\documentclass[dvipsnames]{nature}

\usepackage[british]{babel}
\usepackage[utf8]{inputenc}
\usepackage{babelbib}
\usepackage{url}
\usepackage{graphicx}
\usepackage{subcaption}
\usepackage{calc}
\usepackage{floatflt}
\usepackage{amssymb, amsmath, amsthm}
\usepackage{tabularx}
\usepackage{multirow}
\usepackage{array}
\usepackage{xcolor}
\usepackage{booktabs}
\usepackage{bm}
\usepackage{stmaryrd}
\usepackage{stackrel}
\usepackage{algpseudocode}
\usepackage{algorithm}
\usepackage{rotating}
\usepackage{mwe,tikz}
\usepackage[percent]{overpic}
\usepackage{float}
\usepackage{listings}

\usepackage{numprint}
\usepackage{memhfixc}
\usepackage{makecell}
\usepackage[colorlinks=true,allcolors=blue]{hyperref}
\usepackage{siunitx}
\usepackage{textcomp}
\usepackage{lineno}
\usepackage{hyperref}
\hypersetup{%
    pdfborder = {0 0 0}
}
\usepackage{cleveref}

\usepackage{mathtools}

\usepackage{pdfpages} 
\usepackage{pgffor} 

\makeatletter
\AtBeginDocument{\let\LS@rot\@undefined}
\makeatother

\def\supplementfilename{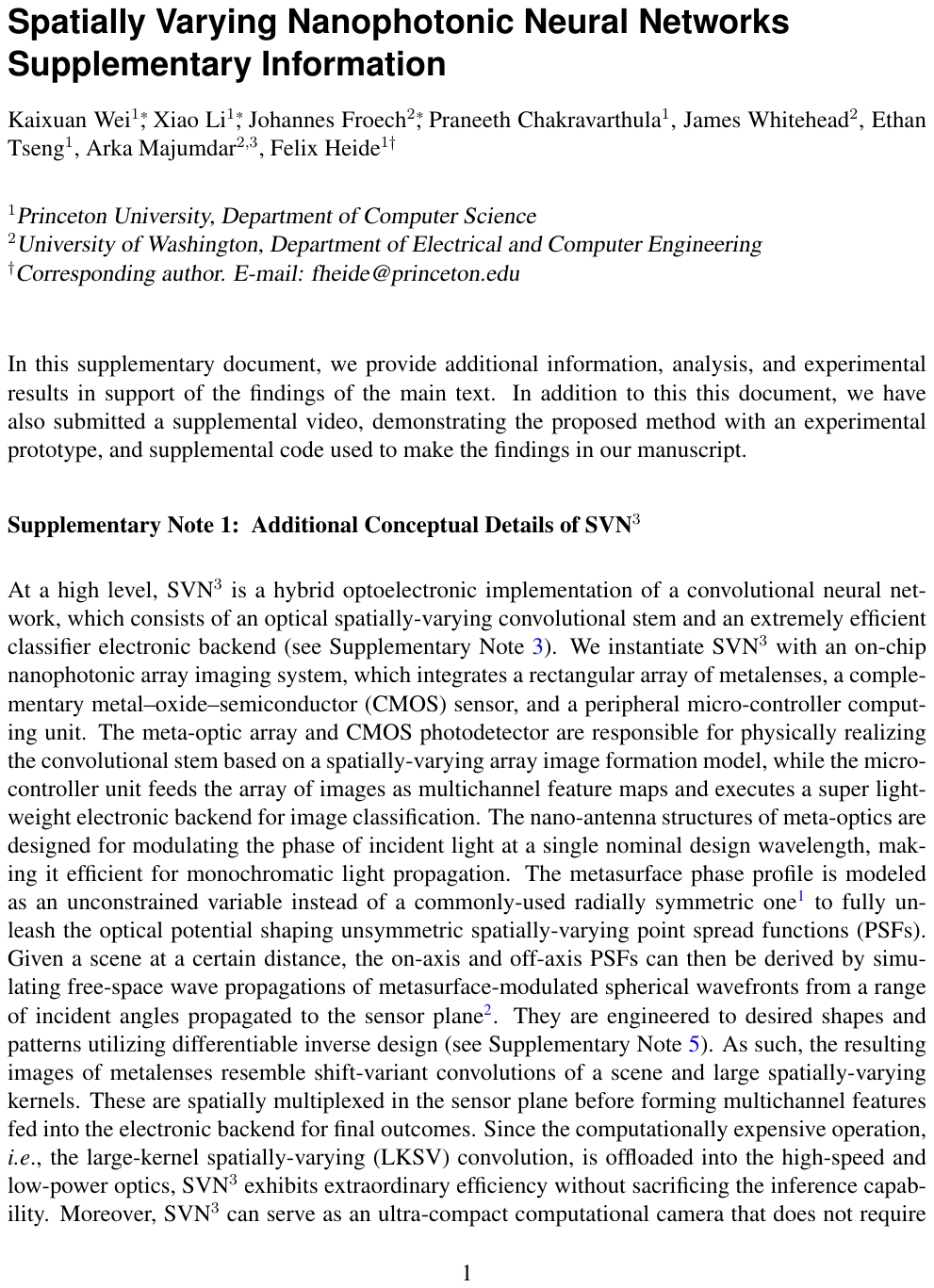}

\pdfximage{\supplementfilename}
\def\numbersupplementpages{\the\pdflastximagepages}

\newif\ifarXiv
\arXivtrue 

\title{Spatially Varying Nanophotonic Neural Networks}

\usepackage[symbol]{footmisc}

\author{Kaixuan Wei$^{1}$\footnotemark[1], Xiao Li$^{1}$\footnotemark[1], Johannes Froech$^2$\footnotemark[1], Praneeth Chakravarthula$^1$,
James Whitehead$^2$, Ethan Tseng$^1$,  Arka Majumdar$^{2,3}$, Felix Heide$^{1\dagger}$}

\begin{document}

\maketitle

\newcommand{\shortname}{SVN$^3$}


\definecolor{Gray}{rgb}{0.5,0.5,0.5}
\definecolor{darkblue}{rgb}{0,0,0.7}
\definecolor{orange}{rgb}{1,.5,0} 
\definecolor{red}{rgb}{1,0,0} 

\newcommand{\heading}[1]{\noindent\textbf{#1}}
\newcommand{\note}[1]{{{\textcolor{orange}{#1}}}}
\newcommand{\todo}[1]{{\textcolor{darkblue}{TODO: #1}}}
\newcommand{\kaixuan}[1]{\textcolor{red}{{[Kaixuan: #1]}}}
\newcommand{\ethan}[1]{\textcolor{purple}{{[Ethan: #1]}}}
\newcommand{\changed}[1]{{\textcolor{blue}{#1}}}
\newcommand{\revision}[1]{{{#1}}}
\newcommand{\place}[1]{ \begin{itemize}\item\textcolor{darkblue}{#1}\end{itemize}}
\newcommand{\de}{\mathrm{d}}

\newcommand{\BEAS}{\begin{eqnarray*}}
\newcommand{\EEAS}{\end{eqnarray*}}
\newcommand{\BEA}{\begin{eqnarray}}
\newcommand{\EEA}{\end{eqnarray}}
\newcommand{\BEQ}{\begin{equation}}
\newcommand{\EEQ}{\end{equation}}
\newcommand{\BIT}{\begin{itemize}}
\newcommand{\EIT}{\end{itemize}}
\newcommand{\BNUM}{\begin{enumerate}}
\newcommand{\ENUM}{\end{enumerate}}

\newcommand{\BA}{\begin{array}}
\newcommand{\EA}{\end{array}}

\makeatletter
\DeclareRobustCommand\onedot{\futurelet\@let@token\@onedot}
\def\@onedot{\ifx\@let@token.\else.\null\fi\xspace}

\def\eg{\emph{e.g}\onedot} \def\Eg{\emph{E.g}\onedot}
\def\ie{\emph{i.e}\onedot} \def\Ie{\emph{I.e}\onedot}
\def\cf{\emph{c.f}\onedot} \def\Cf{\emph{C.f}\onedot}
\def\etc{\emph{etc}\onedot} \def\vs{\emph{vs}\onedot}
\def\wrt{w.r.t\onedot} \def\dof{d.o.f\onedot}
\def\etal{\emph{et al}\onedot}
\makeatother
\newcommand{\aka}{{\it a.k.a.~}}
\newcommand{\ones}{\mathbf 1}

\newcommand{\argmin}{\mathop{\rm argmin}}
\newcommand{\reals}{\mathbb{R}}
\newcommand{\integers}{\mathbb{Z}}
\newcommand{\epi}{\mathop{\bf epi}}
\newcommand{\complex}{\mathbb{C}}
\newcommand{\symm}{{\mbox{\bf S}}}  

\newcommand{\Span}{\mbox{\textrm{span}}}
\newcommand{\Range}{\mbox{\textrm{range}}}
\newcommand{\nullspace}{{\mathcal N}}
\newcommand{\range}{{\mathcal R}}
\newcommand{\Nullspace}{\mbox{\textrm{nullspace}}}
\newcommand{\Rank}{\mathop{\bf Rank}}
\newcommand{\Tr}{\mathop{\bf Tr}}
\newcommand{\diag}{\mathop{\bf diag}}
\newcommand{\lambdamax}{{\lambda_{\rm max}}}
\newcommand{\lambdamin}{\lambda_{\rm min}}

\newcommand{\Expect}{\mathop{\bf E{}}}
\newcommand{\Prob}{\mathop{\bf Prob}}
\newcommand{\erf}{\mathop{\bf erf}}

\newcommand{\Co}{{\mathop {\bf Co}}}
\newcommand{\co}{{\mathop {\bf Co}}}
\newcommand{\dist}{\mathop{\bf dist{}}}
\newcommand{\Ltwo}{{\bf L}_2}
\newcommand{\QED}{~~\rule[-1pt]{8pt}{8pt}}\def\qed{\QED}
\newcommand{\approxleq}{\mathrel{\smash{\makebox[0pt][l]{\raisebox{-3.4pt}{\small$\sim$}}}{\raisebox{1.1pt}{$<$}}}}

\newcommand{\vol}{\mathop{\bf vol}}
\newcommand{\Vol}{\mathop{\bf vol}}
\newcommand{\Card}{\mathop{\bf card}}

\newcommand{\dom}{\mathop{\bf dom}}
\newcommand{\aff}{\mathop{\bf aff}}
\newcommand{\cl}{\mathop{\bf cl}}
\newcommand{\Angle}{\mathop{\bf angle}}
\newcommand{\intr}{\mathop{\bf int}}
\newcommand{\relint}{\mathop{\bf rel int}}
\newcommand{\bd}{\mathop{\bf bd}}
\newcommand{\vect}{\mathop{\bf vec}}
\newcommand{\dsp}{\displaystyle}
\newcommand{\foequal}{\simeq}
\newcommand{\VOL}{{\mbox{\bf vol}}}
\newcommand{\xopt}{x^{\rm opt}}

\newcommand{\Xb}{{\mbox{\bf X}}}
\newcommand{\xst}{x^\star}
\newcommand{\varphist}{\varphi^\star}
\newcommand{\lambdast}{\lambda^\star}
\newcommand{\Zst}{Z^\star}
\newcommand{\fstar}{f^\star}
\newcommand{\xstar}{x^\star}
\newcommand{\xc}{x^\star}
\newcommand{\lambdac}{\lambda^\star}
\newcommand{\lambdaopt}{\lambda^{\rm opt}}

\newcommand{\geqK}{\mathrel{\succeq_K}}
\newcommand{\gK}{\mathrel{\succ_K}}
\newcommand{\leqK}{\mathrel{\preceq_K}}
\newcommand{\lK}{\mathrel{\prec_K}}
\newcommand{\geqKst}{\mathrel{\succeq_{K^*}}}
\newcommand{\gKst}{\mathrel{\succ_{K^*}}}
\newcommand{\leqKst}{\mathrel{\preceq_{K^*}}}
\newcommand{\lKst}{\mathrel{\prec_{K^*}}}
\newcommand{\geqL}{\mathrel{\succeq_L}}
\newcommand{\gL}{\mathrel{\succ_L}}
\newcommand{\leqL}{\mathrel{\preceq_L}}
\newcommand{\lL}{\mathrel{\prec_L}}
\newcommand{\geqLst}{\mathrel{\succeq_{L^*}}}
\newcommand{\gLst}{\mathrel{\succ_{L^*}}}
\newcommand{\leqLst}{\mathrel{\preceq_{L^*}}}
\newcommand{\lLst}{\mathrel{\prec_{L^*}}}

\newtheorem{theorem}{Theorem}[section]
\newtheorem{corollary}{Corollary}[theorem]
\newtheorem{lemma}[theorem]{Lemma}
\newtheorem{proposition}[theorem]{Proposition}

\newenvironment{algdesc}%
{\begin{quote}}{\end{quote}}

\def\figbox#1{\framebox[\hsize]{\hfil\parbox{0.9\hsize}{#1}}}

\makeatletter
\long\def\@makecaption#1#2{
   \vskip 9pt
   \begin{small}
   \setbox\@tempboxa\hbox{{\bf #1:} #2}
   \ifdim \wd\@tempboxa > 5.5in
        \begin{center}
        \begin{minipage}[t]{5.5in}
        \addtolength{\baselineskip}{-0.95pt}
        {\bf #1:} #2 \par
        \addtolength{\baselineskip}{0.95pt}
        \end{minipage}
        \end{center}
   \else
    \hbox to\hsize{\hfil\box\@tempboxa\hfil}
   \fi
   \end{small}\par
}
\makeatother

\newcounter{oursection}
\newcommand{\oursection}[1]{
 \addtocounter{oursection}{1}
 \setcounter{equation}{0}
 \clearpage \begin{center} {\Huge\bfseries #1} \end{center}
 {\vspace*{0.15cm} \hrule height.3mm} \bigskip
 \addcontentsline{toc}{section}{#1}
}
\newcommand{\oursectionf}[1]{  
 \addtocounter{oursection}{1}
 \setcounter{equation}{0}
 \foilhead[-.5cm]{#1 \vspace*{0.8cm} \hrule height.3mm }
 \LogoOn
}
\newcommand{\oursectionfl}[1]{  
 \addtocounter{oursection}{1}
 \setcounter{equation}{0}
 \foilhead[-1.0cm]{#1}
 \LogoOn
}

\newcommand{\Mat}[1]    {{\ensuremath{\mathbf{\uppercase{#1}}}}} 
\newcommand{\Vect}[1]   {{\ensuremath{\mathbf{\lowercase{#1}}}}} 
\newcommand{\Vari}[1]   {{\ensuremath{\mathbf{\lowercase{#1}}}}} 
\newcommand{\Id}				{\mathbb{I}} 
\newcommand{\Diag}[1] 	{\operatorname{diag}\left({ #1 }\right)} 
\newcommand{\Opt}[1] 	  {{#1}_{\text{opt}}} 
\newcommand{\CC}[1]			{{#1}^{*}} 
\newcommand{\Op}[1]     {\Mat{#1}} 
\newcommand{\mini}[1] {{\mbox{argmin}}_{#1} \: \: } 
\newcommand{\argmax}[1] {\underset{{#1}}{\mathop{\rm argmax}} \: \: } 
\newcommand{\minimize}{\mathop{\rm minimize} \: \:}
\newcommand{\minimizeu}[1]{\underset{{#1}}{\mathop{\rm minimize}} \: }
\newcommand{\grad}      {\nabla}
\newcommand{\kron}{\otimes} 

\newcommand{\gradt}     {\grad_\z}
\newcommand{\gradx}     {\grad_\x}
\newcommand{\Drv}     	{\Mat{D}} 
\newcommand{\step}      {\text{\textbf{step}}}
\newcommand{\prox}[1]   {\mathbf{prox}_{#1}}
\newcommand{\ind}[1]    {\operatorname{ind}_{#1}}
\newcommand{\proj}[1]   {\Pi_{#1}}
\newcommand{\pointmult}{\odot} 
\newcommand{\rr}   {\mathcal{R}}

\newcommand{\Basis}{\Mat{D}}         		
\newcommand{\Corr}{\Mat{C}}             
\newcommand{\conv}{\ast} 
\newcommand{\meas}{\Vect{b}}            
\newcommand{\Img}{I}                    
\newcommand{\img}{\Vect{i}}             
\newcommand{\vv}{\Vect{v}}
\newcommand{\p}{\Vect{p}}
\newcommand{\Splitvar}{T}                
\newcommand{\splitvar}{\Vect{t}}         
\newcommand{\Splitbasis}{J}                
\newcommand{\splitbasis}{\Vect{j}}         
\newcommand{\var}{\Vari{z}} 

\newcommand{\FT}[1]			{\mathcal{F}\left( {#1} \right)} 
\newcommand{\IFT}[1]			{\mathcal{F}^{-1}\left( {#1} \right)} 

\newcommand{\func}{f}
\newcommand{\fMat}{\Mat{K}}

\newcommand{\avar}{\Vari{v}} 
\newcommand{\aspvar}{\Vari{z}} 

\newcommand{\mask}{\Mat{M}}

\newcommand{\Pen}      		{F} 
\newcommand{\cardset}     {\mathcal{C}}
\newcommand{\Dat}      		{G} 
\newcommand{\Reg}      		{\Gamma} 

\newcommand{\Trans}{\mathbf{\uppercase{T}}} 
\newcommand{\Ph}{\mathbf{\uppercase{\Phi}}} 

\newcommand{\Tvec}{\Vect{T}} 
\newcommand{\Bvec}{\Vect{B}} 

\newcommand{\Wt}{\Mat{W}} 

\newcommand{\Perm}{\Mat{P}} 

\newcommand{\DiagFactor}[1]     {\Mat{O}_{ #1 }}  

\newcommand{\Proj}{\Mat{P}}             

\newcommand{\Vector}[1]{\mathbf{#1}}
\newcommand{\Matrix}[1]{\mathbf{#1}}
\newcommand{\Tensor}[1]{{\ensuremath{\boldsymbol{\mathscr{#1}}}}}
\newcommand{\TensorUF}[2]{\Matrix{#1}_{(#2)}}

\newcommand{\MatrixKP}[1]{\Matrix{#1}_{\otimes}}
\newcommand{\MatrixKPN}[2]{\Matrix{#1}_{\otimes}^{#2}}

\newcommand{\MatrixKRP}[1]{\Matrix{#1}_{\odot}}
\newcommand{\MatrixKRPN}[2]{\Matrix{#1}_{\odot}^{#2}}

\newcommand{\HP}{\circ}
\newcommand{\HD}{\oslash}

\newcommand{\leftDB}{\left[ \! \left[}
\newcommand{\rightDB}{\right] \! \right]}

\newcommand{\transpose}{T}

\newcommand*\sstrut[1]{\vrule width0pt height0pt depth#1\relax}

\newcommand{\inlineeqnum}{\refstepcounter{equation}~~\mbox{(\theequation)}}
\newcommand{\eqname}[1]{\tag*{#1~(\theequation)}\refstepcounter{equation}}

\newcommand{\lambdas}{\boldsymbol{\lambda}}
\newcommand{\alb}{\boldsymbol{\alpha}} 	
\newcommand{\depth}{\boldsymbol{z}} 	
\newcommand{\albi}{\alpha} 	
\newcommand{\depthi}{z} 	
\newcommand{\ambient}{s}
\newcommand{\jitter}{w}
\newcommand{\z}{\Vect{z}} 							
\newcommand{\x}{\Vect{x}}             	
\newcommand{\y}{\Vect{y}}             	
\newcommand{\Kvar}{\Mat{K}}
\newcommand{\lagrangemult}{\boldsymbol{\nu}}
\newcommand{\scaledlagrange}{\Vect{u}}
\newcommand{\eps}{\epsilon}
\newcommand{\vp}{\Vect{v}}

\begin{affiliations}
 \item Princeton University, Department of Computer Science
 \item University of Washington, Department of Electrical and Computer Engineering
 \item [$^\dagger$] Corresponding author. E-mail: fheide@princeton.edu
\end{affiliations}

\footnotetext{$^*$Indicates equal contribution.}

\begin{abstract}
The explosive growth of computation and energy cost of artificial intelligence has spurred strong interests in new computing modalities as potential alternatives to conventional electronic processors. Photonic processors that execute operations using photons instead of electrons, have promised to enable optical neural networks with ultra-low latency and power consumption. However, existing optical neural networks, limited by the underlying network designs, have achieved image recognition accuracy far below that of state-of-the-art electronic neural networks. In this work, we close this gap by embedding massively parallelized optical computation into flat camera optics that perform neural network computation during the capture, before recording an image on the sensor. Specifically, we harness large kernels and propose a large-kernel spatially-varying convolutional neural network learned via low-dimensional reparameterization techniques. We experimentally instantiate the network with a flat meta-optical system that encompasses an array of nanophotonic structures designed to induce angle-dependent responses. Combined with an extremely lightweight electronic backend with approximately 2K parameters we demonstrate a reconfigurable nanophotonic neural network reaches 72.76\% blind test classification accuracy on CIFAR-10 dataset, and, as such, the first time, an optical neural network outperforms the first modern digital neural network -- AlexNet (72.64\%) with 57M parameters, bringing optical neural network into modern deep learning era. 
\end{abstract}


\section*{Introduction}

Increasing demands for high-performance artificial intelligence in the last decade have levied immense pressure on computing architectures across domains, including robotics, transportation, personal devices, medical imaging and scientific imaging. 
Although electronic microprocessors have undergone drastic evolution over the past 50 years\cite{moore1998cramming}, providing us with general-purpose CPUs and custom accelerator platforms (\eg, GPU, DSP ASICs), this growth rate is far outpaced by the explosive growth of AI models.
Specifically, Moore's law delivers a doubling in transistor counts every two years\cite{waldrop2016chips} whereas deep neural networks (DNN)\cite{lecun2015deep}, arguably the most influential algorithms in AI, have doubled in size every six months\cite{sevilla2022compute}. However, in fact, the end of voltage scaling has made the power consumption, and not the number of transistors, the principal factor limiting further improvements in computing performance\cite{horowitz20141}. Overcoming this limitation and radically reducing compute latency and power consumption could drive unprecedented applications from low-power edge computation in the camera, potentially enabling computation in thin eye-glasses or micro-robots, and reducing power consumption in data centers used for training of neural network architectures.

Optical computing has been proposed as a potential avenue to alleviate several inherent limitations of digital electronics, \eg, compute speed, heat dissipation, and power, and could potentially boost computational throughput, processing speed, and energy efficiency by orders of magnitude\cite{solli2015analog,caulfield2010future,miller2017attojoule}. Such optical computers leverage several advantages of photonics to achieve high throughput, low latency, and low power consumption\cite{mcmahon2023physics}. These performance improvements are achieved by sacrificing reconfigurability. Thus, although general-purpose optical computing has yet to be practically realized due to obstacles such as larger physical footprints and inefficient optical switches\cite{miller2010optical,tucker2010role}, several significant advances have already been made towards optical/photonic processors tailored specifically for AI\cite{wetzstein2020inference,shastri2021photonics,wu2021analog}.
Representative examples include optical computers that perform widely-used signal processing operators\cite{liu2016fully,kwon2018nonlocal,silva2014performing,zhu2017plasmonic,ferrera2010chip}, (\eg, spatial/temporal differentiation, integration, and convolution) and mathematical solvers\cite{xu2020scalable,mohammadi2019inverse} with performance far beyond those of contemporary electronic processors. Most strikingly, optical neural networks (ONN)\cite{feldmann2019all,xu202111,shen2017deep,feldmann2021parallel,ashtiani2022chip,tait2017neuromorphic,teugin2021scalable,lin2018all,mengu2019analysis,yan2019fourier,rahman2021ensemble,luo2022metasurface,hamerly2019large,zhou2021large,shi2022loen,zheng2022meta,chang2018hybrid,chen2023accel} can perform AI inference tasks such as image recognition when implemented as fully-optical or hybrid opto-electronical computers.

Existing ONNs can be broadly classified into two categories based on either integrated photonics\cite{feldmann2019all,xu202111,shen2017deep,feldmann2021parallel,ashtiani2022chip,tait2017neuromorphic,teugin2021scalable} (\eg, Mach–Zehnder interferometers\cite{shen2017deep}, phase change materials\cite{feldmann2019all}, microring resonators \cite{tait2017neuromorphic}, multimode fibers\cite{teugin2021scalable}) for physically realizing multiply–accumulate (MAC) operations, or with free-space optics\cite{lin2018all,mengu2019analysis,yan2019fourier,rahman2021ensemble,luo2022metasurface,hamerly2019large,zhou2021large,shi2022loen,zheng2022meta,chang2018hybrid,colburn2019optical} that implement convolutional layers with light propagation through diffractive elements (\eg, 3D-printed surfaces\cite{lin2018all}, 4F optical correlators\cite{chang2018hybrid}, optical masks\cite{shi2022loen}, meta-surfaces \cite{zheng2022meta}). 
The design of these ONN architectures has been fundamentally restricted by the underlying network design, including the challenge of scaling to large numbers of neurons (within integrated photonic circuits) and the lack of scalable energy-efficient nonlinear optical operators. 
As a result, even the most successful ensemble ONNs\cite{rahman2021ensemble} that employ dozens of ONNs in parallel, have only achieved LeNet\cite{lecun1989handwritten}-level accuracy on image classification, which was achieved by their electronic counterparts over 30 years ago. 
Moreover, most high-performance ONNs can only operate under coherent illumination,
prohibiting the integration into the camera optics under natural lighting conditions. Although hybrid opto-electronic networks\cite{shi2022loen,zheng2022meta} working on incoherent light do exist, they do not yield favorable results as their optical frontend is designed to execute only a single convolutional layer.

In this work, we report a novel nanophotonic neural network that lifts the aforementioned limitations, allowing us to close the gap to the first modern DNN architectures\cite{krizhevsky2012imagenet} with optical compute in a flat form factor of only $\SI{4}{mm}$ length -- akin to performing computation on the sensor cover glass. We leverage the ability of a lens system to perform large-kernel spatially-varying convolutions tailored specifically for image recognition and semantic segmentation. These operations are performed during the capture before the sensor makes a measurement. We learn large kernels via low-dimensional reparameterization techniques which circumvent spurious local extremum caused by direct optimization. To physically realize the ONN, we develop a differentiable spatially varying inverse design framework that solves for metasurfaces\cite{khorasaninejad2017metalenses,dorrah2022tunable} that can produce the desired angle-dependent responses under spatially incoherent illumination. Because of the compact footprint and CMOS sensor compatibility, the resulting optical system is not only a photonic accelerator, but also an ultra-compact computational camera that directly operates on the ambient light from the environment before the analog to digital conversion. We find that this approach facilitates generalization and transfer learning to other tasks, such as semantic segmentation, outperforming AlexNet\cite{krizhevsky2012imagenet} in 1000-category ImageNet~\cite{deng2009imagenet} classification and PASCAL VOC~\cite{everingham2010pascal} semantic segmentation.

By on-chip integration of the flat-optics frontend ($>99$\% MACs) with an extremely lightweight electronic backend ($<1$\% MACs), we achieve higher classification performance than modern fully-electronic classifiers (73.80\% in simulation and 72.76\% in experiment, compared to 72.64\% by AlexNet\cite{krizhevsky2012imagenet} on CIFAR-10\cite{krizhevsky2009learning} test set) while simultaneously reducing the number of electronic parameters by four orders of magnitude, thus bringing optical neural networks into the modern deep learning era.

\section*{Results}

\begin{figure*}[t]
    \centering
    \includegraphics[width=\textwidth]{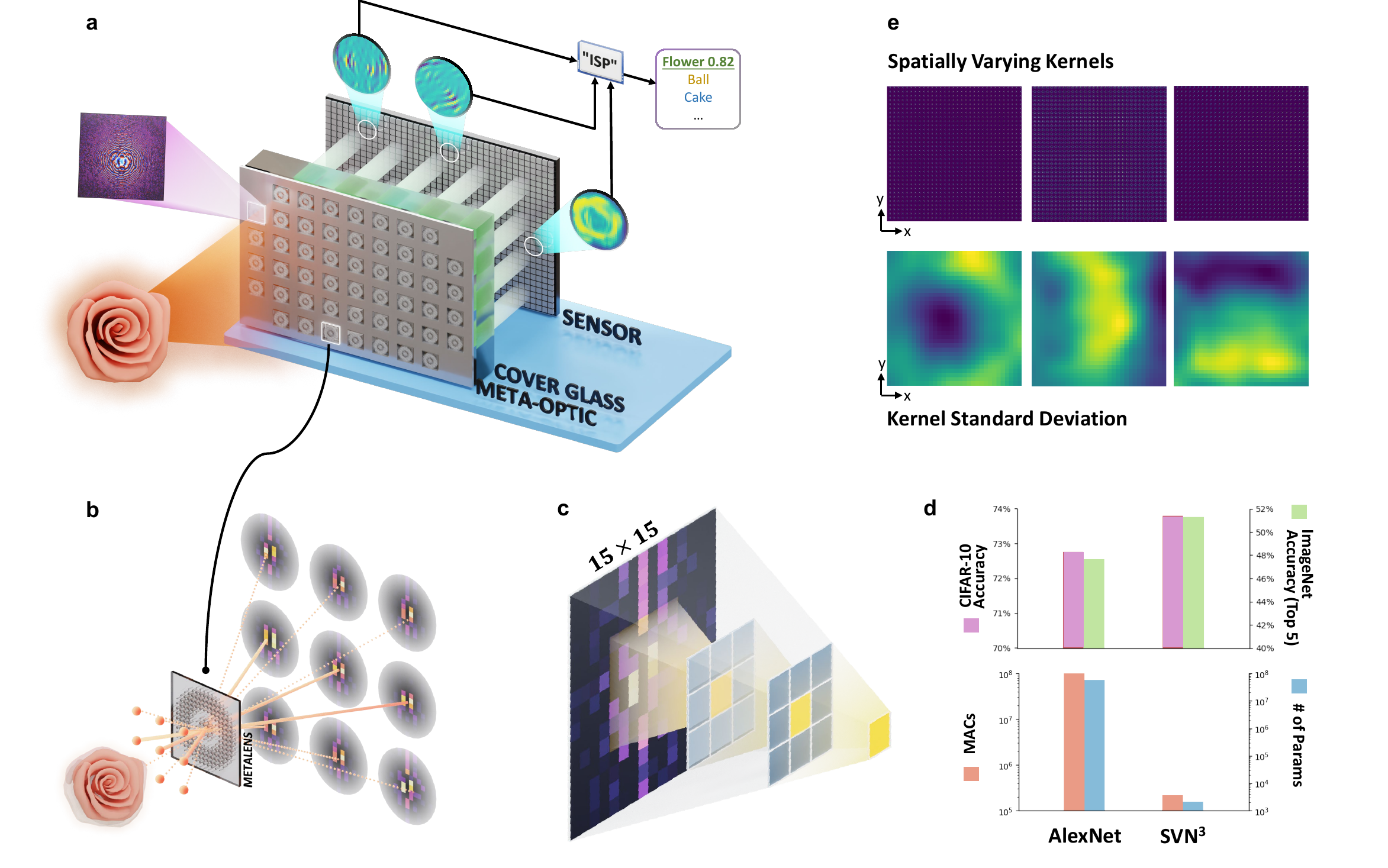}
    \caption{\textbf{Spatially varying nanophotonic neural networks}. (a) Illustration of the proposed opto-electronic network, which comprises a nanophotonic array front-end that optically encodes the scene into multichannel image features and a lightweight electronic back-end that performs the final prediction, in a programmable manner, for image classification or semantic segmentation; (b) Each metalens is designed for specific learned large and angularly varying point spread functions that comprise the feature kernels of the early network layers which vary over the sensor. These kernels are learned electronically using a spatially varying reparameterization. (c) We learn large kernels of size $15 \times 15$ (for digital $32 \times 32$ image classification) by factorizing them into a cascade of smaller ones. (d) Assessment of purely electronic AlexNet~\cite{krizhevsky2012imagenet} compared to \shortname~: we report classification accuracies on CIFAR-10 and ImageNet datasets (top barplot), digital multiply–accumulate (MACs) operations, and digital parameters (bottom barplot) for CIFAR-10 image recognition. The proposed method outperforms a network with multiple orders of magnitude more electronic parameters with multiple orders of magnitude fewer multiply–accumulate operations, see Table~\note{S2} for details.
    (e) Representative spatially-varying kernels plotted over space (see Figure~\note{S3} for high-resolution illustration) and the corresponding kernel standard deviation, illustrating the variation, at each spatial location (second row). }    
    \label{fig:metaonn-overview}
\end{figure*}

\paragraph{Large-Kernel Spatially-Varying Parameterization}
The working principle and optoelectronic implementation of the proposed spatially varying nanophotonic neural network (\shortname) are illustrated in Figure~\hyperref[fig:metaonn-overview]{1a}. The \shortname~is an optoelectronic neuromorphic
computer 
that comprises a metalens array nanophotonic front-end and a lightweight electronic back-end (embedded in a low-cost micro-controller unit) for image classification or semantic segmentation. The metalens array front-end consists of 50 metalens elements that are made of 390 nm pitch nano-antennas and are optimized for incoherent light in a band around 525 nm. The wavefront modulation induced by each metalens can be represented by the optical convolution of the incident field and the point spread functions (PSF) of the individual device. Therefore, the nanophotonic front-end performs parallel multichannel convolutions, at the speed of light, without any power consumption. We also refer to  Supplementary Note \note{1} and \note{3} for additional details on the physical forward model and the neural network design, respectively.


Unlike existing ONNs\cite{chang2018hybrid,colburn2019optical,shi2022loen,zheng2022meta,fu2022ultracompact} that engineer the optical response to mimic a convolutional layer that consists of spatially-invariant small-sized kernels, the \shortname~ employs large-sized angularly varying PSFs (Figure~\hyperref[fig:metaonn-overview]{1b}) as the convolution kernels to construct a large-kernel spatially-varying (LKSV) convolutional layer.
Such an LKSV convolutional layer is not used in conventional deep neural networks due to immense computation costs and challenges in training. Nevertheless, we demonstrate that with low-dimensional reparametrization techniques, namely large kernel factorization, and low-rank spatially-varying reparameterization, this computing layer can be effectively learned in silicon,  circumventing spurious local minima that can arise from na\"ive over-parametrization (Supplementary Note \note{3}).

We reparameterize a large ($15\times15$) convolutional kernel into a stack of (seven) small $3\times3$ kernels, which are convolved sequentially to the large kernel (Figure~\hyperref[fig:metaonn-overview]{1c}). The spatially-varying structure is reparameterized through a spatially-variant weighted linear combination of a (large) kernel basis, which resembles the low-rank approximation of a general spatially-varying kernel.
As such, we construct a 3-layer convolutional neural network (CNN) composed of an LKSV convolutional stem, a depth-wise separable convolutional layer, and a fully-connected classification head, for CIFAR-10 image classification. This CNN is trained in silicon by minimizing the standard cross-entropy loss with tailored regularizations (an isotropic total variation regularization and a specialized spectrum regularization) on the spatially-varying kernels
(Supplementary Note \note{4}). Validated by the spatial combining weights and the Fourier spectrum profiles of learned kernels in Figure \note{S2}, these regularizations enforce smooth transitions of spatially-varying kernels (Figure~\hyperref[fig:metaonn-overview]{1e}) and penalize high-pass and ill-conditioned kernels, which are challenging to implement in an optical system. 
After in-silicon training, our LKSV design performs favorably compared to the conventional small-kernel spatially-invariant (SKSI) counterpart by a sizable margin,
lifting from the LeNet-level accuracy (65.45\%) to the AlexNet-level accuracy (73.80\%), see also Figure~\hyperref[fig:metaonn-overview]{1d} and Table~\note{S1}, \note{S2}. 

The high computational cost of LKSV convolution in silicon can be entirely eliminated by designing a passive optical system with metalenses whose PSFs are inverse designed to mimic the designated target kernels. While the target kernels may contain both positive and negative values, optical PSFs contain only non-negative values. Thus, to generate each target kernel we employ a pair of metalenses and we take the subtraction of their image features post-convolution to achieve positive and negative values.

To optically realize a 25-channel LKSV convolutional layer, we instantiate an on-chip metalens array that consists of 50 metalenses with the device layout shown in Figure~\hyperref[fig:metaonn-overview]{1a} and \hyperref[fig:exp-val-cifar10-1]{2a}.
To engineer spatially-varying PSFs, we simulate the optical system and use a differentiable spatially-varying inverse design framework 
to compute the phase profiles of the metalenses via stochastic gradient-based optimization.
The angularly varying PSFs are optimized by minimizing the mean square error loss with respect to the target electronic kernels and employing an energy regularization to maximize the localized energy in the region of interest on the sensor plane. By employing energy regularization, we improve the light efficiency of the designed metalenses from 39.37\% to 93.88\% without impacting the PSF accuracy 
and make the ONNs more robust to unwanted scattering light and other noise in real-world measurement (Supplementary Note \note{5}).

\paragraph{Experimental Validation}
The inverse-design-optimized metalens array was fabricated on a single chip in a silicon nitride on quartz film. We used a nanopatterning approach using electron beam lithography to define the outline of the design in a resist, deposited a hard mask, and subsequently transferred the pattern into the underlying silicon nitride using reactive ion etching. To exclude transmission of light through non-patterned sections we further deposited a metal aperture around the ONN metalens kernels. 
The close-up of the resulting metalens array camera and a metalens array device before mounting are shown in Figure~\hyperref[fig:exp-val-cifar10-1]{2a}. 
The PSFs (over $3\times3$ varied sampling incident angles) of three randomly selected kernels are illustrated in Figure~\hyperref[fig:exp-val-cifar10-1]{2c}, which illustrates the spatially-varying features of the designed optical kernels. To experimentally realize the optical system and measure the image features of the metalenses, we devise the setup shown in Figure~\hyperref[fig:exp-val-cifar10-1]{2b}. The green channel of a smartphone OLED display, which is placed at the designed object distance, is used as the incoherent light source, and a large-area CMOS sensor is placed at the focal plane of the metalens array device. When the dataset images are displayed on the display, the sensor captures the corresponding image features of all the metalens elements in a \emph{single shot}.
The captured positive, negative, and real-valued features through subtraction closely resemble the electronic ground truth from both qualitative and quantitative comparisons, which verifies the effectiveness of the implemented inverse design framework (Figure~\hyperref[fig:exp-val-cifar10-1]{2d}, \hyperref[fig:exp-val-cifar10-2]{3a}).
Interested readers are also referred to the Supplementary Video \note{1} for prototype demonstration of \shortname~for dynamic content. 

To extensively assess the performance of our opto-electronic neural network \shortname, we captured the entire grayscale CIFAR-10 dataset, including 50,000 training images and 10,000 test images, with the setup described above and shown in Figure~\hyperref[fig:exp-val-cifar10-1]{2b}. The image features in each frame are equally spaced in a regular $6\times9$ array with the four corners being traditional hyperbolic metalenses used for device alignment (Figure~\note{S4}).
After cropping the image features of all the metalenses and computing the real-valued target features through paired subtraction, the resulting multichannel optical features are fed into the pretrained lightweight electronic backend to obtain the final predictions.  We finetune the electronic backend using the cross-entropy loss on the experimentally captured CIFAR-10 training dataset. The finetuning procedure is identical to the prior in-silicon training of the target electronic neural network, except no extra regularization losses are applied (Supplementary Note \note{4}). \shortname~reaches to $72.76\%$ on the CIFAR-10 test dataset, which is comparable to $73.80\%$ of the corresponding electronic model. 
Similar observations are also drawn in the confusion matrices in Figure~\hyperref[fig:exp-val-cifar10-2]{3b}, which reveals the similar recognition behavior of the \shortname~in real experiment and simulation. Figure~\ref{fig:cls-vis-res} reports predictions on random samples from CIFAR-10 testset. The method consistently assigns a high probability to the true class (Top-2).
These experimental results collectively validate the effectiveness of \shortname~in classifying common objects, extending beyond the realm of handwritten digit recognition investigated in existing work.
Furthermore, we emphasize that almost all computation ($> 99\%$ of MACs) of \shortname~is executed on the optical side with zero energy consumption (Table~\note{S2}). This AlexNet-level classification accuracy is thus achieved with an ultra-low power device.

\paragraph{Versatile Reconfigurable Computational Camera} Our approach is generic which we validate by instantiating \shortname~for other datasets and tasks. Next, we describe such an instance for ImageNet classification with 1000 object categories. ImageNet is the first large-scale image classification dataset with 1.28 million labeled training data, serving as a major driver to advance modern AI. To the best of our knowledge, no existing ONN has reported results on 1000-class ImageNet classification so far. To tackle this challenging 1000-class recognition task, we use an enlarged electronic backend with four depth-wise separable convolutional layers and one fully-connected classifier. We inverse design and fabricate an on-chip metasurface array to optically encode features for $64\times64$ low-resolution ImageNet classification. 
Akin to the CIFAR-10 experiment, the entire training and validation datasets of ImageNet are encoded into optical features by the imaging system for finetuning and evaluation. The experimentally captured features consistently align with their electronic ground truth (Figure~\hyperref[fig:exp-imagenet]{5a}), validating the scalability and effectiveness of \shortname~to process large-sized image features.  
After finetuning the electronic backend on the ImageNet training set, \shortname~achieves $48.64\%$ top-5 classification accuracy in ImageNet validation set, outperforming AlexNet ($47.60\%$) by $1.03\%$. 
Note that the \shortname~for $64\times64$ ImageNet classification has 1.67M digital multiply-accumulate operations (MACs), which is only $0.9\%$ of AlexNet (180.26M). 

Although the optical frontend (encoder) in \shortname~ is not programmable after being fabricated, we demonstrate that \shortname~can serve as a reconfigurable versatile computational camera with a universal optical encoder. By adjusting the electronic backend (decoder) using transfer learning, \shortname~is capable of performing diverse vision tasks beyond the initially designed task. Using the same physical setup for ImageNet classification, we conduct image recognition experiments on the CIFAR-100\cite{krizhevsky2009learning}, Flowers-102\cite{nilsback2008automated}, Food-101\cite{bossard2014food} and Pet-37\cite{parkhi2012cats} datasets. For all of these datasets, we achieve comparable or better performance than AlexNet (Figure~\hyperref[fig:exp-imagenet]{5b}), consistently validating the flexibility of our hybrid opto-electronic system without adapting the optical frontend. We also validate this capability for other computer vision tasks, \eg, semantic segmentation in PASCAL VOC\cite{everingham2010pascal} dataset, where our hybrid network is competitive to the AlexNet-based segmentation network as validated in Figure~\hyperref[fig:exp-imagenet]{5c}. Our \shortname~achieves a pixel accuracy of $65.73\%$ compared with $66.34\%$ of AlexNet-based segmentation on the PASCAL VOC test set.

\section*{Discussion}
In this work, we investigate a novel nanophotonic neural network, that lifts the limitations of existing optical neural networks, propelling them to performance parity with the first modern digital neural network, AlexNet. To this end, we embed computation in the camera lens, performed during the image capture, and we exploit the spatially varying nature of large optical aberrations. Specifically, we propose a large-kernel spatially-varying convolutional neural network, learned via low-dimensional reparameterization techniques, and physically realizing it via a meta-optical system. We find that this approach achieves an image classification accuracy of (top-1) $72.76\%$ on CIFAR-10 and (top-5) $48.64\%$ on (1000-class) ImageNet. The proposed method shifts almost all computation from electronic processors into the optical domain, while allowing for an ultra-thin optical stack of only 4~mm, akin to performing computation on the sensor cover glass. Specifically, we reduce the number of multiply-add floating point operations by $99.64\%$, while ensuring generalization to diverse vision tasks without needing to fabricate new optics.
We believe that the proposed optical neural network is a first step to bridging the gap between photonic and electronic artificial intelligence, and we anticipate that these devices could enable ultra-low latency computing at the edge.

\begin{figure*}[t]
    \centering
    \includegraphics[width=0.95\textwidth]{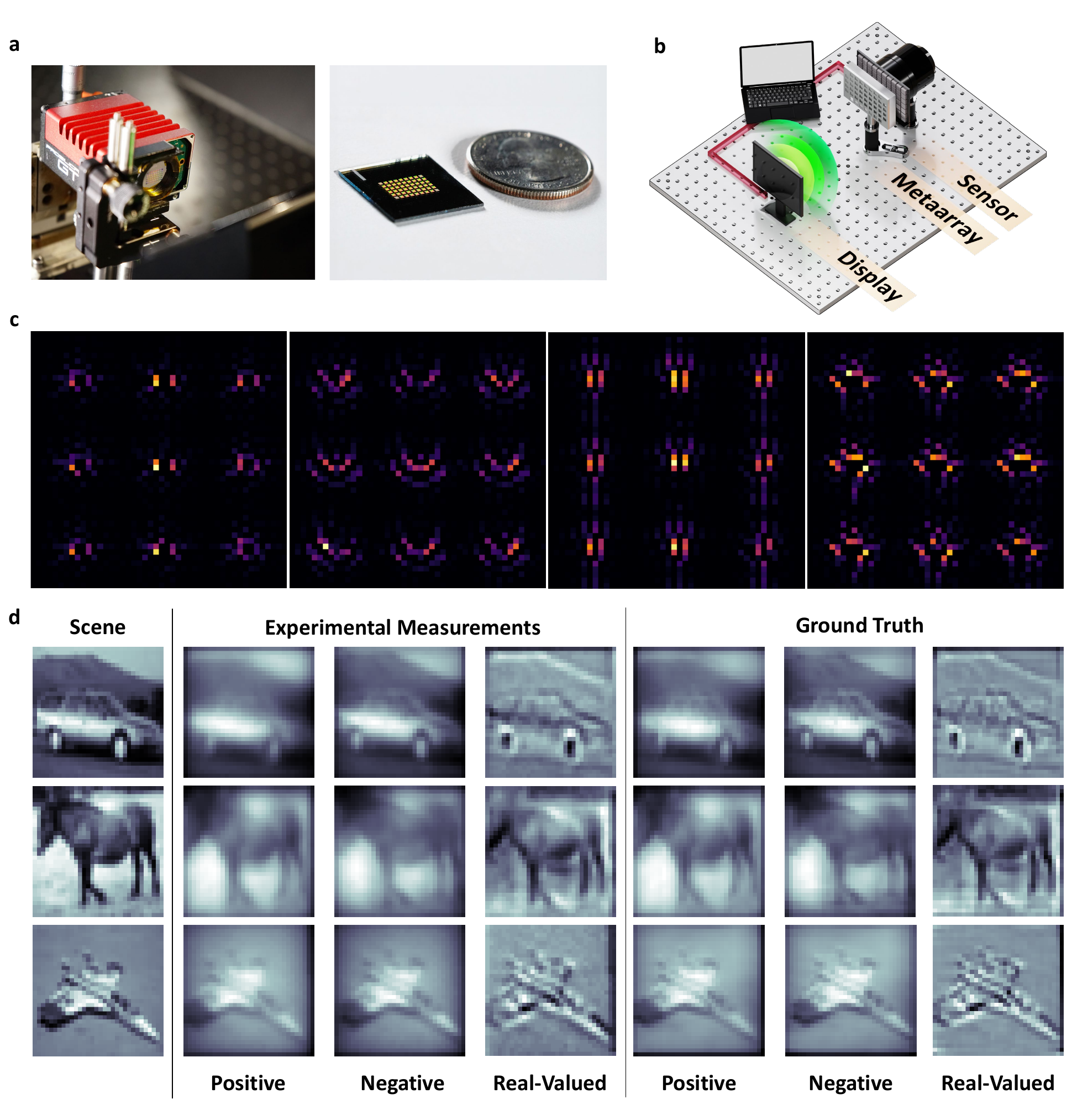}
    \caption{\textbf{Experimental validation of \shortname~}. (a) Flat camera prototype (left) and a metalens array device before mounting (right); (b) Illustration of the experimental setup, consisting of an OLED display placed at the designated object distance, metalens array, and CMOS sensor. Note that no additional optics are used. Camera and display are synchronized for data capture; (c) Spatially-varying PSF visualization on a $3\times 3$ sampling grid of incident angles. Here, we show four representative kernels; (d) Side-by-side comparison of the experimental measurements that match the corresponding ground truth feature channels. ``Real-valued'' denotes the target feature channel, the negative image feature subtracted from positive image features post-convolution.}
    \label{fig:exp-val-cifar10-1}
\end{figure*}

\begin{figure*}[t]
    \centering
    \includegraphics[width=\textwidth]{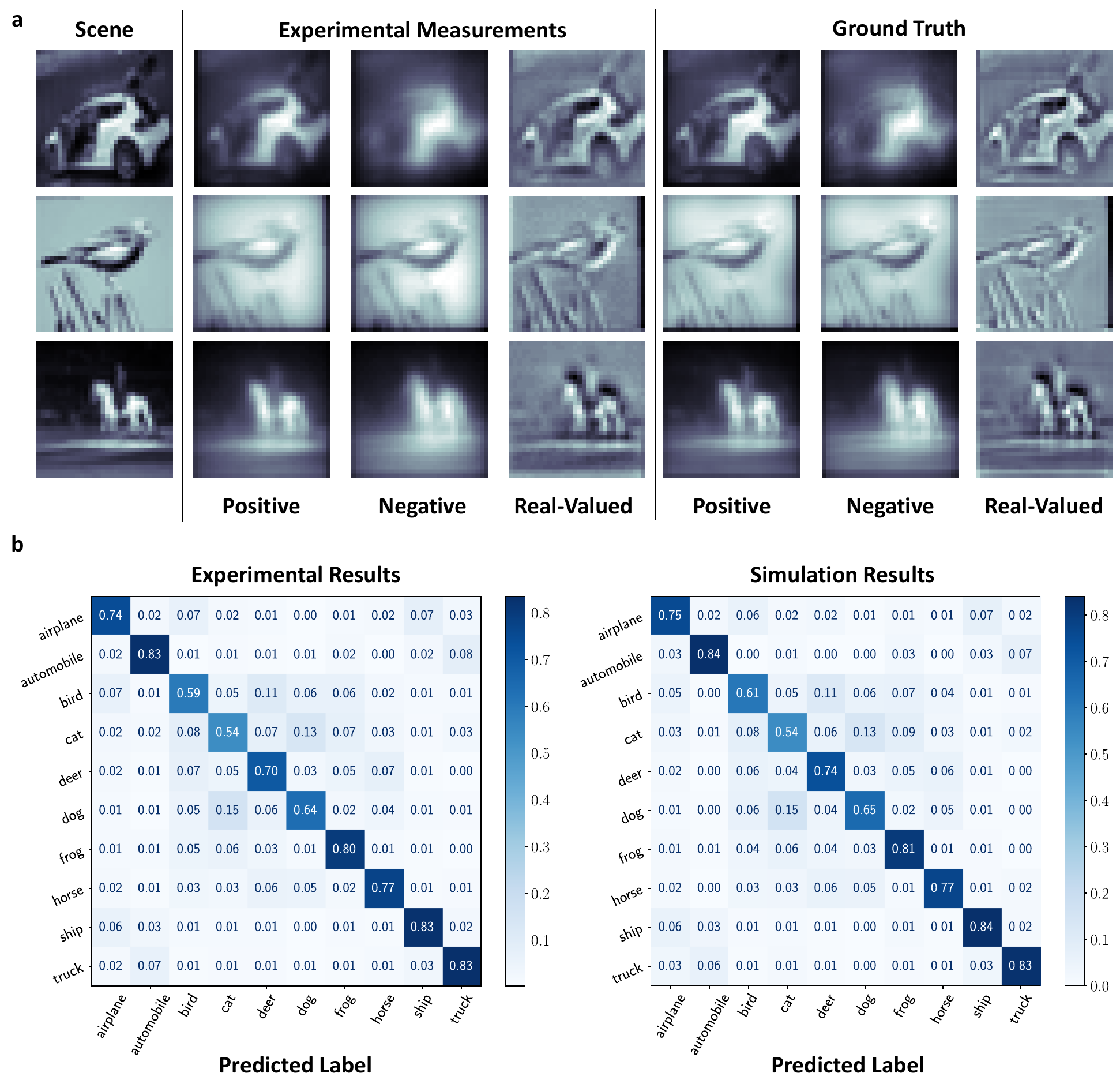}
    \caption{\textbf{Experimental measurements of a fabricated chip of a design for CIFAR-10 image classification}. (a) Qualitative assessment of the experimental measurements compared with the ground truth feature channels. ``Real-valued'' again denotes the target feature channels via subtracting the negative from the positive image features post-convolution. (b) The confusion matrices of the experimental and simulation results on the CIFAR-10 test dataset validate the effectiveness of the method. }
    \label{fig:exp-val-cifar10-2}
\end{figure*}

\begin{figure*}[t]
    \centering
    \includegraphics[width=\textwidth]{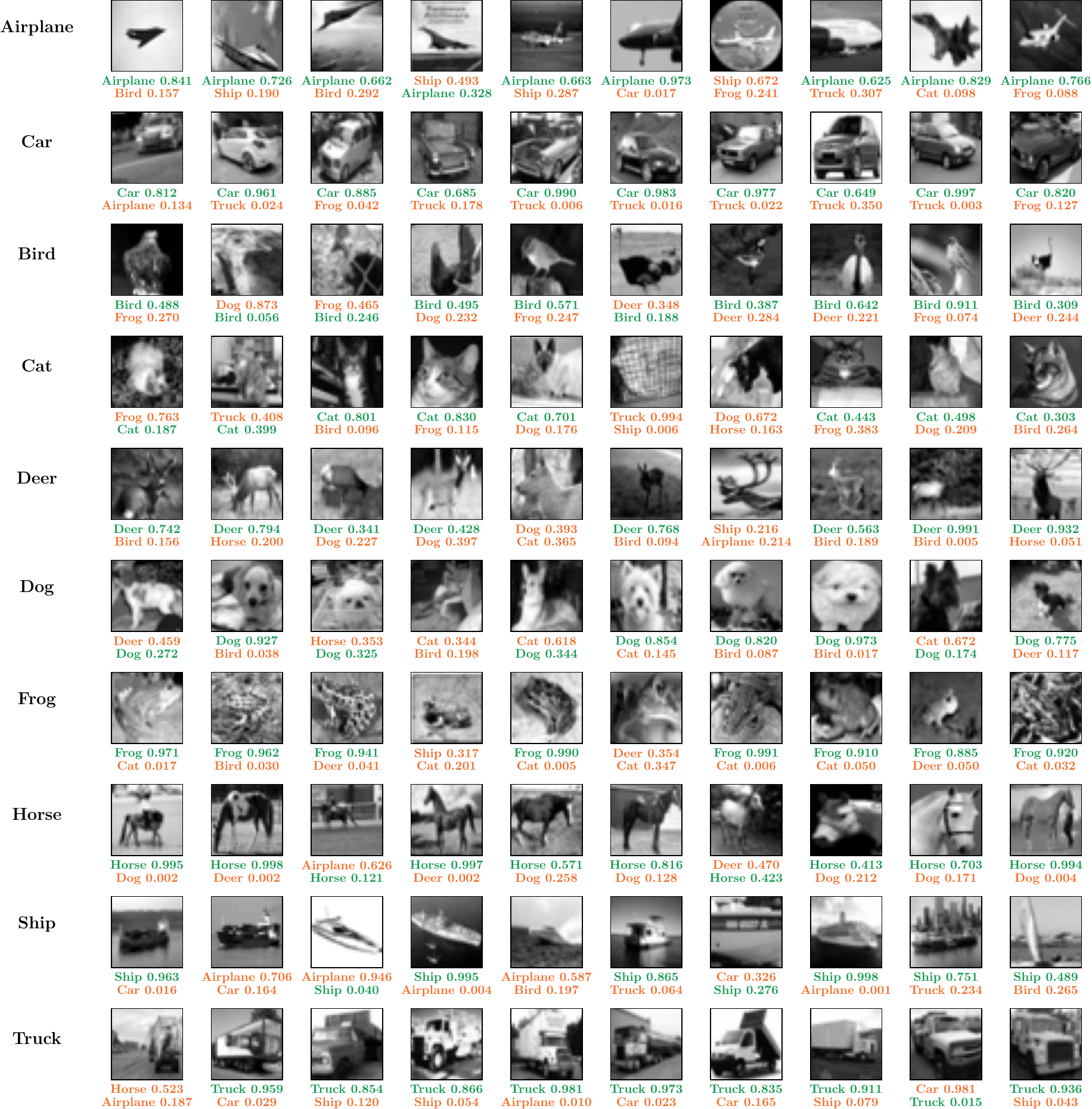}
    \caption{\textbf{Experimental (top-2) classification (probability) results on random samples from CIFAR-10 test set}. \textcolor{ForestGreen}{Green} and \textcolor{Orange}{Orange} colored labels under the images denote the correct and incorrect predictions, respectively. The method accurately predicts the correct class or a visually similar class. See Figure~\note{S15} and \note{S16} for additional examples.}
    \label{fig:cls-vis-res}
\end{figure*}

\begin{figure*}[t]
    \centering
    \includegraphics[width=\textwidth]{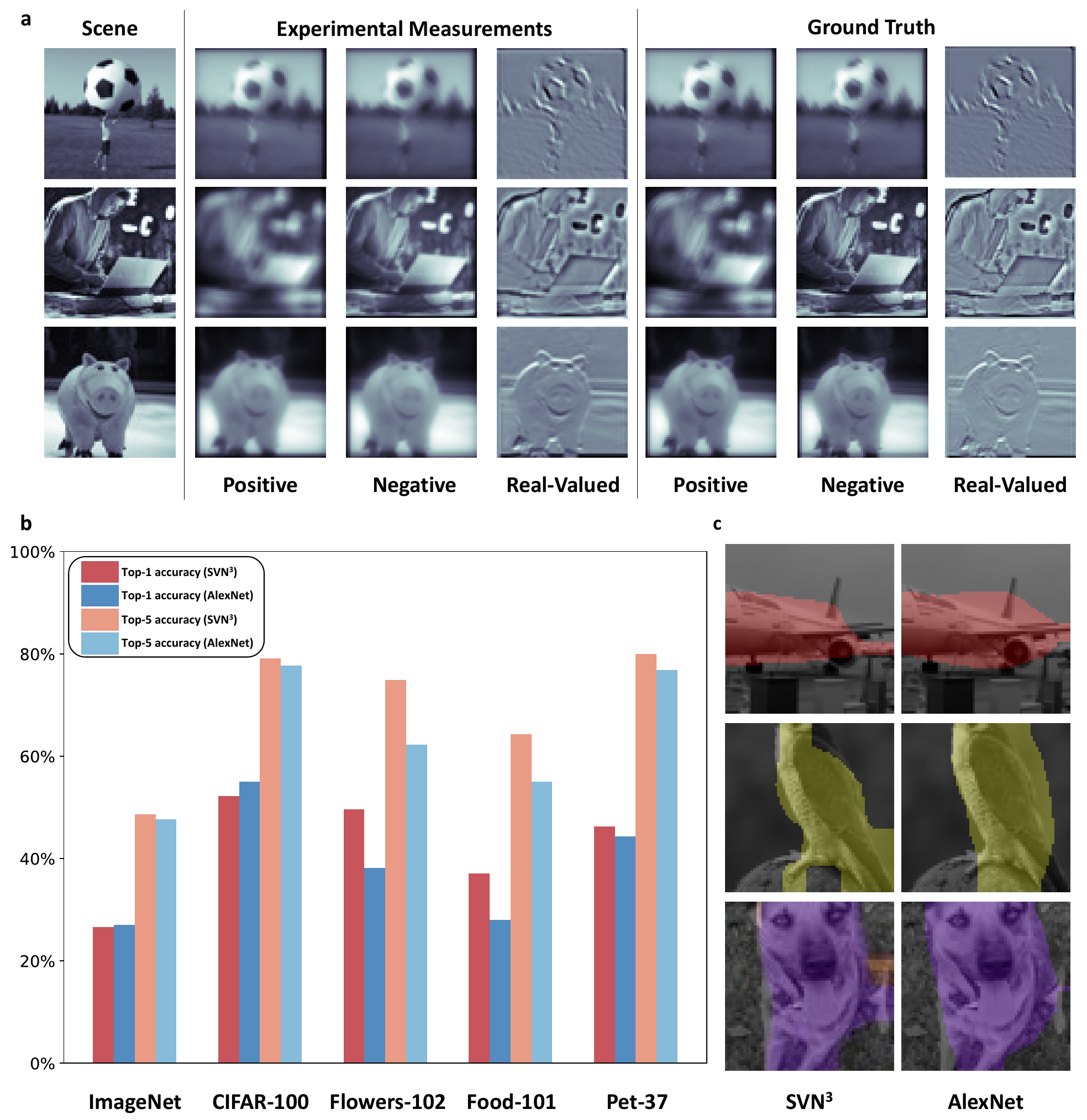}
    \caption{\textbf{Validation of \shortname~as a versatile camera for diverse vision tasks}. (a) Experimentally measured feature maps of \shortname~on the ImageNet dataset. (b) Recognition on ImageNet and other downstream datasets (CIFAR-100, Flowers-102, Food-101, and Pet-37) using the same optical front-end and the transfer-learned electronic decoder. (c) Transfer learning for semantic segmentation on PASCAL VOC dataset. \shortname~again achieves comparable or better performance than the AlexNet-based segmentation network (see Figure~\note{S17} for additional examples).  These findings validate that the proposed camera, with a fixed optical encoder, can generalize to diverse tasks by adapting the electronic backend.} 
    \label{fig:exp-imagenet}
\end{figure*}

\begin{methods}
\subsection{Design and Optimization}
We used PyTorch to design and evaluate our spatially-varying nanophotonic neural network. See Supplementary Note \note{3}, \note{4} and \note{5} for details on the architectural design, 
in-silicon training, and differentiable inverse design of \shortname.

\subsection{Sample Fabrication}
We fabricated the meta-optic on top of a 500 $\mu$m thick double-side polished fused silica wafer. First, a $\SI{800}{nm}$ film of silicon nitride was deposited via plasma-enhanced chemical vapor deposition (PECVD) in a SPTS DeltaX PECVD using Silane and Ammonia as the precursor for a growth at 350 $^{\circ}$C. After growth, the wafer is diced in pieces of $2\times2$ cm and cleaned in a sonicating bath of Acetone, followed by a rinse in Iso-Propyl Alcohol (IPA). Then the sample was shortly cleaned in a O2 plasma using a barrell etcher at 100 W for $\sim$ 15 s. After the cleaning step, we spin-coated the sample with ZEP 520A resist ($\sim$ 400 nm), followed by a layer of a discharging polymer (DisCharge H2O). The arrays of kernels were then written on single chips for the spatially varying and spatially invariant designs via electron beam lithography (EBL) using a JEOL-JBX6300FS with acceleration voltage of 100 kV and 8 nA beam current. After EBL, the sample was rinsed in IPA and developed in amyl acetate for 2 min and rinsed in IPA. To define a hard mask, we evaporated $\SI{65}{nm}$ of alumina using a lab-built e-beam evaporator and a Al2O3 evaporation source. The resist was then lift-off overnight in NMP at 110 $^{\circ}$C and the sample was further cleaned in a brief O2 plasma etch to remove remaining organic residues. We then used inductively-coupled reactive ion etching (Oxford Instruments, PlasmaLab100) with an etch chemistry based on Fluorine to transfer the metasurface layout from the hard mask into the silicon nitride film to a thickness of $\sim$ $\SI{750}{nm}$, whereas the remaining 50 nm of PECVD ensures higher stability of the etched device layer. After fabrication of the device layer, we deposited a metal aperture layer surrounding the metasurfaces to exclude any stray light. These apertures were created through optical direct write lithography (Heidelberg-DWL66) and subsequent deposition of a ~$\SI{150}{nm}$ thick metal film (Cr).

\subsection{Experimental Setup}
We built two experimental setups to characterize the optical performance of metalens array samples, as described in detail in Supplementary Note \note{7}: The first one is used to experimentally measure the PSFs of the metalens array samples. In this setup, a $\SI{520}{nm}$ pigtailed single-mode fiber laser is used to mimic a point light source, and a CMOS sensor is used as the detector to measure the intensity response of a metalens array sample upon the incidence of a point light source positioned at the designed object distance. A microscope objective, together with a relay lens, is used to magnify the PSF measurement on the detector plane. The second setup is used to realize the designed optical system and to measure the image features, as shown in Figure~\hyperref[fig:exp-val-cifar10-1]{2a} and Figure~\hyperref[fig:exp-val-cifar10-1]{2b}. In this setup, the green channel of a smartphone OLED display that is placed at the designated object distance is used as the incoherent light source, and a large-area CMOS sensor is placed at the focal plane of the metalens array device. The smartphone and the sensor are controlled by a computer and synchronized such that when the dataset images are displayed on the smartphone sequentially, the sensor captures the corresponding image features of all the metalens elements in a single shot.

\subsection{Data Availability}
The source images used throughout this work are publicly available. All captured data used to generate the findings in this work will be made public.



\end{methods}

\clearpage

\ifarXiv
    

\section*{Supplementary material}

# Spatially Varying Nanophotonic Neural Networks Supplementary Information

Kaixuan Wei[1]*, Xiao Li[1]*, Johannes Froech[2]*, Praneeth Chakravarthula[1], James Whitehead[2], Ethan Tseng[1], Arka Majumdar[2,3], Felix Heide[1†]

[1]*Princeton University, Department of Computer Science*
[2]*University of Washington, Department of Electrical and Computer Engineering*
[†]*Corresponding author. E-mail: fheide@princeton.edu*

In this supplementary document, we provide additional information, analysis, and experimental results in support of the findings of the main text. In addition to this this document, we have also submitted a supplemental video, demonstrating the proposed method with an experimental prototype, and supplemental code used to make the findings in our manuscript.

**Supplementary Note 1: Additional Conceptual Details of SVN³**

At a high level, SVN³ is a hybrid optoelectronic implementation of a convolutional neural network, which consists of an optical spatially-varying convolutional stem and an extremely efficient classifier electronic backend (see Supplementary Note 3). We instantiate SVN³ with an on-chip nanophotonic array imaging system, which integrates a rectangular array of metalenses, a complementary metal–oxide–semiconductor (CMOS) sensor, and a peripheral micro-controller computing unit. The meta-optic array and CMOS photodetector are responsible for physically realizing the convolutional stem based on a spatially-varying array image formation model, while the microcontroller unit feeds the array of images as multichannel feature maps and executes a super lightweight electronic backend for image classification. The nano-antenna structures of meta-optics are designed for modulating the phase of incident light at a single nominal design wavelength, making it efficient for monochromatic light propagation. The metasurface phase profile is modeled as an unconstrained variable instead of a commonly-used radially symmetric one[1] to fully unleash the optical potential shaping unsymmetric spatially-varying point spread functions (PSFs). Given a scene at a certain distance, the on-axis and off-axis PSFs can then be derived by simulating free-space wave propagations of metasurface-modulated spherical wavefronts from a range of incident angles propagated to the sensor plane[2]. They are engineered to desired shapes and patterns utilizing differentiable inverse design (see Supplementary Note 5). As such, the resulting images of metalenses resemble shift-variant convolutions of a scene and large spatially-varying kernels. These are spatially multiplexed in the sensor plane before forming multichannel features fed into the electronic backend for final outcomes. Since the computationally expensive operation, *i.e.*, the large-kernel spatially-varying (LKSV) convolution, is offloaded into the high-speed and low-power optics, SVN³ exhibits extraordinary efficiency without sacrificing the inference capability. Moreover, SVN³ can serve as an ultra-compact computational camera that does not require

1

any conventional and costly image signal processing (ISP) routines (*e.g.*, demosaicking, denoising, and tone mapping), rendering it a promising energy-economic alternative for applications that entail always-on inference functionalities.

**Supplementary Note 2: Comparative Analysis of Existing ONNs**

Benchmarking existing optical computing architectures poses challenges due to the substantial differences in their implementation. Nonetheless, comparing their best-reported classification results on established datasets and examining high-level system features offers valuable insights into the landscape of optical computing.

As our work focuses on free-space optical computing, we restrict the comparison to existing ONNs that utilize free-space wave propagation, omitting considerations regarding integrated photonic circuitry. Table S1 summarizes blind testing accuracies achieved by diverse architectures (all-electronic, all-optical, and hybrid optoelectronic) on commonly used image classification datasets. Nearly all architectures demonstrate saturated performance on the MNIST dataset for handwritten digit recognition.

However, challenges arise when tackling the more complex task of generic object recognition on the CIFAR-10 dataset. Even the most high-performing optical architecture, leveraging dozens of parallel diffractive optical networks, achieved only 62.13% blind test accuracy on the CIFAR-10 test set. This accuracy falls short compared to the classic electronic network LeNet, developed three decades ago. In contrast, our SVN$^3$ achieves a notable 73.80% blind test accuracy on the CIFAR-10 dataset. By fully harnessing the capabilities of free-space optical computing, our system surpasses even the first modern electronic neural network—AlexNet on CIFAR-10. Furthermore, with a slightly expanded electronic backend, our hybrid system demonstrates superior performance compared to AlexNet, even when dealing with the challenging 1000-class ImageNet dataset, one of the most broadly used image classification datasets.

Table S3 provides a comparison of various crucial features (e.g., percentage of optical compute, light source type, and system form factor) across different types of optical neural networks. In comparison to all-optical neural networks, SVN$^3$ is capable of processing incoherent light as input, turning it into a *optical computer in the camera rather than solely an optical accelerator*. In contrast to hybrid optoelectronic networks, SVN$^3$ features a compact form factor easily integrable with off-the-shelf CMOS photosensors, owing to the small footprint of metasurfaces. As such, SVN$^3$ provides a unique perspective in the spectrum of optical computing, while elevating performance to a new paradigm.

Table S1: **Benchmark image classification performance of existing ONNs.** We report the results (classification accuracy) obtained by numerical simulation on four widely used image recognition datasets. For ImageNet, the top-5 accuracy is reported on the resized $64 \times 64$ ImageNet dataset with 1000 categories. For other datasets, we report the standard top-1 blind testing accuracy. The performance of the classic electronic neural networks, LeNet and AlexNet is provided for reference. Entries as "-" indicate that the results are not applicable.

| TYPE | METHODS | RECOGNITION ACCURACY | | | |
|---|---|---|---|---|---|
| | | MNIST | FASHIONMNIST | CIFAR-10 | IMAGENET |
| ALL-ELECTRONIC | LeNet[3] | 99.20 % | 90.30 % | 66.43 % | - |
| | AlexNet[4] | - | - | 72.64 % | 47.68 % |
| ALL-OPTICAL | D$^2$NN[5] | 97.51 % | 89.85 % | 45.20 % | - |
| | Class-specific D$^2$NN[6] | 98.52 % | 91.48 % | 50.82 % | - |
| | D$^2$NN Ensemble[7] | - | - | 62.13 % | - |
| | Metasurface D$^2$NN[8] | 93.80 % | 87.50 % | - | - |
| OPTO-ELECTRONIC | 4F System[9] | - | - | 51.00 % | - |
| | DPU[10] | 99.00 % | 90.20 % | - | - |
| | LOEN[11] | 97.21 % | - | - | - |
| | Meta-optic Accelerator[12] | 94.70 % | - | - | - |
| | All-analog photoelectronic chip[13] | 98.00 % | 89.00 % | - | - |
| | Ours | **99.20 %** | **91.55 %** | **73.80 %** | **51.28 %** |

3

Table S2: **Ablation and Complexity Experiments for the Proposed Design Variant (LKSV)**: We report the number of multiply–accumulate (MAC) operations, the inference runtime (TIME) and the number of parameters (# of PARAMS) for the optical and electronic parts of different models. The runtime is reported for a NVIDIA GTX 1080Ti GPU; note that the true optical runtime is only a function of the optical path length and is $\approx 3$ ps in our implementation. We report here the runtime numbers of the optical model component when implemented electronically for reference, *i.e.*, higher is better. The labels SK/LK refer to small and large kernel respectively; SI/SV indicate spatially invariant and spatially varying respectively. * denotes large kernel and/or spatially varying convolutional layer learned *without* low-dimensional reparameterization. ACC denotes the blind testing accuracy on the CIFAR-10 dataset. The result of AlexNet is also listed as a reference.

| MODEL | ACC | MAC | | TIME | | # of PARAMS | |
|---|---|---|---|---|---|---|---|
| | | OPTICAL ↑ | ELECTRONIC ↓ | OPTICAL ↑ | ELECTRONIC ↓ | OPTICAL ↑ | ELECTRONIC ↓ |
| AlexNet | 72.64 % | - | 174.85 M | - | 70.58 us | - | 57.03 M |
| | | - | 100 % | - | 100 % | - | 100 % |
| SKSI | 65.45 % | 243.20 K | 125.30 K | 0.41 us | 2.64 us | 0.25 K | 2.03 K |
| | | 66.00 % | 34.00 % | 13.56 % | 86.44 % | 10.94 % | 89.06 % |
| LKSI* | 67.13 % | 5.78 M | 125.30 K | 3.54 us | 2.64 us | 5.65 K | 2.03 K |
| | | 97.88 % | 2.12 % | 57.25 % | 42.75 % | 73.52 % | 26.48 % |
| LKSI | 71.64 % | 5.78 M | 125.30 K | 3.54 us | 2.64 us | 5.65 K | 2.03 K |
| | | 97.88 % | 2.12 % | 57.25 % | 42.75 % | 73.52 % | 26.48 % |
| SKSV* | 57.69 % | 1.41 M | 125.30 K | 27.72 us | 2.64 us | 6.52 K | 2.03 K |
| | | 91.83 % | 8.17 % | 91.30 % | 8.70 % | 76.21 % | 23.79 % |
| SKSV | 71.14 % | 1.41 M | 125.30 K | 27.72 us | 2.64 us | 6.52 K | 2.03 K |
| | | 91.83 % | 8.17 % | 91.30 % | 8.70 % | 76.21 % | 23.79 % |
| LKSV | **73.80** % | 34.59 M | 125.30 K | 302.66 us | 2.64 us | 38.92 K | 2.03 K |
| | | **99.64** % | **0.36** % | **99.14** % | **0.86** % | **95.03** % | **4.97** % |

**Table S3: Footprint and Optical Computation in Existing ONN Methods**: We show 1) the percentage of the optical compute within the whole system in terms of the multiply–accumulate (MAC) operations; 2) the working light source requirement; and 3) the system's form factor (height × width × depth, estimated if not available). We list the results of the most performant architectures reported in literature, *e.g.*, the 16-kernel LEON[11] with 3 fully-connected (FC) layers, and the 2-FC version of the meta-optic accelerator[12].

| TYPE | METHODS | FEATURES | | |
| --- | --- | --- | --- | --- |
| | | OPTICAL COMPUTE | LIGHT SOURCE | FORM FACTOR (mm) |
| ALL-OPTICAL | $D^2NN$[5] | Almost 100%, except the analog-to-digital conversion in sensor | Coherent | (80, 80, 120) |
| | Class-specific $D^2NN$[6] | | | (80, 80, 120) ×20 |
| | $D^2NN$ Ensemble[7] | | | (80, 80, 120) ×30 |
| | Metasurface $D^2NN$[8] | | | (0.1, 0.1, 0.1) |
| OPTO-ELECTRONIC | 4f System[9] | 95.45 % | Incoherent | (10, 10, 600) |
| | DPU[10] | 99.99 % | Coherent | (20, 12, 600) |
| | LOEN[11] | 2.70 % | Incoherent | (25, 25, 1) |
| | Meta-optic Accelerator[12] | 48.69 % | Incoherent | (2.4, 2.4, 200) |
| | Ours | **99.88 %** | Incoherent | (6, 9, 4) |

**Supplementary Note 3: Architectural Design of the LKSV-CNN**

**Design Overview.** To fully harness the capabilities of optical computing, we develop a novel convolutional neural network (CNN) that is optimized for hybrid optoelectronic deployment. At the core of our approach is a convolutional stem design (the first layer of the CNN) which is tailored for photonic implementation and has superior representation capabilities, leveraging the innate characteristics of light-matter interaction. Our approach is motivated by a rethinking of the inherent nature of optical aberration, which is conventionally considered a destructive property of imaging systems that blurs and/or distorts resulting images. For centuries, optical engineering has worked with the goal to correct for aberration, designing near-perfect imaging systems[14]. We depart from this approach and treat aberration as a computational tool rather than a problem to be solved. With the design freedom enabled by metasurfaces[15,16], we precisely engineer metalenses with desired spatially nonuniform aberrations over large spatial support. These metalenses enable optical encoding of a multichannel LKSV convolutional layer, which we use used to build a feature embedding for high-level reasoning tasks. To design these convolutional kernels, we first create an LKSV-CNN as an electronic proxy model and train it in silicon for image recognition. We then use the learned kernels to optimize nanophotonic structures via differentiable inverse design, resulting in target point spread functions.

As shown in Figure S1, the proposed LKSV-CNN mainly consists of three parameterized learnable ingredients — an LKSV convolutional stem $f^{(s)}$, a depthwise separable convolutional backbone[17,18] $f^{(b)}$ and a fully-connected classification head $f^{(h)}$, each of which is separated by a rectified linear unit (ReLU) $\phi$ and/or a pooling function[19] $\psi$. The architecture $f$ is defined as

$$f = f^{(h)} \circ \psi \circ f^{(b)} \circ \psi \circ \phi \circ f^{(s)}, \tag{S1}$$

that is a a a mapping $\hat{y} = f(x; \theta)$ from an input image $x$ to a predicted category $\hat{y}$, where $\circ$ denotes the function composition operator, $\theta$ indicate the associated parameters such as convolutional kernels that are learned end-to-end. Building the LKSV convolutional stem $f^{(s)}$, at first glance, seems only to require simply enlarging the kernels and untying the weight-sharing mechanism of a convolutional layer. However, as demonstrated in Table S2 (also see Supplementary Note 8), Such a naïve approach usually results in rather poor performance in practice, suffering from both higher computational cost and lower accuracy. This is nevertheless **not** surprising as rules of thumb found in deep learning literatures: 1) modern deep learning practices seldom use convolutional kernels of size beyond 3×3 following the well-established deep architectures[20,21] and 2) the spatially-varying convolutional layer (*a.k.a.* locally-connected layer in machine learning terminology) is empirically inferior to its weight-sharing counterpart possibly owing to the lack of spatial invariance, a highly successful inductive bias for visual reasoning[22–26].

With these negative results in mind, we identify the bad performance of LKSV convolutional layer is not attributed to its representation capability but to the optimization strategy. Mathematically, functions expressible by small-kernel spatially-invariant (SKSI) convolutional layers are a subset of those expressible by LKSV layers, which indicates the LKSV layer is ideally at least

as good as the SKSI counterpart in terms of representational capacity. The terrible performance of LKSV layer, therefore, results from the highly non-convex optimization landscape of neural networks[27–29], which is especially difficult for layers with a massive number of learnable parameters. To circumvent the optimization difficulty, we propose low-dimensional reparameterization techniques coined large kernel factorization and low-rank spatially-varying reparameterization that reformulate the LKSV convolution into a low-dimensional one with a sequence of small-kernel basis. This strategy significantly stabilizes the gradient flow during network training with a stochastic gradient-based optimizer, allowing us to unlock the modeling power of the proposed LKSV design. After training, the reparameterized layer can be equivalently transformed back into the original structure as if it was learned as usual.

**Large Kernel Factorization.** Before diving into the spatially-varying convolution, we first show how a large-kernel convolution can be structurally reparameterized as a sequence of small-kernel ones using large kernel factorization. To avoid ambiguity, we refer to *convolution* in optics/signal processing context, which is the transposed convolution under deep learning convention. The readers should not be confused by these two kinds of convolution operators as they are mutually equivalent with appropriate zero padding[30]. Given a (transposed) convolutional layer with $C$ input channels, $D$ output channels and kernel size $K \times K$, its kernel is a 4th-order tensor $\mathbf{K} \in \mathbb{R}^{C \times D \times K \times K}$. This computing layer feeds a $C$-channel input $\mathbf{I} \in \mathbb{R}^{C \times H \times W}$ and outputs a $D$-channel feature map $\mathbf{O} \in \mathbb{R}^{D \times H' \times W'}$, *i.e.*, $\mathbf{O} = \mathbf{I} * \mathbf{K}$, where $*$ denotes the convolution operator and the output shape $H'$ and $W'$ are determined by $K$, padding, and stride configurations. We focus on optic-realizable cases with stride one and without padding. The value at $(h, w)$-entry on the $d$-th output channel is thus given by

$$\mathbf{O}_{d,h,w} = (\mathbf{I} * \mathbf{K})_{d,h,w} = \sum_{c=1}^{C} \sum_{p=1}^{K} \sum_{q=1}^{K} \mathbf{K}_{c,d,p,q} \mathbf{I}_{c,h-p+1,w-q+1} \tag{S2}$$

This equation also represents the (multichannel) discrete analog of the image formation process, where the kernel $\mathbf{K}$ is analog to a point spread function, the response of a focused optical imaging system to a point source. Applying multiple convolutions sequentially with a bank of kernels $\{\mathbf{K}_1, \mathbf{K}_2, \cdots, \mathbf{K}_N\}$, as illustrated in Figure S1b, can be reduced to a single convolution operation with a large kernel $\mathbf{K}_L$, thanks to the algebraic properties of the convolution operator[30], Specifically, the convolution operator satisfies *associativity*,

$$\mathbf{I} * \mathbf{K}_1 * \mathbf{K}_2 = \mathbf{I} * (\mathbf{K}_1 * \mathbf{K}_2) \tag{S3}$$

Applying Equation (S3) iteratively, we thereby have

$$\mathbf{I} * \mathbf{K}_1 * \mathbf{K}_2 * \cdots * \mathbf{K}_N = \mathbf{I} * \mathbf{K}_L \tag{S4}$$

where $\mathbf{K}_L = (\mathbf{K}_1 * \mathbf{K}_2 * \cdots * \mathbf{K}_N)$, implying the integrated large kernel $\mathbf{K}_L$ can be computed by convolving all sub-kernels sequentially. The Equation (S4) lays the foundation of the proposed

7

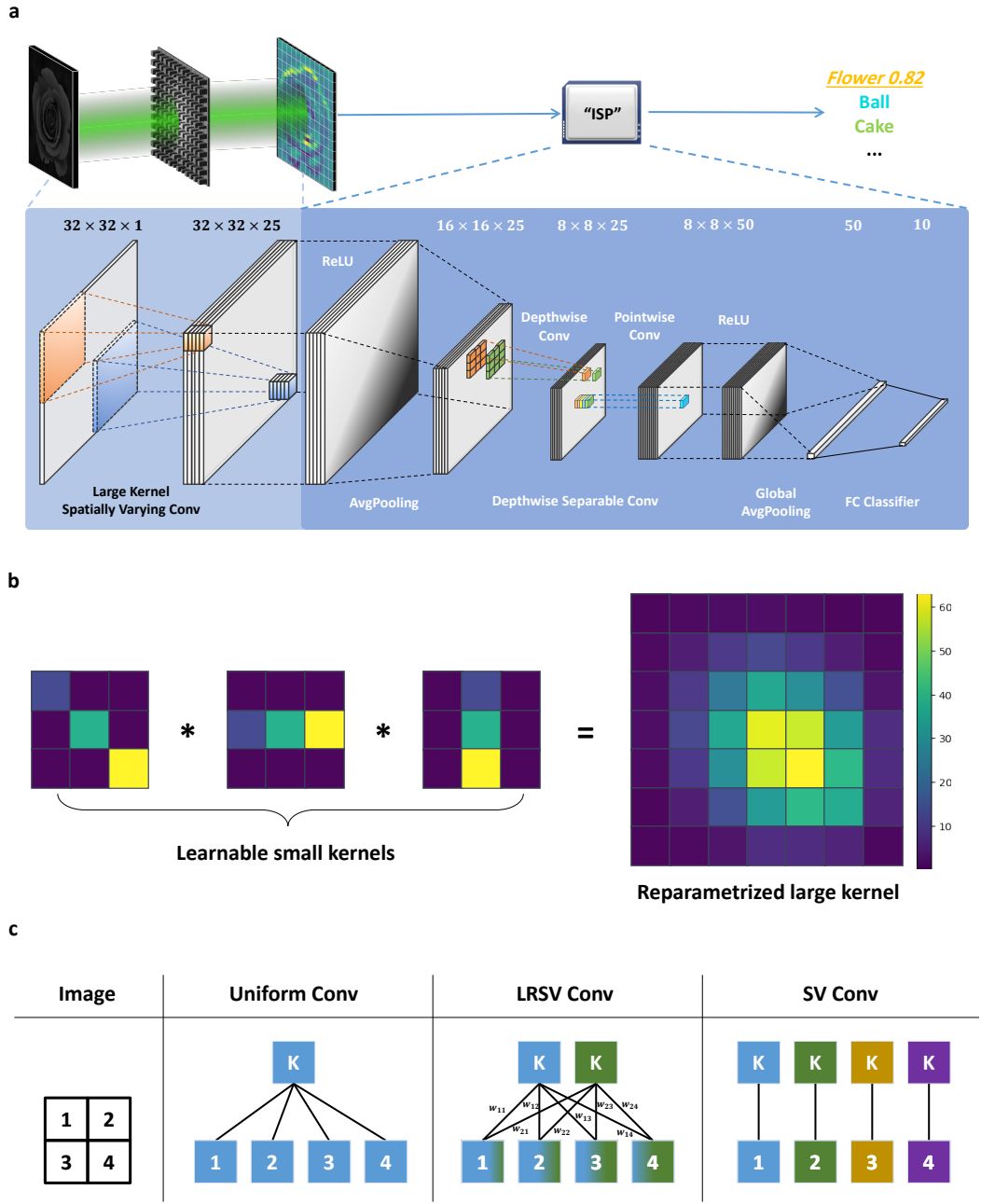

**Figure S1: Architectural overview of LKSV-CNN.** (a) The overall architecture of the proposed LKSV-CNN, where the optical frontend performs a multichannel large-kernel spatially varying convolution, and the resulting features are then processed through a lightweight electronic backend to obtain the classification results. (b) Example illustration of the principle of large-kernel factorization. The large kernel can be represented as a convolution of smaller basis kernels (c) Schematic comparison of different types of convolution (spatially uniform convolution; low-rank spatially varying convolution and naïve spatially varying convolution).

8

large kernel factorization technique — instead of directly optimizing a large-kernel convolutional layer, we reparameterize it by a cascade of multiple layers with small ($3 \times 3$) kernels, which facilitates learning effectively using stochastic gradient descent. For example, to learn a large kernel with size $15 \times 15$, we could first factorize it into seven cascaded $3 \times 3$ kernels during training, and then recombine them into the large kernel after finishing the training. The large kernel factorization introduces dependencies/correlations onto the kernel parameters, which therefore imposes a low-dimensional manifold on kernel parameter space, effectively acting as a regularizer. Learning on this regularized parameter space enables stochastic optimization towards robust critical points that fit and generalize well.

**Low-Rank Spatially-Varying Reparameterization.** The structural design of CNN is often compared to the primate visual system[31] but the visual system has no direct mechanism to share weights across space — Neurons comprising retinotopic maps have selectivity properties that vary with their position within the map[32–35]. This naturally drives the design of spatially varying convolutional layers that depart from spatial invariance. However, naïvely employing spatially varying kernels empirically results in performance degradation compared to their spatially-invariant counterparts, revealing a similar optimization dilemma encountered when using large kernels. We again leverage a structural reparameterization technique namely low-rank spatially-varying reparameterization to address this problem. Specifically, we reformulate the spatially-varying convolution using a linear combination of a basis set of convolutional kernels with spatially-variant combining weights. As illustrated in Figure S1c, instead of using the same kernel (convolutional layer) or a separate independent kernel (locally-connected layer) for each spatial location, our proposed low-rank spatially-varying convolutional layers use a separate kernel generated from combining a shared kernel basis set for each spatial location. In detail, for each input spatial location $(h, w)$, we construct a kernel $\mathbf{K}^{(h,w)}$ that is a linear combination of the members of a basis set, *i.e.*,

$$\mathbf{K}^{(h,w)} = \sum_{r=1}^{R} \boldsymbol{\omega}_{r,h,w} \mathbf{K}_{basis}^{(r)}, \tag{S5}$$

where $R$ is the number of elements in the basis set, Here, $\boldsymbol{\omega} \in \mathbb{R}^{R \times H \times W}$ are the corresponding spatial combining weights, $\mathbf{K}_{basis}^{(r)}$ denotes the $r$-th (large) kernel of the basis set, which is reparameterized by a stack of small ($3 \times 3$) kernels as $\mathbf{K}_{basis}^{(r)} = (\mathbf{K}_1^{(r)} * \mathbf{K}_2^{(r)} * \cdots * \mathbf{K}_N^{(r)})$ using the aforementioned large kernel factorization. Notice the formulation (S5) resembles a low-rank factorization of a spatially-varying kernel with rank $R$, suggesting the "low-rank" term in its name. We parameterize the spatial weights $\boldsymbol{\omega}$ as subsampled pixel-wise free variables $\tilde{\boldsymbol{\omega}} \in \mathbb{R}^{R \times \frac{H}{s} \times \frac{W}{s}}$ where $s$ is the scale factor (set as 4 in our experiment). The weights $\boldsymbol{\omega}$ are then computed via bicubic upsampling followed by a softmax function for normalization, that is

$$\boldsymbol{\omega}_{r,h,w} = \frac{\exp\left(\tilde{\boldsymbol{\omega}}_{r,h,w}\right)}{\exp\left(\sum_{r=1}^{R} \tilde{\boldsymbol{\omega}}_{r,h,w}\right)} \tag{S6}$$

9

**Algorithm S1** LKSV Convolutional Layer

**Trainable Parameters:**
    // A set of factorized small kernels that induces large-kernel basis set $\{\mathbf{K}_{basis}^{(r)}\}$ in Equation (S5)
    $\{\mathbf{K}_i^{(r)} \in \mathbb{R}^{C \times D \times 3 \times 3} \mid i = 1, ..., N; \ r = 1, ..., R\}$
    $\tilde{\boldsymbol{\omega}} \in \mathbb{R}^{R \times \frac{H}{s} \times \frac{W}{s}}$                                                             ▷ subsampled spatial combining weights

1: **procedure** LKSV-CONV($\mathbf{I}$)                                          ▷ $\mathbf{I} \in \mathbb{R}^{C \times H \times W}$ is the input image feature
2:     Compute spatial combining weights $\boldsymbol{\omega} \in \mathbb{R}^{R \times H \times W}$ via Equation (S6).
3:     Form the large-kernel basis $\mathbf{K}_{basis}^{(r)} = (\mathbf{K}_1^{(r)} * \mathbf{K}_2^{(r)} * \cdots * \mathbf{K}_N^{(r)}) \quad \forall \ r \in \{1, ..., R\}$
4:     Initialize $\mathbf{O} = \mathbf{0} \in \mathbb{R}^{D \times H' \times W'}$
5:     **for** $r$ = 1 to $R$ **do**
6:         $\hat{\mathbf{I}} = \boldsymbol{\omega}_{r,:,:} \otimes \mathbf{I}$               ▷ $\otimes$ denotes elementwise multiplication with broadcasting
7:         $\mathbf{O} = \mathbf{O} + (\hat{\mathbf{I}} * \mathbf{K}_{basis}^{(r)})$
8:     **end for**
9:     **return** $\mathbf{O}$
10: **end procedure**

During the training phase, we implement the proposed LKSV convolutional layer $f^{(s)}$ with convolution and elementwise multiplication operations, as in Algorithm S1, rather than forming the equivalent locally-connected layer. The kernel in the basis set and the combining weights can all be learned end-to-end. See Figure S2 for a visualization of learned kernel basis (with rank $R = 6$) and combining weights $\boldsymbol{\omega}$. After training, the resulting spatially-varying kernels in each spatial location can be computed by the weighted combination of kernel basis (See Figure S3 for example spatially-varying kernel visualization).

**Extremely Lightweight Electronic Backend.** Above, we have presented the LKSV convolutional stem that is ultimately offloaded into a linear optical frontend. To complete the proposed hybrid optoelectronic neural network, we further design a simple yet effective electronic backend that introduces nonlinearity with minimal complexity. This electronic backend is composed of a depthwise separable convolution (DSC) backbone $f^{(b)}$, a fully-connected classification head $f^{(h)}$, as well as associated auxiliary functions (See Equation (S1)), all operating on low-resolution spaces with marginal computational overhead. This electronic component of ultra-low complexity, and, akin to the ISP of a camera sensor, can be performed during readout.

    DSC is a simplified convolutional layer decoupling the cross-channels and spatial correlations modeling with low computational complexity and memory usage[18]. DSC encompasses two consecutive layers: a depthwise convolution layer, *i.e.*, a spatial convolution performed independently over each channel of an input $\mathbf{I} \in \mathbb{R}^{C \times H \times W}$, followed by a pointwise convolution layer, *i.e.*, a $1 \times 1$ convolution, projecting the channels $\tilde{\mathbf{O}} \in \mathbb{R}^{C \times H' \times W'}$ output by the depthwise convolution

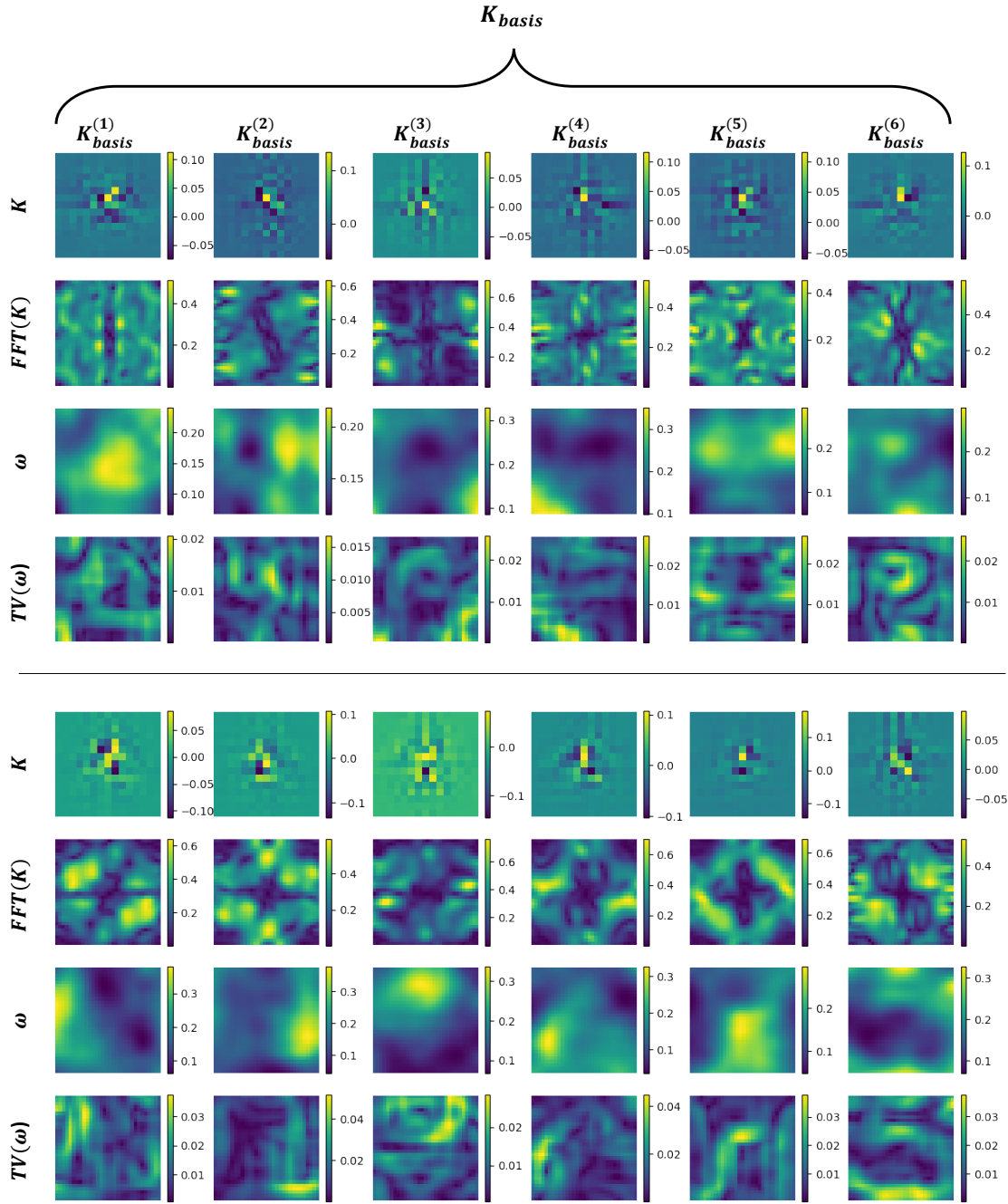

**Figure S2: Two example kernels learned via low-rank spatially-varying reparameterization.** At each spatial location, the kernel value is determined by the weighted average of six basis kernels $K_{basis}^{(i)}$, where the spatial combining weight $\omega$ and its corresponding total variation map $TV(\omega)$ are displayed in the third and fourth rows respectively. The centered Fourier spectrum of each basis kernel $FFT(K)$ is shown alongside the kernel.

11

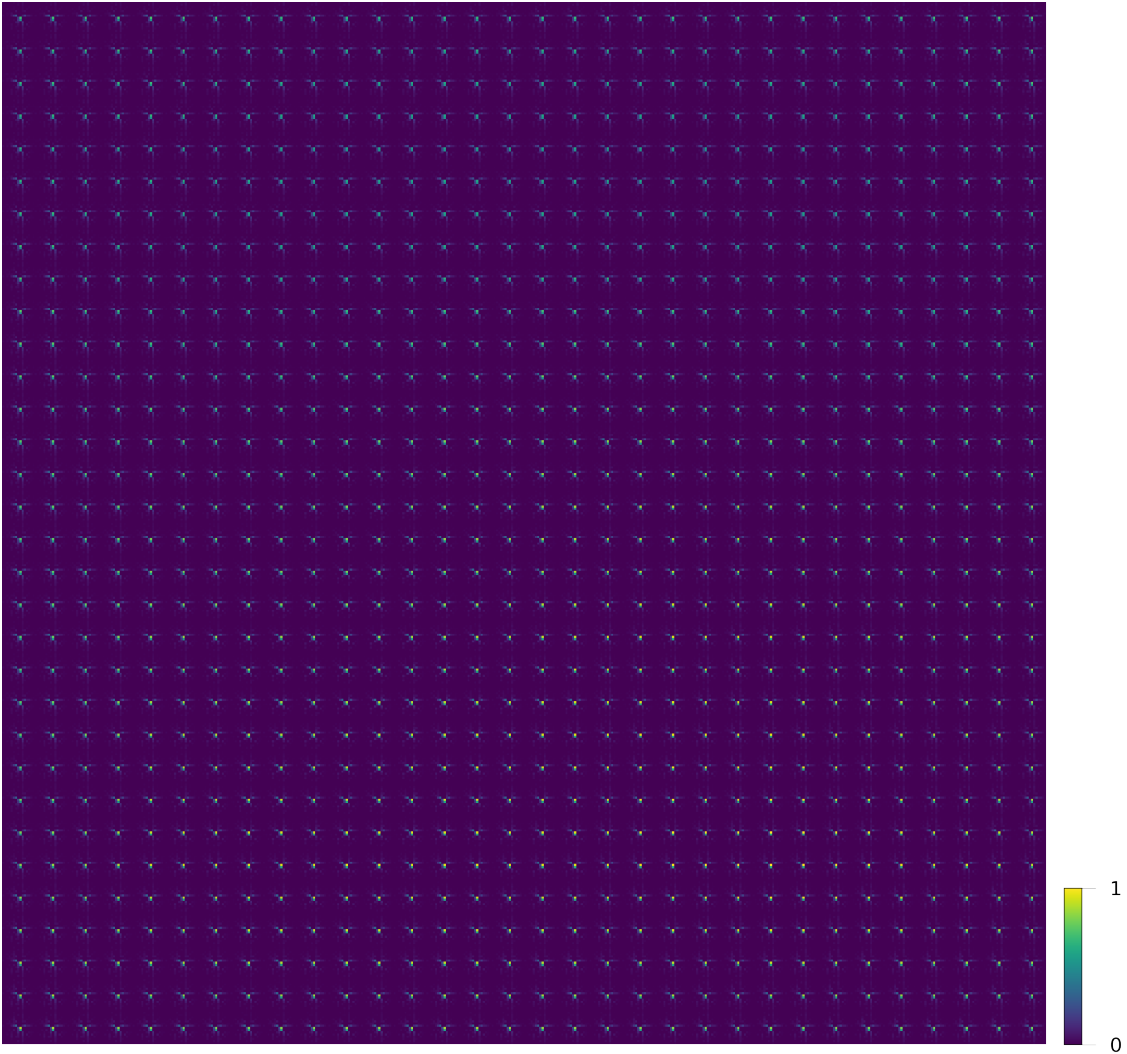

Figure S3: **Example spatially-varying kernels #1.** We visualize the kernels (PSFs) over $32 \times 32$ incident angles. This illustration serves as a high-resolution copy of Figure 1d in the main paper. We refer the reader to the electronic version to zoom in for details.

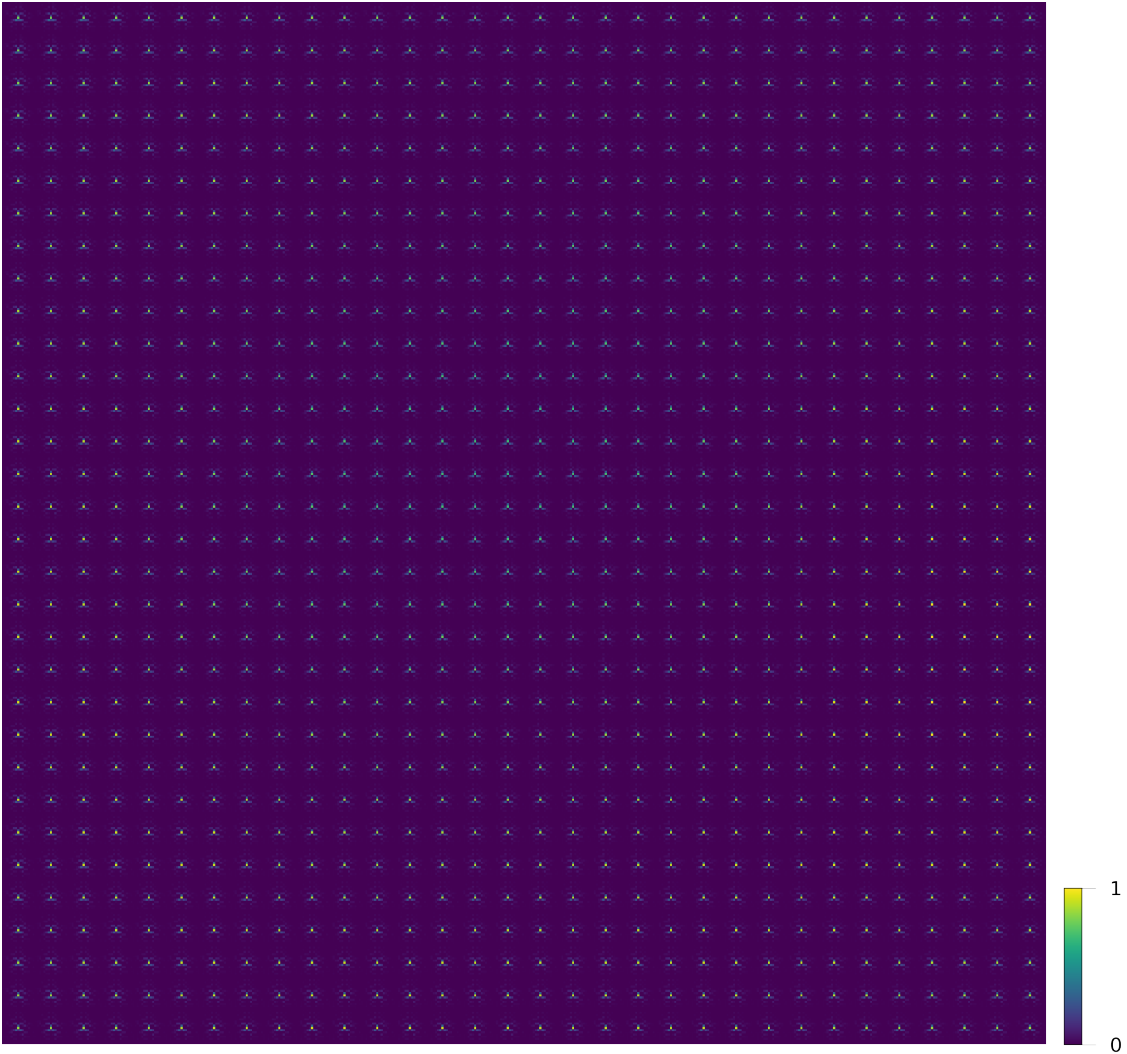

**Figure S3: Example spatially-varying kernels #2.** We visualize the kernel (PSFs) over $32 \times 32$ incident angles. This illustration serves as a high-resolution copy of Figure 1d in the main paper. We refer the reader to the electronic version to zoom in for details.

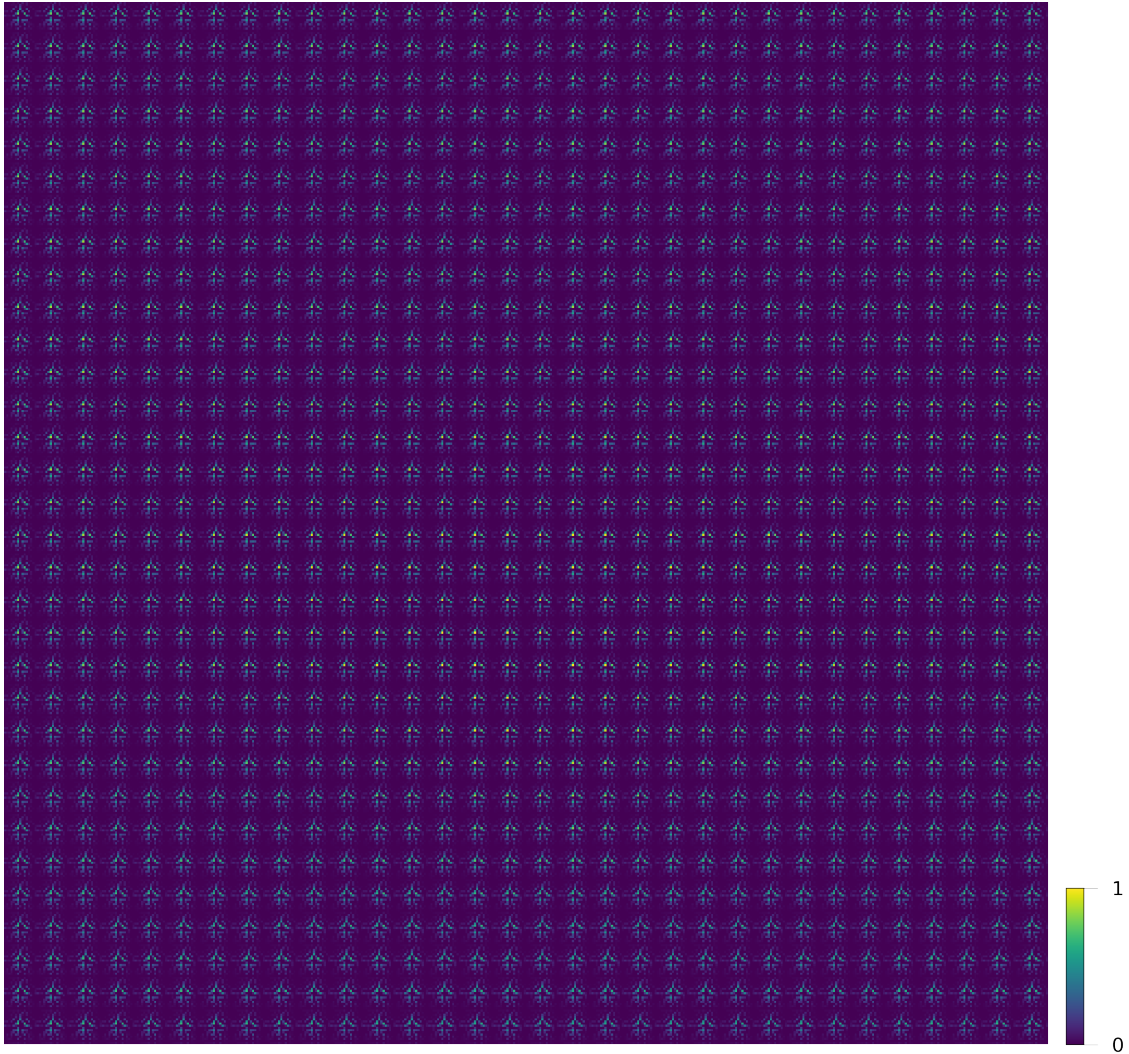

**Figure S3: Example spatially-varying kernels #3.** We visualize the kernel (PSFs) over $32 \times 32$ incident angles. This illustration serves as a high-resolution copy of Figure 1d in the main paper. We refer the reader to the electronic version to zoom in for details.

onto a new channel space $\mathbf{O} \in \mathbb{R}^{C' \times H' \times W'}$, which formally reads,

$$\tilde{\mathbf{O}}_{c,h',w'} = (\mathbf{I} \star \mathbf{K}^{(d)})_{c,h',w'} = \sum_{p=1}^{K} \sum_{q=1}^{K} \mathbf{K}^{(d)}_{c,p,q} \mathbf{I}_{c,(h'-1)\times s+p,(w'-1)\times s+q}$$

$$\mathbf{O}_{c',h',w'} = (\tilde{\mathbf{O}} \odot \mathbf{K}^{(p)})_{c',h',w'} = \sum_{c=1}^{C} \mathbf{K}^{(p)}_{c',c} \tilde{\mathbf{O}}_{c,h',w'} \tag{S7}$$

where $\star$ and $\odot$ denote the depthwise and pointwise convolution parameterized by $\mathbf{K}^{(d)} \in \mathbb{R}^{C \times K \times K}$, $\mathbf{K}^{(p)} \in \mathbb{R}^{C' \times C}$ respectively. Here, $s$ is the stride of the depthwise convolution, which we set to 2 for downsampling in the LKSV-CNN. The DSC backbone $f^{(b)}$ consists of one or several cascaded DSC layers (separated by ReLU nonlinearity), depending on the complexity of the task in hand (*e.g.*, for CIFAR-10 image classification, and we use only one DSC layer to minimize electronic computing burden while maintaining high accuracy). Once we obtain the features extracted by $f^{(b)}$, a global average pooling $\psi^{(g)}$ is applied followed by a fully-connected classifier head $f^{(h)}$ to generate the final category scores as in Equation (S1), where the $f^{(h)}(\boldsymbol{x}; \mathbf{W}, \boldsymbol{b})$ is a linear transformation parameterized by weight matrix $\mathbf{W}$ and bias vector $\boldsymbol{b}$, *i.e.*, $\hat{\boldsymbol{y}} = \mathbf{W}^T \boldsymbol{x} + \boldsymbol{b}$.

## Supplementary Note 4: Neural Network Training and Finetuning

After defining the LKSV-CNN, training is conducted in-silicon to learn convolutional kernels of SVN[3] as and target PSFs via inverse design.

**Training Loss** The main training loss function for image classification is the standard cross-entropy loss, *i.e.*,

$$\mathcal{L}_{ce}(\boldsymbol{y}, \hat{\boldsymbol{y}}) = -\frac{1}{B} \sum_{n=1}^{B} \sum_{i=1}^{C} y_{n,i} \log \hat{y}_{n,i} \tag{S8}$$

where $C$ is the number of classes, $B$ is the batch size, $y_{n,i}$ and $\hat{y}_{n,i}$ denote the true and predicted labels for the $n$-th example in the batch, respectively.

To facilitate the optical realization of LKSV convolution, we further introduce two regularization loss functions including a second-order isotropic total variation loss and a high-pass loss that encourage the smoothness of the spatial combining weights and penalize kernels with high-pass characteristics. To be specific, the second-order isotropic total variation loss $\mathcal{L}_{tv}$ is imposed on the spatial combining weights $\boldsymbol{\omega} \in \mathbb{R}^{C \times H \times W}$ (See Equation (S5)) as

$$\mathcal{L}_{tv}(\boldsymbol{\omega}) = \bar{\boldsymbol{\omega}}_{tv} + \sigma(\boldsymbol{\omega}_{tv}), \tag{S9}$$

where $\bar{\boldsymbol{\omega}}_{tv}$ and $\sigma(\boldsymbol{\omega}_{tv})$ are the mean value and the standard deviation of $\boldsymbol{\omega}_{tv}$, $\boldsymbol{\omega}_{tv} \in \mathbb{R}^{H \times W}$ is the isotropic total variation map of the $\boldsymbol{\omega}$ that can be computed as

$$\boldsymbol{\omega}_{tv} = \sum_{c=1}^{C} \sqrt{(\boldsymbol{\omega}_{c,:,:\text{end}-1} - \boldsymbol{\omega}_{c,:,2:})^2 + (\boldsymbol{\omega}_{c,:\text{end}-1,:} - \boldsymbol{\omega}_{c,2:,:})^2}. \tag{S10}$$

The : in subscript above indicates a slice operation along an axis, $(\boldsymbol{\omega}_{c,:,:\text{end}-1} - \boldsymbol{\omega}_{c,:,2:})$ and $(\boldsymbol{\omega}_{c,:\text{end}-1,:} - \boldsymbol{\omega}_{c,2:,:})$, therefore, represent horizontal and vertical finite differences, respectively. Minimizing the loss function $\mathcal{L}_{tv}$ ensures the smoothness of the varying weights across the spatial domain.

Besides, a high-pass loss is proposed to regularize the frequency response of kernels in the convolutional basis set $\mathbf{K}_{basis}^{(r)} \in \mathbb{R}^{D \times K \times K}$ (we omit the input channel dimension $C$ as $C = 1$) since the optical convolution is by nature a low-pass filter[2] that has limited expressiveness for a high-pass one. The high-pass loss can be characterized as

$$\mathcal{L}_{hp}(\mathbf{K}) = \frac{1}{D} \sum_{d=1}^{D} \max \left( \sum_{i,j} (|\mathcal{F}(\mathbf{K}_{d,:,:})| \otimes \mathbf{M}), 0 \right), \tag{S11}$$

where $\mathcal{F}(\cdot)$ denotes the (centered) Fourier transform, $|\cdot|$ takes the amplitude of a complex input, $\otimes$ indicates the elementwise multiplication, $\sum_{i,j}$ summarizes all items along the spatial dimensions. $\mathbf{M}$ is a normalized zero-mean frequency mask whose values are linearly proportional to the spatial frequencies. Consequently, $\sum_{i,j}(|\mathcal{F}(\mathbf{K}_{d,:,:})| \otimes \mathbf{M})$ is a summary metric that measures the overall frequency response level of the convolutional kernel $\mathbf{K}_{d,:,:}$, where the negative value suggests a low-pass filter while the positive one tend to be a high-pass filter. Since our goal is to discourage high-pass responses without special preference for low-pass filters, a max operator is introduced to avoid the tendency that renders the kernel frequency response levels as low as possible.

The overall loss $\mathcal{L}$ for LKSV-CNN training can be written as,

$$\mathcal{L} = \lambda_1 \mathcal{L}_{ce} + \lambda_2 \mathcal{L}_{tv} + \lambda_3 \mathcal{L}_{hp} \tag{S12}$$

where we empirically set the weights as $\lambda_1 = 1$, $\lambda_2 = 0.1$, $\lambda_3 = 0.1$ for all experiments.

**Training Recipes for Different Datasets**  Our training recipes follow common deep learning practice. We have two SVN[3] designs with electronic backends with different layers (2 and 5 layers) for CIFAR-10[36] and ImageNet classification, respectively. The other results (image recognition on CIFAR-100[36], Flowers-102[37], Pet-37[38], Food-101[39] and image semantic segmentation on PASCAL VOC[40]) are obtained using the generic SVN[3] (designed for ImageNet features) by transfer learning (*i.e.* finetuning the electronic backend for each dataset). For all datasets, images are preprocessed into grayscale to conform with the setup of SVN[3] under monochromatic light.

- *CIFAR-10:* CIFAR-10 dataset is a well-known image classification dataset created by the Canadian Institute for Advanced Research (CIFAR) in 2009 that consists of 60,000 images in 10 classes, with 6,000 images in each class. The classes are airplane, automobile, bird, cat, deer, dog, frog, horse, ship, and truck. The images are of size $32 \times 32$ pixels and are divided into 50,000 training images and 10,000 test images. We use an AdamW[41] optimizer with a batch size of 128, weight decay of $1 \times 10^{-4}$, an initial learning rate of $1 \times 10^{-3}$, decayed through cosine annealing with a minimum value of $1 \times 10^{-5}$ for 200 epochs. We also use random cropping and horizontal flipping for data augmentation.

- *ImageNet:* ImageNet is a widely used and extensive dataset in the field of computer vision. Experiments are conducted on the ImageNet Large Scale Visual Recognition Challenge (ILSVRC) 2012-2017 image classification dataset. This dataset spans 1000 object classes and contains 1,281,167 training images and 50,000 validation images. Subsampled low-resolution images of the size of $64 \times 64$ are used due to the form factor constraints. We use an AdamW optimizer with a batch size of 256, weight decay of $1 \times 10^{-4}$, an initial learning rate of $1 \times 10^{-3}$, decayed through cosine annealing with a minimum value of $1 \times 10^{-5}$ for 100 epochs. Random cropping and horizontal flipping are adopted as data augmentation.

- *Transfer Learning for Other Image Recognition Tasks:* Transfer learning on CIFAR-100, Flowers-102, Pet-37, and Food-101 entails finetuning the electronic backend (of the generic ImageNet design). The finetuning recipe is almost the same as the one for ImageNet training, except only cross-entropy is used as training loss without other regularization losses.

- *Transfer Learning for PASCAL VOC Semantic Segmentation:* We use the same setting as the above transfer learning tasks except the batch size is set to 8.

**Supplementary Note 5: Differentiable Spatially-Varying Inverse Design**

To physically realize the LKSV convolutional layer, we propose a differentiable spatially-varying inverse design framework to engineer a meta-optic system yielding LKSV-PSFs as desired. We first describe the overall physical system setup, then derive the corresponding image formation model of the optical system, and finally formulate the inverse design as an optimization problem solved by a stochastic gradient-based optimizer.

**Physical System Setup.** Modeling the on-chip nanophotonic array imaging system (introduced in Supplementary Note 1 and the main paper) entails specifying associated system physical parameters, including the working wavelength $\lambda$, the object distance $d_o$, image distance $d_i$, discretization grid and resolution, sensor pixel pitch, and the certain spatial arrangement of the metalens array. Those physical parameters constitute a large design space with a high degree of freedom, which cannot be exhaustive in a trial-and-error approach. Instead, selecting these parameters must adhere to image formation physics and fabrication constraints. In our proof-of-concept prototype

17

design, the working monochromatic light wavelength is set to 525 nm and the sensor is modeled with a resolution of 1,936×1,216 and a pixel pitch of 5.86 μm to be consistent with the sensing device (Sony IMX174 CMOS sensor) we use in experiment. We use 50 metalenses arranged in a $6 \times 9$ array (as in Figure S4) are designed in parallel to induce 25 pairs of LKSV-PSFs, where PSFs of each pair are responsible for mimicking positive and negative parts of a target electronic kernel respectively. These metalenses are placed with equal horizontal and vertical spacing (0.375 mm). The scenes of interest are assumed to be located in a single-depth plane and the image of each metalens is designated to occupy 128×128 sensor pixels ensuring all 50 images can be simultaneously recorded in the sensor plane in a non-overlapped spatially-multiplexed manner. Field of view (FOV) and numerical aperture (NA) are manipulated by controlling the image distance $d_i$ (the object distance $d_o$ is linearly proportional to $d_i$ to maintain target magnification), which is optimally determined through a brute-force search to render the optical responses resemble the target PSFs as much as possible. An instantiation of our SVN[3] physical setup is summarized in Table S4.

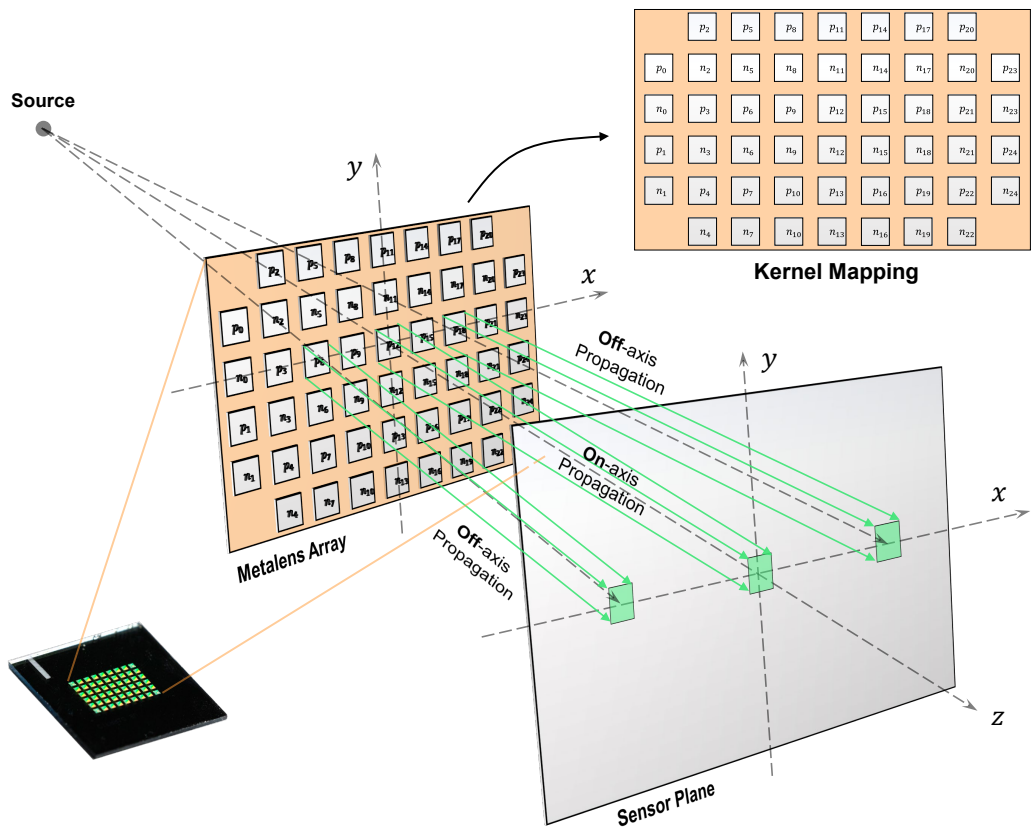

Figure S4: **Metalens array pattern and coordinate system.** The metalens $P_{12}$ is on-axis illuminated while the other metalenses are illuminated with off-axis illumination of varying angles. The front view of the chip is shown in the upper right. The kernel identifiers are also shown ($p$: positive; $n$: negative).

18

**Image Formation Model.** We adopt a point-based incoherent wave-optics model to simulate the image formation process. For CIFAR-10 classification, the scene at a single depth layer is modeled by 32×32 mutually incoherent point sources, each of which emits a spherical wave propagating through the metasurfaces with different incident angles. The coordinate system and the model geometry are illustrated in Figure S4, where the optical axis (crossing the scene center) is assumed to perpendicularly cross the center of metalens $P_{12}$. PSFs for different incident angles can then be derived by simulating free-space wave propagations of wavefronts transformed by metasurface phase profiles to the sensor plane. More formally, given a point source at coordinate $(\xi, \eta)$ and a metalens, the (normalized) source field $\mathbf{U}_s^{(\xi,\eta)}$ in the metalens plane is given by

$$\mathbf{U}_s^{(\xi,\eta)}(x,y) = \frac{e^{jkr_o}}{r_o} e^{j\mathbf{\Phi}(x,y)}, \tag{S13}$$

where $k = 2\pi/\lambda$ is the wavenumber, $r_o = \sqrt{(\xi-x)^2 + (\eta-y)^2 + d_o^2}$ is the Euclidean distance between the point source $(\xi, \eta)$ on the object plane and any pixel $(x, y)$ on the metalens plane. $\mathbf{\Phi}$ denotes the phase profile of the metalens subject to the working wavelength $\lambda$. Then, the destination field $\mathbf{U}_d$ in the sensor plane can be calculated by Rayleigh-Sommerfeld scalar diffraction integral[2], numerically implemented by a band-limited (shifted) angular spectrum propagation model[42,43] as

$$\begin{aligned}
\mathbf{U}_d^{(\xi,\eta)} &= \mathcal{F}^{-1}\{\mathcal{F}\{\mathbf{U}_s^{(\xi,\eta)}\} \otimes \mathbf{H}\} \\
\mathbf{H}(u,v) &= e^{j2\pi(x_0 u + y_0 v + d_i w)} \text{rect}\left(\frac{u-u_0}{u_{\text{width}}}\right) \text{rect}\left(\frac{v-v_0}{v_{\text{width}}}\right) \\
\boldsymbol{w}(u,v) &= \begin{cases} e^{j2\pi d_i \sqrt{\frac{1}{\lambda^2} - u^2 - v^2}} & \text{if } \sqrt{u^2+v^2} \leq \frac{1}{\lambda} \\ 0 & \text{otherwise} \end{cases}
\end{aligned} \tag{S14}$$

where $\mathcal{F}\{\cdot\}$ represents the fast Fourier transform, and $u, v$ are the corresponding spatial frequencies. $x_0, y_0$ indicate the spatial coordinate shifts — the origin of the destination coordinates $(\hat{x}, \hat{y})$ is shifted from the source coordinates as $\hat{x} = x - x_0$, $\hat{y} = y - y_0$. Note $x_0 = \frac{d_i}{d_o}\Delta d_x$, $y_0 = \frac{d_i}{d_o}\Delta d_y$ vary with metalenses, where $\Delta d_x$ and $\Delta d_y$ are the horizontal and vertical distances between the given metalens and the central one $P_{12}$. $\text{rect}(u)$ is a rectangular function with unity width, and $u_0$, $v_0$, $u_{\text{width}}$, $v_{\text{width}}$ are constants used for the band limit to avoid sampling aliasing errors[43]. Since the photodetectors in the sensor plane measure intensities of the wave field via the photoelectric effect, the PSF $\mathbf{K}_\mathbf{\Phi}^{(\xi,\eta)}$ resulted from the point source $(\xi, \eta)$ is written as

$$\mathbf{K}_\mathbf{\Phi}^{(\xi,\eta)} = |\mathbf{U}_d^{(\xi,\eta)}|^2 \tag{S15}$$

where $|\cdot|$ takes the amplitude of a complex field. The measured image $\mathbf{I}$ taken by the given metalens is therefore formed by

$$\mathbf{I} = \iint A(\xi,\eta) \mathbf{K}_\mathbf{\Phi}^{(\xi,\eta)} d\xi d\eta, \tag{S16}$$

where $A(\xi,\eta)$ is the point source $(\xi,\eta)$ intensity, the factor normalized in Equation (S13). As the system geometry ensures no cross-talks/overlaps among measured images on the sensor plane, all 50 images are formed by the nanophotonic array camera independently yet simultaneously.

**Inverse Design as Stochastic Optimization.** We implement the aforementioned image formation model as a differentiable forward operator parameterized by a discretized phase mask $\boldsymbol{\Phi}$ that can be optimized for specified angle-sensitive PSFs. Given the LKSV convolutional stem $f^{(s)}$ learned in an in-silicon training phase, the target PSFs (ground truth) $\mathbf{K}_{\text{gt}}^{(\xi,\eta)}$ are generated by transforming canonical basis components through $f^{(s)}$ akin to probing impulse responses of a linear system,

$$\mathbf{K}_{\text{gt}}^{(\xi,\eta)} = f^{(s)}\left(\mathbf{E}^{(H-\xi,W-\eta)}\right) \tag{S17}$$

where $\mathbf{E}^{(H-\xi,W-\eta)} \in \mathbb{R}^{H \times W}$ denotes a canonical basis component whose entries are all zero, except one at coordinate $(H-\xi, W-\eta)$ that equals 1. The negative signs of $\xi$ and $\eta$ reflect the inverted real image formed through a single thin lens by nature. To faithfully mimic the SV property of $f^{(s)}$, the optical PSFs of a metalens should be optimized jointly with respect to different incident angles (discretized in source plane coordinates). Since the ground truth is only defined up to a scale factor, we define a loss function that calculates scale-invariant mean square error between the optical and electronic PSFs as

$$\mathcal{L}_{psf}(\boldsymbol{\Phi}) = \sum_{\xi,\eta} \|\hat{\alpha} \mathbf{K}_{\boldsymbol{\Phi}}^{(\xi,\eta)} - \mathbf{K}_{\text{gt}}^{(\xi,\eta)}\|_2^2 \tag{S18}$$

where $\hat{\alpha} = \arg\min_{\hat{\alpha}} \mathcal{L}_{psf}$ is a single scalar analytically computed to minimize the mean square error between scaled optical PSFs and target ones at each loss evaluation. Besides, we add light efficiency regularization, defined as

$$\mathcal{L}_{light}(\boldsymbol{\Phi}) = \sum_{\xi,\eta} \left(1 - \sum_{h,w} \mathbf{K}_{\boldsymbol{\Phi}}^{(\xi,\eta)}\right) \tag{S19}$$

to boost the light efficiency of the designed optical system, where the inner sum $\sum_{h,w} \mathbf{K}_{\boldsymbol{\Phi}}^{(\xi,\eta)}$ is the received energy in sensor plane for a specific incident angle. The final loss function for inverse design thus reads:

$$\mathcal{L}_{inv} = \mathcal{L}_{psf} + \lambda_{light} \mathcal{L}_{light} \tag{S20}$$

where $\lambda_{light}$ is empirically set as $10^{-4}$ in experiment. It is then minimized via gradient descent to leverage the differentiability of the $\mathbf{K}_{\boldsymbol{\Phi}}^{(\xi,\eta)}$ with respect to $\boldsymbol{\Phi}$. Nevertheless, backpropagating the loss (S20) entails a large volume of computing resources due to massive ($H \times W$ times) PSF computations involved in the loss evaluation. As a result, we adopt a stochastic mini-batch version of (S20) with randomly sampled incident angles $(\xi, \eta)$ and use an adaptive stochastic optimizer namely Adam[44] to minimize it. The batch size is 32 and the total number of iterations is 5,000. The learning rate follows a cosine annealing rule that decays from 1 to $1 \times 10^{-6}$ during optimization. The same optimization process is repeatedly executed for each of the 50 meta lenses, where the forward model (on- and off-axis propagation) varies according to the system geometry.

The proposed optical simulation framework allows for pre-fabrication evaluation of the design quality through metrics such as PSF approximation normalized root mean square error (NRMSE),

image feature discrepancy in NRMSE, and light efficiency. See Figure S5 and S6 for the prefabrication assessment of CIFAR-10-based and generic SVN[3] designs, respectively. Take CIFAR-10-based design as an example, the PSF error map, as shown in Figure S7, indicates that the simulated spatially varying PSFs at each of the $32 \times 32$ sampling incident angles closely resemble those of the target electronic kernels with $7.67\%$ mean NRMSE. The bar plots in Figure S5b and S5c show the PSF error and image feature discrepancies (NRMSE) for the 50 metalenses, respectively. The feature discrepancies in NRMSE ($39.06\%$ at average) are higher than the PSF errors owing to inaccuracies that can arise from representing discrete convolution kernels using spatially continuous light fields, as well as the error amplification caused by ill-conditioned target kernels. Note that these target kernels were regularized by the specialized spectrum penalty at the training phase (Supplementary Note 4), which reduced their condition numbers and thus reduced the average image feature discrepancy (NRMSE) from $66.46\%$ to $39.06\%$. Although enforcing stronger condition regularization can further close the gap between simulated device performance and real-world device performance, this would severely degrade the model recognition accuracy, as less regularized kernels could extract more discriminative features for classification.

**Phase to Structure Mapping.** The aforementioned optimization approach yields 50 individual phase profiles (Figure S8) representing the positive and negative components of the 25 kernels. To physically implement the specific phase profiles as a meta-optic structure, we mapped it to the phase response of nanopillars. The different phase delays of the nanopillars are realized by varying the side width, assuming a local periodic approximation. For this purpose, we first created a library of the phase response for square nanopillars with varying side widths in SiN of 750 nm height on top of a thin film of 50 nm SiN on top of a 0.5 mm quartz substrate. We use a period of 390 nm, which matches a third of the period of the ideal kernel phase profiles. Thus by creating pillar sub-arrays with a size of $3 \times 3$, we can match the lattice constant of the kernel. The phase response was calculated using rigorous coupled wave analysis (RCWA), specifically using the S4 package[45]. The resulting transmission and phase for transmitted light is plotted in Figure S9a. Also shown are the 15 levels of nanopillars that were chosen to discretize the phase and mapped to the phase profile. A section of a kernel, overlayed with the structural dimension of the meta optic is shown in Figure S9b.

**Supplementary Note 6: Metasurface Fabrication**

We fabricated the meta-optic in a SiN thin film on a double-side polished fused silica wafer, measuring 500 micrometers in thickness. In detail, initially a thin film of silicon nitride with a thickness of $800\,\mathrm{nm}$ was grown on the wafer, using plasma-enhanced chemical vapor deposition (PECVD) within a SPTS DeltaX PECVD apparatus. This growth operation utilized Silane and Ammonia as the precursor gases at a temperature of 350 degrees Celsius.

Upon completion of the growth, the wafer was diced into pieces measuring $2 \times 2$ centimeters

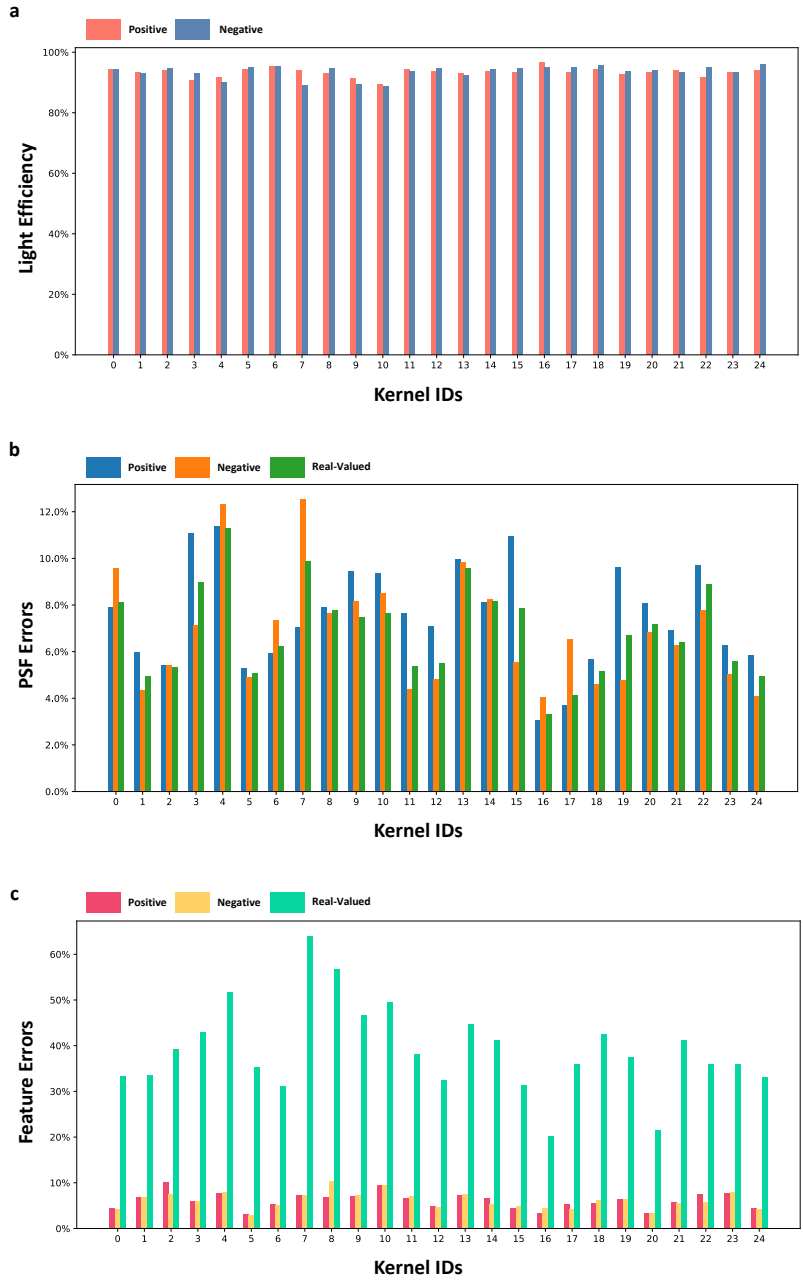

Figure S5: **Quantitative characterization of CIFAR-10-based SVN[3] design.** We assess all 50 metalenses (25 positive + 25 negative kernels) in three different aspects, including (a) light efficiency: the percentage of input light energy preserved in the output sensor plane after metalens modulation and free-space diffractive wave propagation; (b) optical PSF deviation from their electronic ground truth in terms of normalized root mean square error (NRMSE); and (c) optical image feature NRMSE errors averaged over 50 randomly sampled images.

22

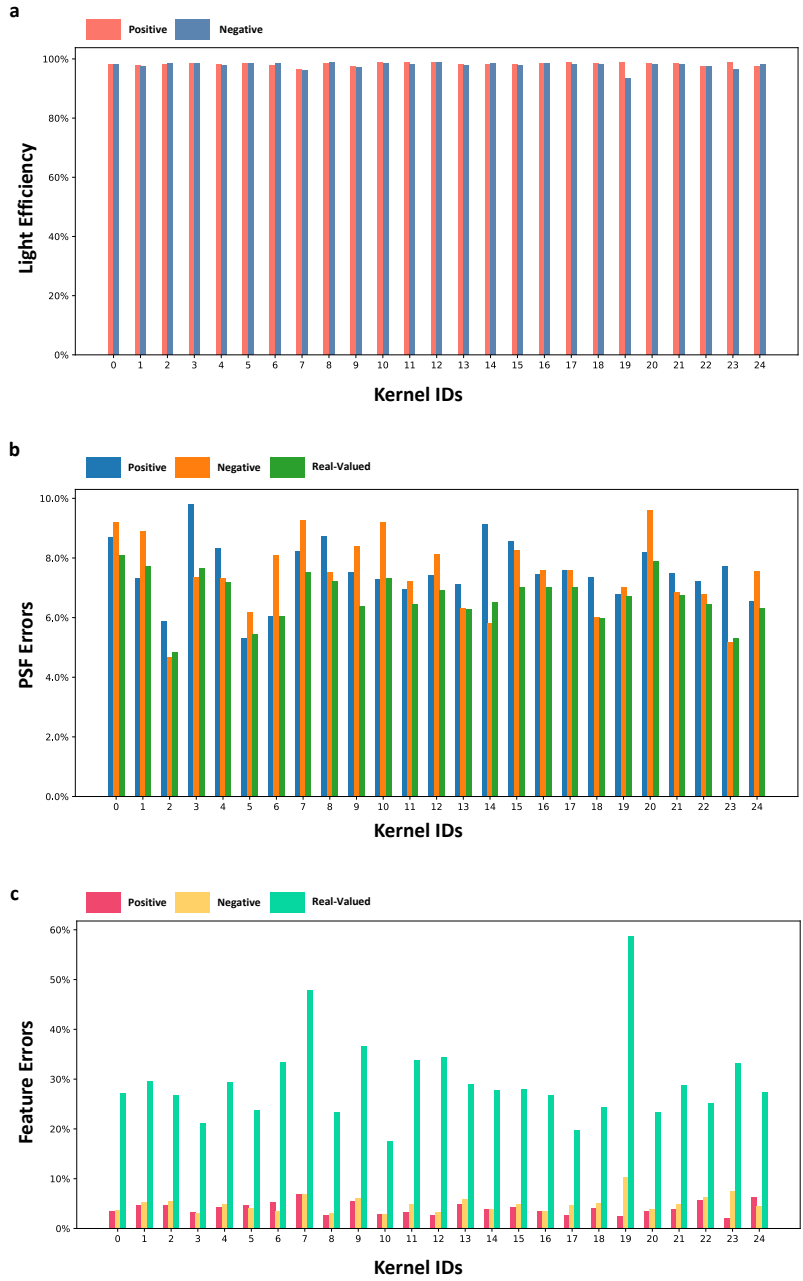

Figure S6: **Quantitative characterization of the generic (ImageNet)-based SVN[3] design.** We assess all 50 metalenses (25 positive + 25 negative kernels) in three different aspects, including (a) light efficiency: the percentage of input light energy preserved in the output sensor plane after metalens modulation and free-space diffractive wave propagation; (b) optical PSF deviation from their electronic ground truth in terms of normalized root mean square error (NRMSE); and (c) optical image feature NRMSE errors averaged over 50 randomly sampled images.

23

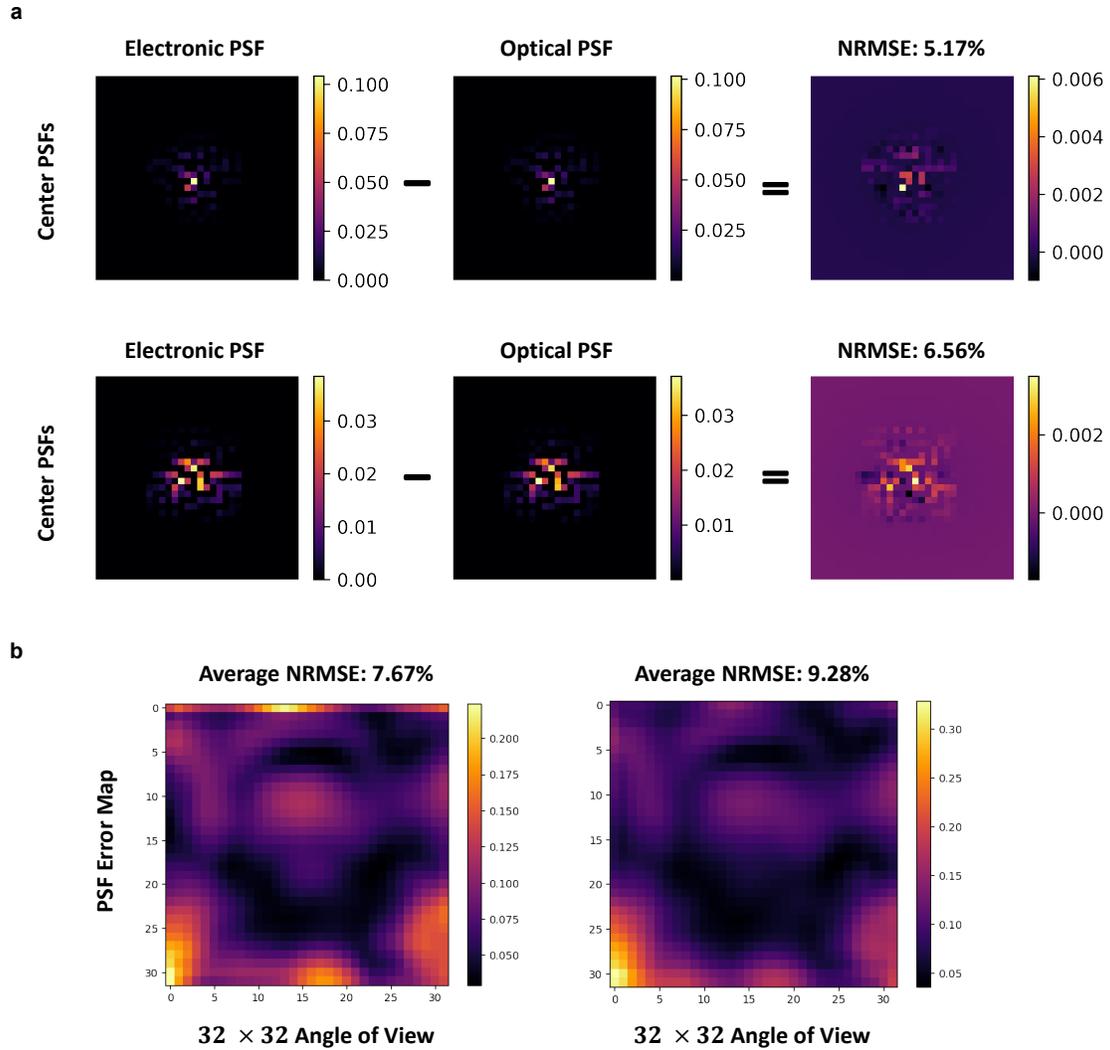

**Figure S7: Side-by-side comparisons of inverse-designed optical PSFs and their corresponding electronic ground truth.** (a) Visualization of center electronic and optical PSFs along with the residual error map. The residual error map is calculated as the difference between the electronic and the optical PSF. The numerical metrics, *i.e.*, normalized root mean square error (NRMSE) are shown above the residual error map. (b) PSF error map displaying the NRMSE across $32 \times 32$ different angles of view.

24

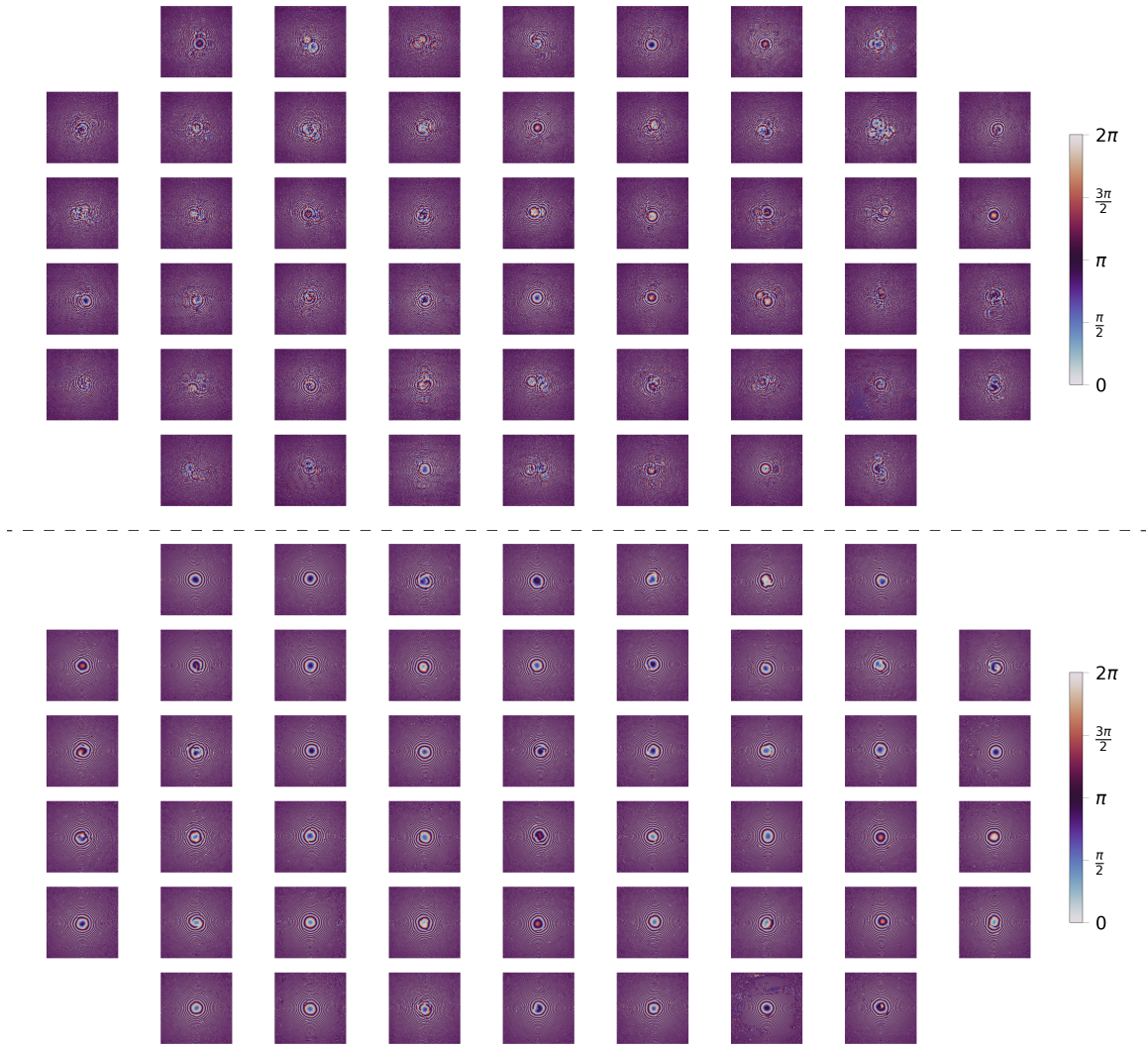

Figure S8: **Inverse designed phase profiles of SVN$^3$.** Top: CIFAR-10-specialized design; Bottom: Generic front-end design.

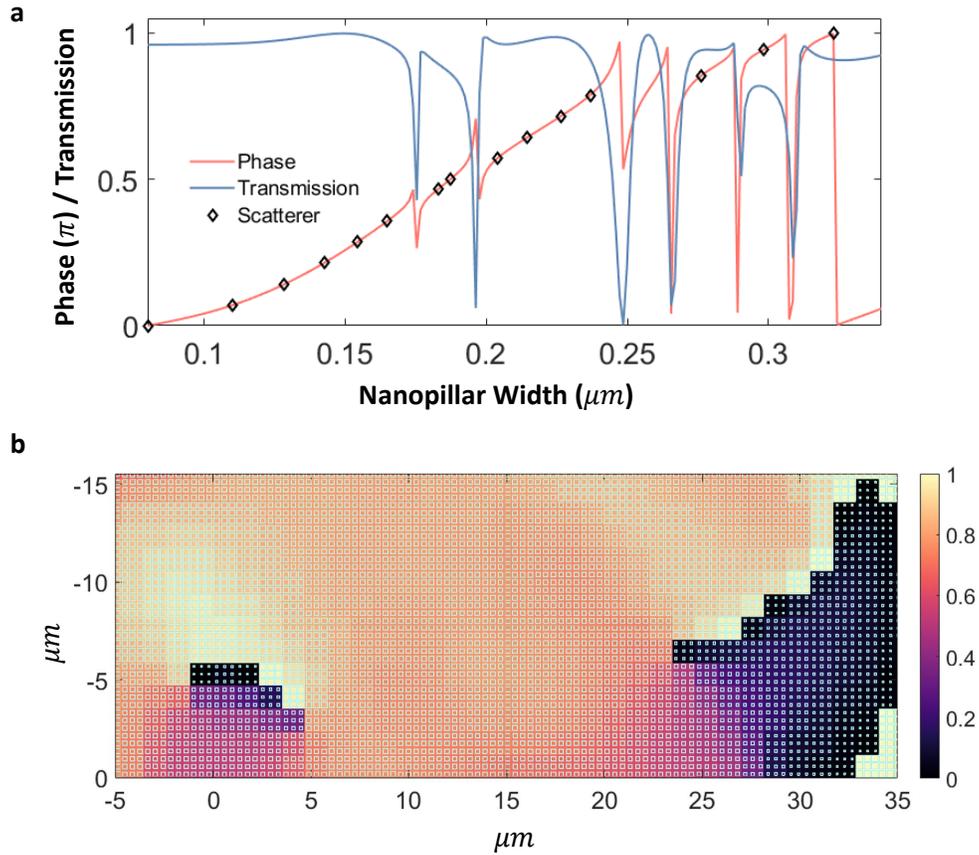

**Figure S9: Phase to structure mapping.** (a) Dependence of the phase (red) and transmission (blue) on the structural parameters for SiN nanopillars with a period of 350 nm and varying sidewidth. The 15 levels of nanopillars for the structural implementation are marked as diamonds. (b) Example of phase-to-structure mapping. The phase response of the pillar is mapped to the kernel phase, and the resulting structure is shown as outlines overlayed with the corresponding kernel phase.

and underwent a thorough cleaning procedure. The cleaning involved immersing the wafer in an Acetone sonicating bath, followed by a rinse in Iso-Propyl Alcohol (IPA). Subsequently, a brief exposure to O2 plasma using a barrell etcher at 100 watts for approximately 15 seconds further cleansed the sample. Post-cleaning, a layer of ZEP 520A resist was spin-coated on the sample (approximately 400 nanometers thick), succeeded by spin-coating of a discharging polymer film (DisCharge H2O).

Next, the kernels arrays for both spatially varying and invariant designs, were inscribed onto individual chips utilizing electron beam lithography (EBL). This process employed a JEOL-JBX6300FS machine with an acceleration voltage of 100 kV and a beam current of 8 nA. Following the EBL procedure, the discharging polymer layer was removed, by immersing the sample in IPA and subsequently the resist was developed in amyl acetate for 2 minutes, again followed by immersion in IPA to stop the developer. To create a hard mask, $65\,\text{nm}$ of alumina was deposited using a lab-built e-beam evaporator with an Al2O3 evaporation source. Subsequently, the resist layer underwent an overnight lift-off in N-Methyl-2-pyrrolidone (NMP) at a temperature of 110 degrees Celsius. Additional cleaning in a short-duration O2 plasma etch effectively eliminated any remaining organic residues.

In the next step, we utilized an inductively-coupled reactive ion etching (using an Oxford Instruments PlasmaLab100 system), employing a Fluorine-based etch chemistry (SF6 and C4F8 with a ratio of 1 to 2). This process transferred the hard mask into the underlying silicon nitride film, whereas the etching time was set to achieve a height in the SiN pillars of approximately 750 nm. The remaining 50 nm PECVD layer served as to enhance the stability of the etched device layer.

Subsequent to the fabrication of the meta optic, we deposited a metal aperture layer surrounding the structured surface. This prevents undesired spurious stray light reaching the sensor. The creation of these apertures was accomplished through optical direct write lithography using a Heidelberg-DWL66 apparatus, and a subsequent deposition of an approximately 150 nm Cr film using sputter deposition.

**Supplementary Note 7: SVN$^3$ Prototype**

We built two experimental setups to characterize the optical performance of various metalens array samples that we have designed in the development of the final device. The first one is used to measure the on-axis Point Spread Functions (PSFs) of the metalens array samples, while the second one is used to realize the designed optical system and to measure the image features, which are the images of various datasets, such as CIFAR-10 and ImageNet images, through the metalens array samples.

**Setup 1: PSF Measurement** In this setup, a 520 nm pigtailed single-mode fiber laser is used to mimic a point light source, and a monochromatic CMOS sensor is used as the detector to measure the intensity response of a metalens array sample upon the incidence of a point light source positioned at the designed object distance. In the light path between the metalens array and the detector, a microscopic objective together with a relay lens is used to increase the angular resolution of the PSF measurements thus generating magnified high-resolution PSFs on the detector plane. The fiber tip, the microscopic objective, the relay lens, and the sensor are all positioned on-axis to the central metalens element in a metalens array. The metalens array sample is mounted on a 3D translational stage, such that the on-axis PSFs of different metalens elements in an array can be measured. In addition, since the measurement Field-of-View is large enough, the PSFs of a small 3x3 array can be captured in one shot and thus for spatially-varying metalens samples, we can measure the off-axis PSFs at the incident angle determined by the object distance and metalens spacing.

Figure S10 shows the qualitative comparison of the experimentally measured on-axis PSFs and the corresponding electronic ground truths for four metalens elements. We observe that the fabricated samples adequately match the PSFs from simulation.

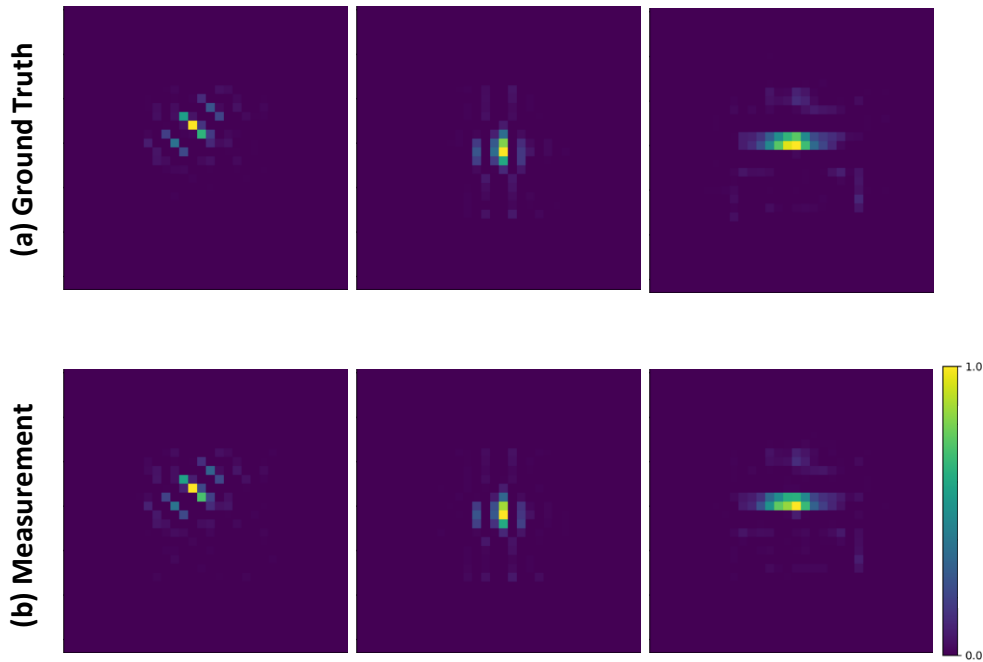

**Figure S10: Visual comparison of the experimentally measured center PSFs and the corresponding electronic ground truths.** We measure the on-axis PSFs from three metalens elements using the setup described in Supplementary Note 7.

**Setup 2: SVN$^3$ System**   In this setup, as in Figure S11, we placed OLED display smartphone at the designed object distance and used the green channel of the display as the incoherent light source. A large-area monochromatic CMOS sensor is placed at the focal plane of the metalens array device to capture the images of various datasets being displayed on the phone, such as CIFAR-10. The display is mounted on a 1D translational stage such that its distance to the metalens array sample can be accurately adjusted. The sensor is mounted on a 3D translational stage such that through fine adjustments, the images can be focused and fit into the sensor active area. The display and the sensor are controlled by a computer and synchronized such that when the dataset images are displayed on the smartphone sequentially as objects, the sensor captures the corresponding images in a single shot. We experimentally measured the display gamma of the smartphone to be $\approx 2.2$ and applied a gamma correction of 0.45 to the dataset images to compensate for the display gamma. In addition, to reduce the intensity variation of the captures caused by the 60Hz refresh rate of the display, we chose the exposure time of the sensor to be integer times of 16.7 ms. Experimentally captured raw images produced by this SVN$^3$ prototype are shown in Figure S12, which are further processed into 50 optical feature maps by cropping the designated regions in the raw measurement. Figure S13 and S14 exhibit side-by-side comparisons of measured optical features and their electronic counterparts for CIFAR-10 and generic SVN$^3$ designs respectively, demonstrating the high accuracy of our optical computing engine to execute designated large-kernel spatially-varying convolution.

**Supplementary Note 8: Ablation Study on LKSV Design**

As detailed in Supplementary Notes 1 and 3, the LKSV is designed to fully harnessing the capabilities of free-space optical computing. Table S2 summarizes the model profiles, encompassing both optical and electronic aspects, across various design variants. These range from the ($3 \times 3$) small-kernel spatially-invariant (SKSI) convolutional stem to the full ($15 \times 15$) large-kernel spatially-varying (LKSV) design. All opto-electronic architectures share a consistent two-layer electronic backend (refer to Supplementary Note 3), with 2.18K parameters and 216.50K multiply-accumulate operations (MAC) when processing a $32 \times 32$ input image from the CIFAR-10 dataset.

We analyze MAC, inference runtime, and parameter count of the optical model component when implemented digitally, underscoring the computational advantages intrinsic to optical computing. Leveraging low-dimensional re-parameterization techniques, our proposed LKSV model significantly outperforms its naïve SKSI counterpart by a substantial margin (65.45% to 73.80%), akin to the transition from LeNet to AlexNet. The accompanying substantial increase in MAC operations (280.60K to 34.61M) can be entirely offloaded into optics for "free". It is noteworthy that the performance of LK and/or SV models suffers significantly without the proposed low-dimensional reparameterization. This highlights the pivotal role of model parameterization in unleashing the potential of free-space optical computing.

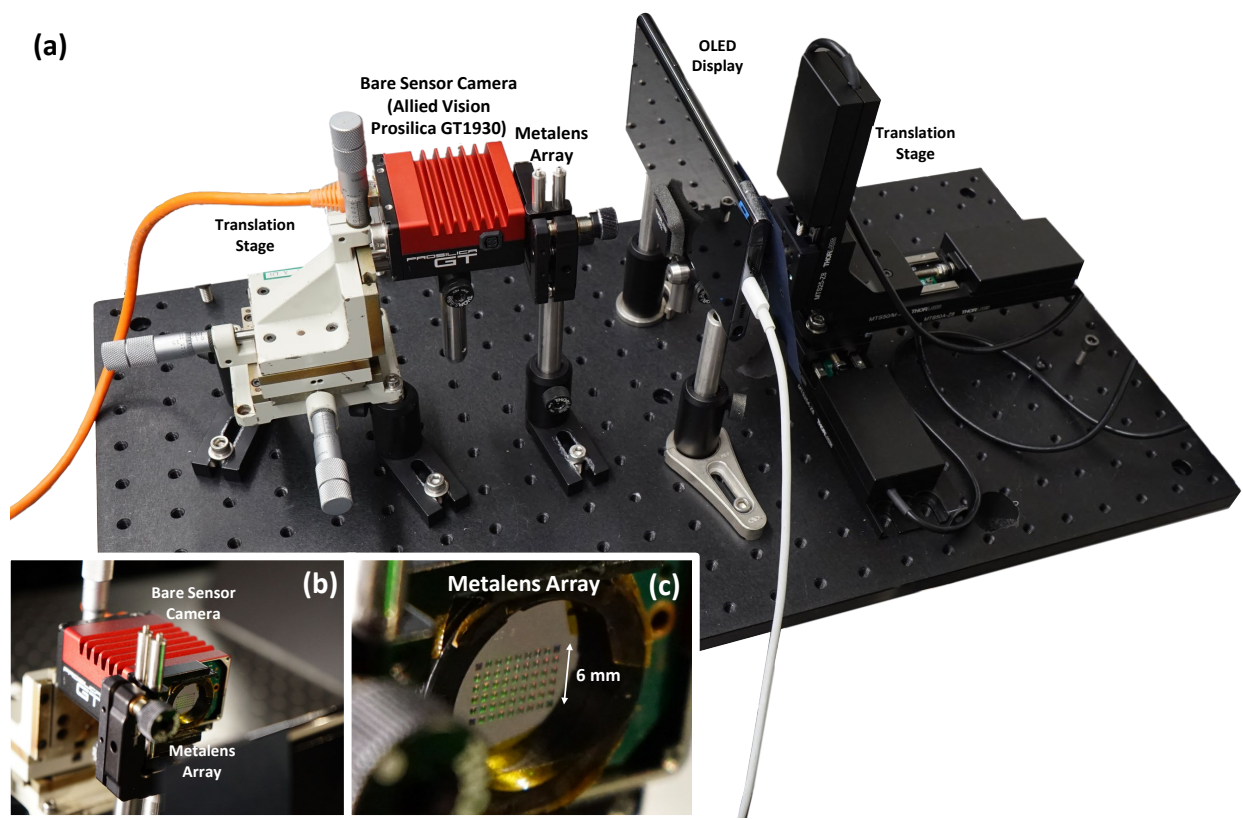

**Figure S11: SVN**[3] **experimental prototype.** (a) A smartphone OLED display serves as the incoherent illumination source, and the designed metasurface array is mounted in front of a CMOS photosensor. We show (b) another perspective of the computational camera setup and (c) a detailed close-up showcasing the individual miniature metalenses. The entire metalens array spans $6\,\mathrm{mm}$ length in height, $9\,\mathrm{mm}$ in width and $0.5\,\mathrm{mm}$ in depth.

30

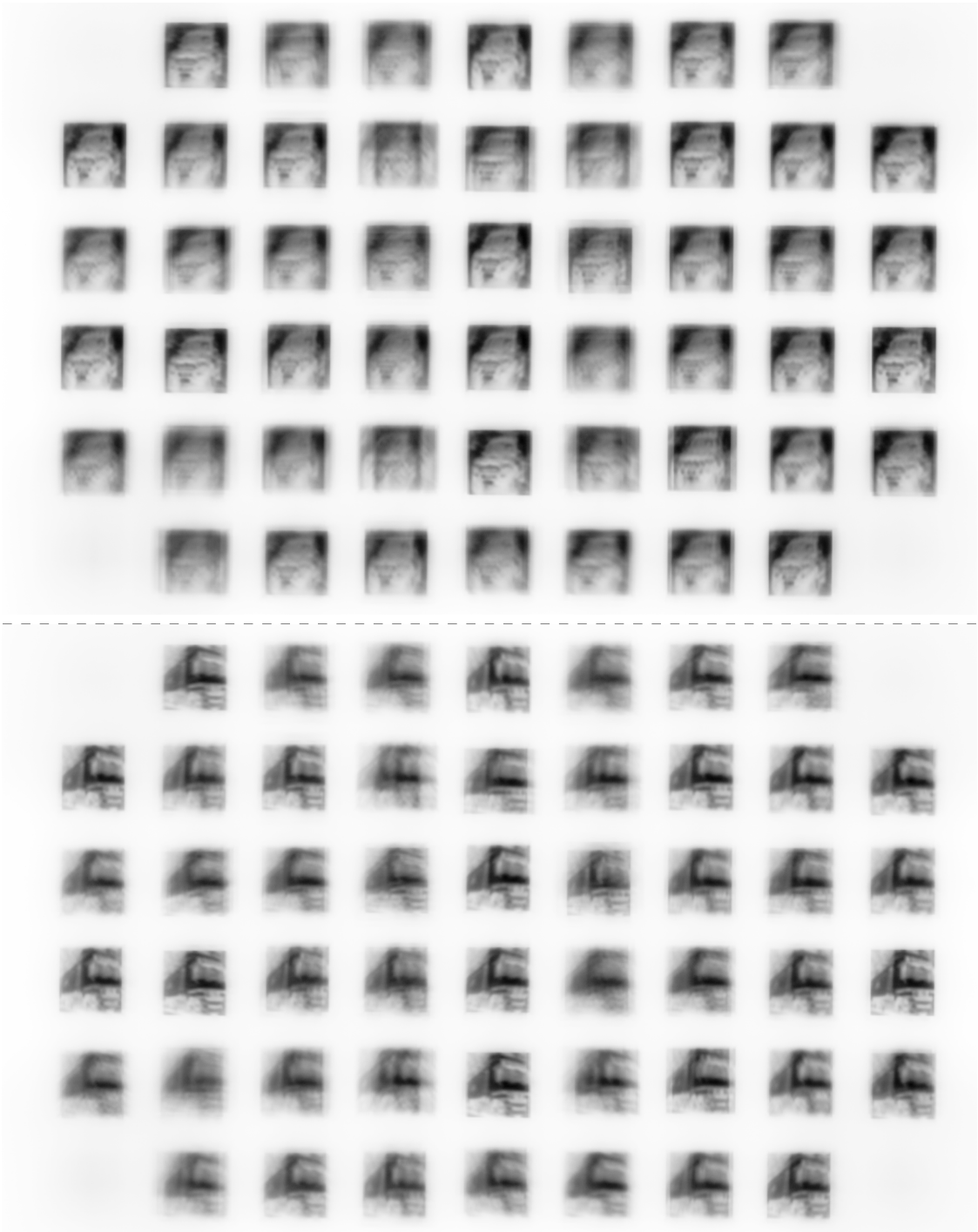

**Figure S12: Experimentally measured raw feature samples produced by SVN³.** Alignment images formed by traditional hyperbolic lenses located at the four corners are discarded.

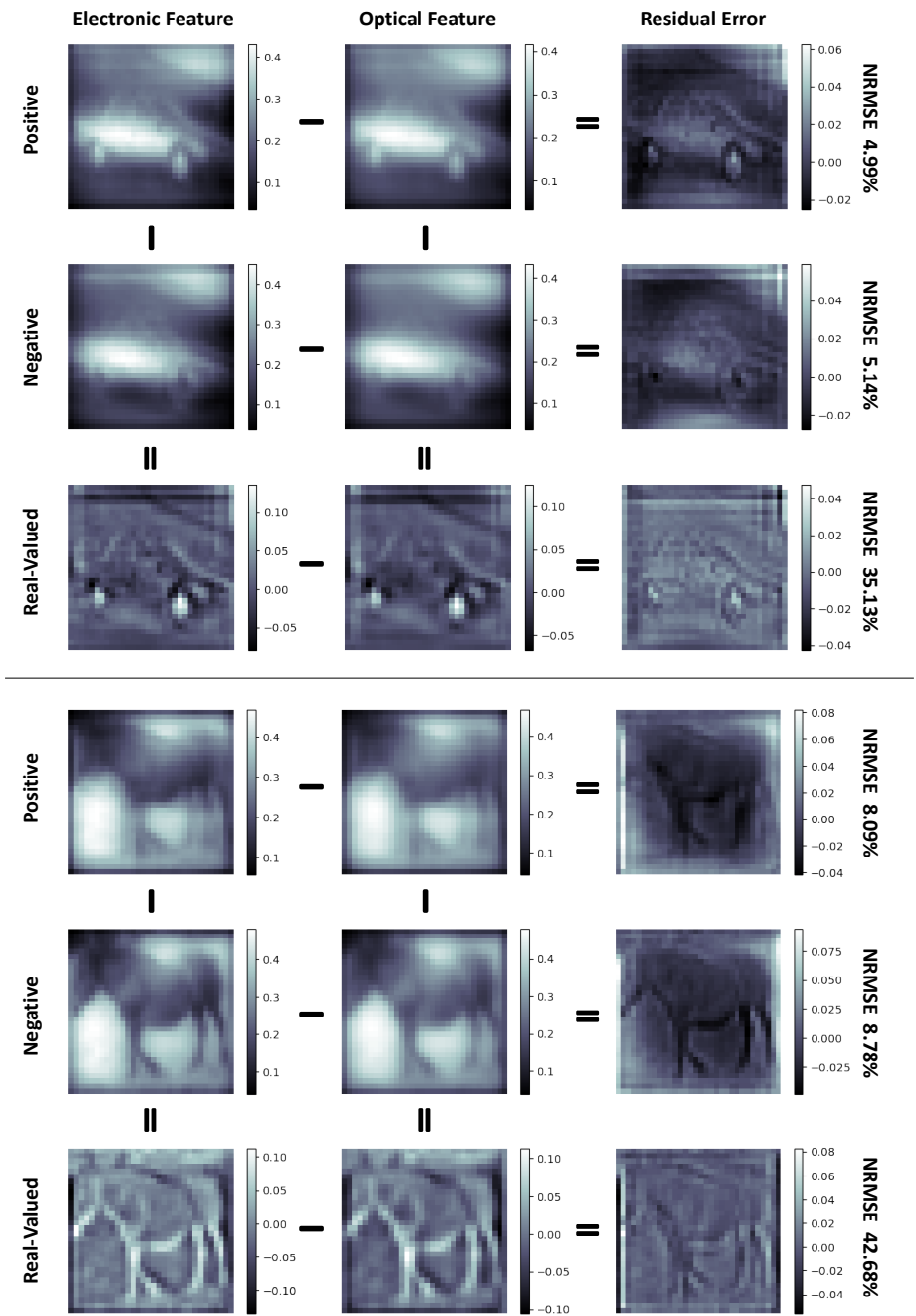

**Figure S13: Side-by-side comparisons of optical features and their corresponding electronic ground truth for the CIFAR-10-based SVN³ design.** The real-valued features are obtained by subtracting the positive features from the corresponding negative ones. (See minus "-" and equal "=" signs above)

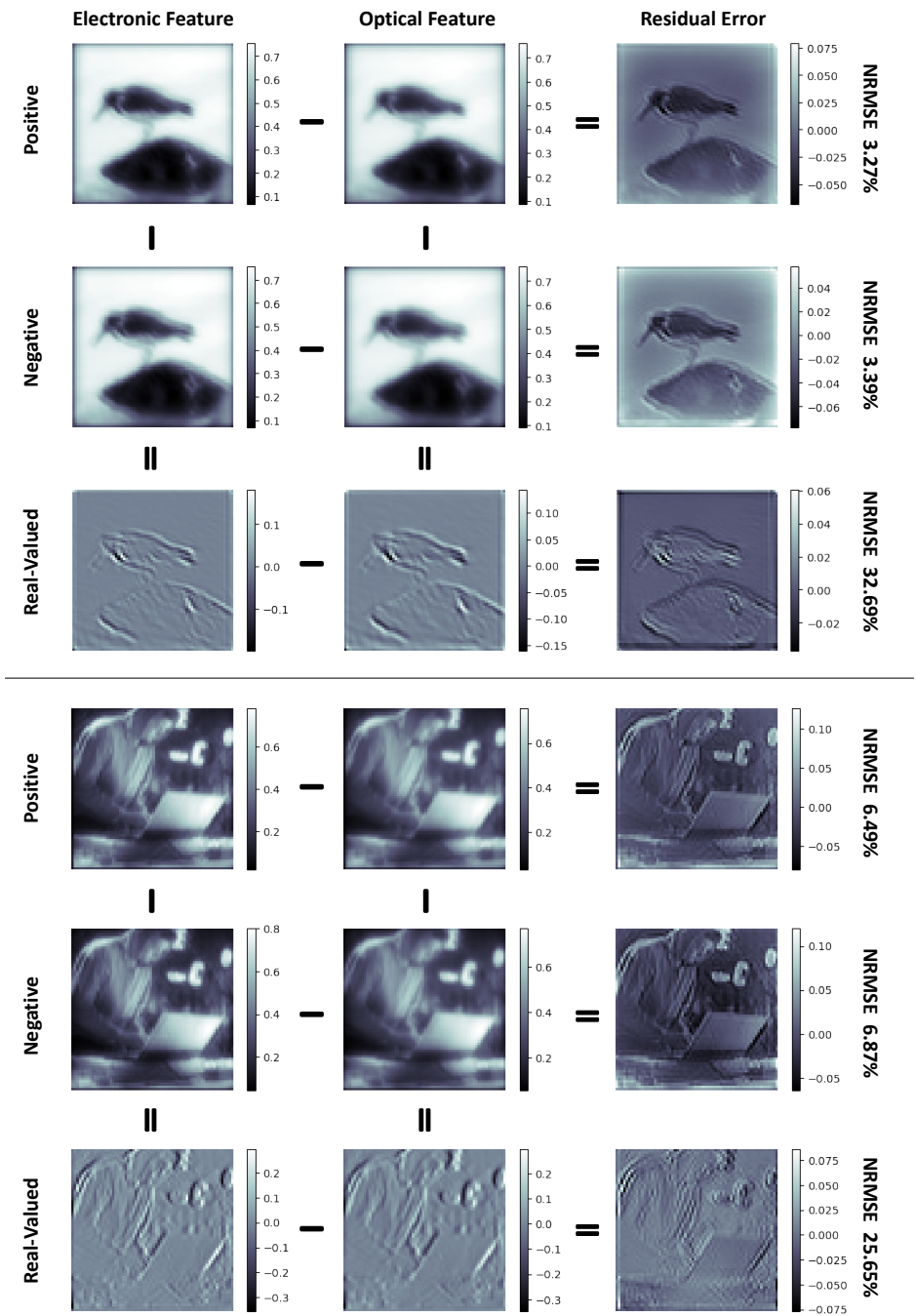

**Figure S14: Side-by-side comparisons of optical features and their corresponding electronic ground truth for the generic (ImageNet)-based SVN³ design.** The real-valued features are obtained by subtracting the positive features from the corresponding negative ones. (See minus "-" and equal "=" signs above)

33

**Supplementary Note 9: Additional Experimental Results**

Additional experimental classification results by SVN$^3$ on random samples from CIFAR-10 and ImageNet datasets are shown in Figure S15 and S16 respectively.

As demonstrated in our main paper, leveraging an inverse-designed optical frontend based on ImageNet-trained kernels, SVN$^3$ functions as a versatile computational camera featuring a universal optical encoder. Through fine-tuning of the electronic backend for different datasets and tasks, the optoelectronic network becomes readily applicable across diverse downstream applications without necessitating a redesign of the optical system. The quantitative performance of the generic SVN$^3$ on various downstream tasks is summarized in Table S5. Notably, SVN$^3$ significantly surpasses AlexNet in fine-grained object recognition tasks, such as the detailed identification of flowers, food items, and pets, validating its generalizability.

Furthermore, we extend our evaluation by comparing the performance of SVN$^3$ and AlexNet in the image semantic segmentation task using the PASCAL VOC dataset. To achieve semantic segmentation, we removed the global average pooling and the fully-connected classification head, replacing them with a $1 \times 1$ convolutional layer (adjusting the channel dimension as per the number of classes) followed by bilinear upsampling to predict dense segmentation masks. we report three metrics to quantitatively evaluate the performance of semantic segmentation, focusing on pixel accuracy and region intersection over union (IoU). Expressed as mathematical formulas, these metrics include 1) Pixel accuracy: $\sum_{ii} n_{ii} / \sum_i t_i$; 2) Mean accuracy: $1/n_{cl} \sum_i n_{ii}/t_i$; and 3) Mean IoU: $1/n_{cl} \sum_i n_{ii}/(t_i + \sum_j n_{ji} - n_{ii})$.

The numerical and visual results presented in Table S6 and Figure S17 show that SVN$^3$ yields comparable results to AlexNet in this challenging task of image semantic segmentation. Notably, SVN$^3$ accurately segments elements such as flights, animals, and human figures within complex scenes, demonstrating the capability of the optical neural network to handle complex high-level vision tasks beyond image classification.

Table S4: **Design parameter specification and equipment list** for the optical system design and setup of SVN$^3$ in CIFAR-10 and ImageNet classification experiments.

| PARAMETERS | CIFAR-10 MODEL | IMAGENET MODEL |
|---|---|---|
| Wavelength | 525 nm | 525 nm |
| Nanopillar size | 390 nm | 390 nm |
| Metalens size | 0.75 mm$^2$ | 0.75 mm$^2$ |
| Metalens spacing | 0.375 mm | 0.375 mm |
| Object distance | 86.5 mm | 103.8 mm |
| Image distance | 4.00 mm | 4.80 mm |
| Sensor pixel pitch | 5.86 μm | 5.86 μm |
| Sensor image pixels | $128 \times 128$ | $128 \times 128$ |
| EQUIPMENT | PRODUCT NAME ||
| Camera | Allied Vision Prosilica GT1930 ||
| Smartphone OLED display | SM-S908U1 ||

Table S5: **Performance comparison of SVN$^3$ and AlexNet for image classification tasks.** We show the top-1 and top-5 recognition accuracy. Here, SVN$^3$ denotes the generic instance consisting of a 5-layer electronic backend. The classification accuracy obtained with simulated measurements is shown in the middle column. The classification accuracy obtained with experimental measurements obtained with our hardware prototype is shown in the right column.

| CLASSIFICATION | ALEXNET | | SVN$^3$ (SIMULATION) | | SVN$^3$ (EXPERIMENT) | |
|---|---|---|---|---|---|---|
| | TOP-1 ↑ | TOP-5 ↑ | TOP-1 ↑ | TOP-5 ↑ | TOP-1 ↑ | TOP-5 ↑ |
| ImageNet-1000 | 26.99 % | 47.68 % | 28.43 % | 51.28 % | 26.62 % | 48.64 % |
| CIFAR-100 | 55.00 % | 77.74 % | 52.89 % | 80.34 % | 52.22 % | 79.13 % |
| Flowers-102 | 38.15 % | 62.22 % | 52.94 % | 76.29 % | 49.63 % | 74.89 % |
| Food-101 | 27.97 % | 54.97 % | 39.47 % | 66.71 % | 37.12 % | 64.36 % |
| Pet-37 | 44.37 % | 76.81 % | 47.97 % | 81.11 % | 46.31 % | 80.01 % |

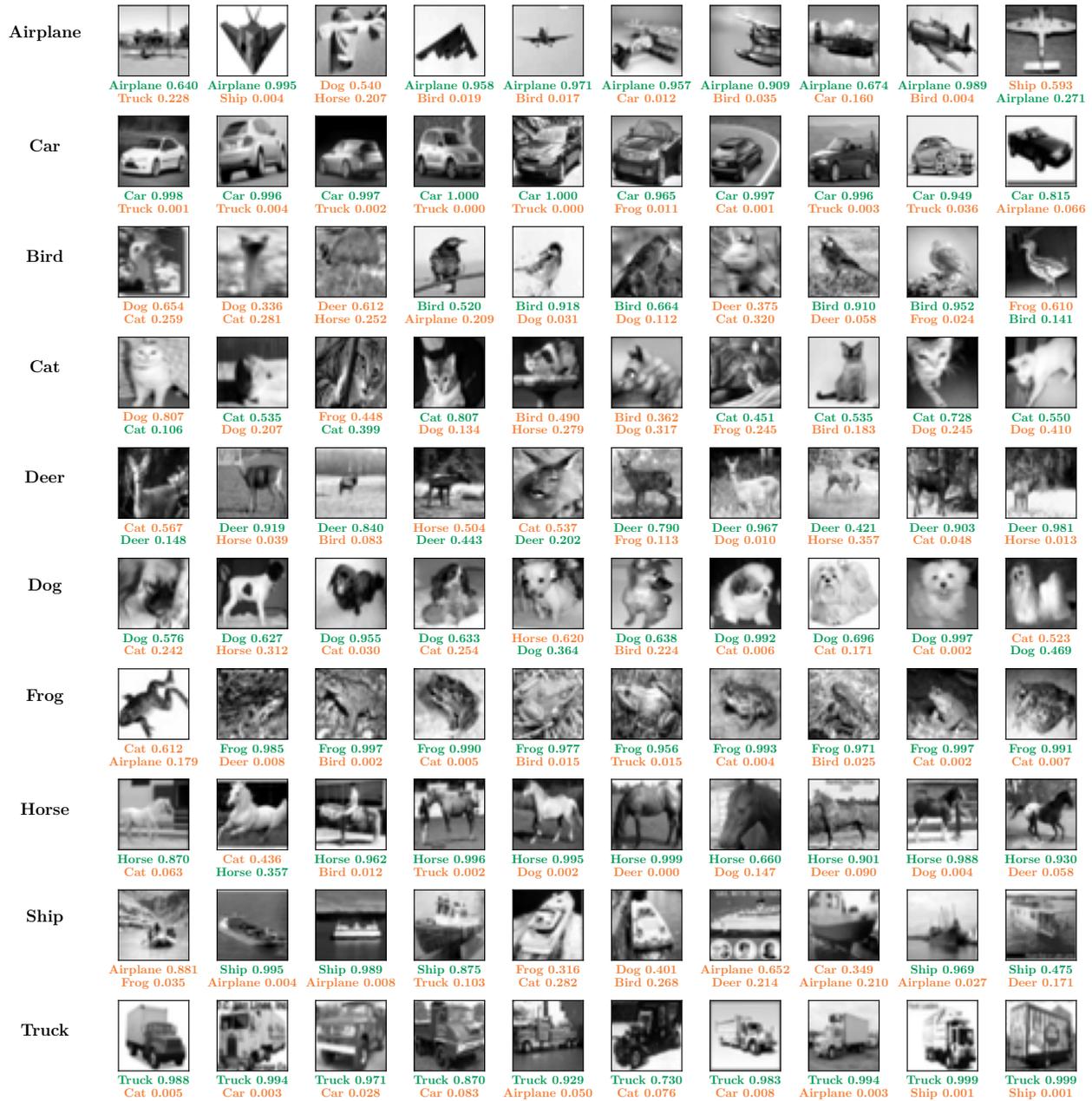

Figure S15: **Additional experimental classification results on random samples from CIFAR-10 test set (Part-1)**. We show SVN³'s predicted categories with top-2 highest probabilities below each test image. Green and Orange colored labels under the images denote the correct and incorrect predictions, respectively.

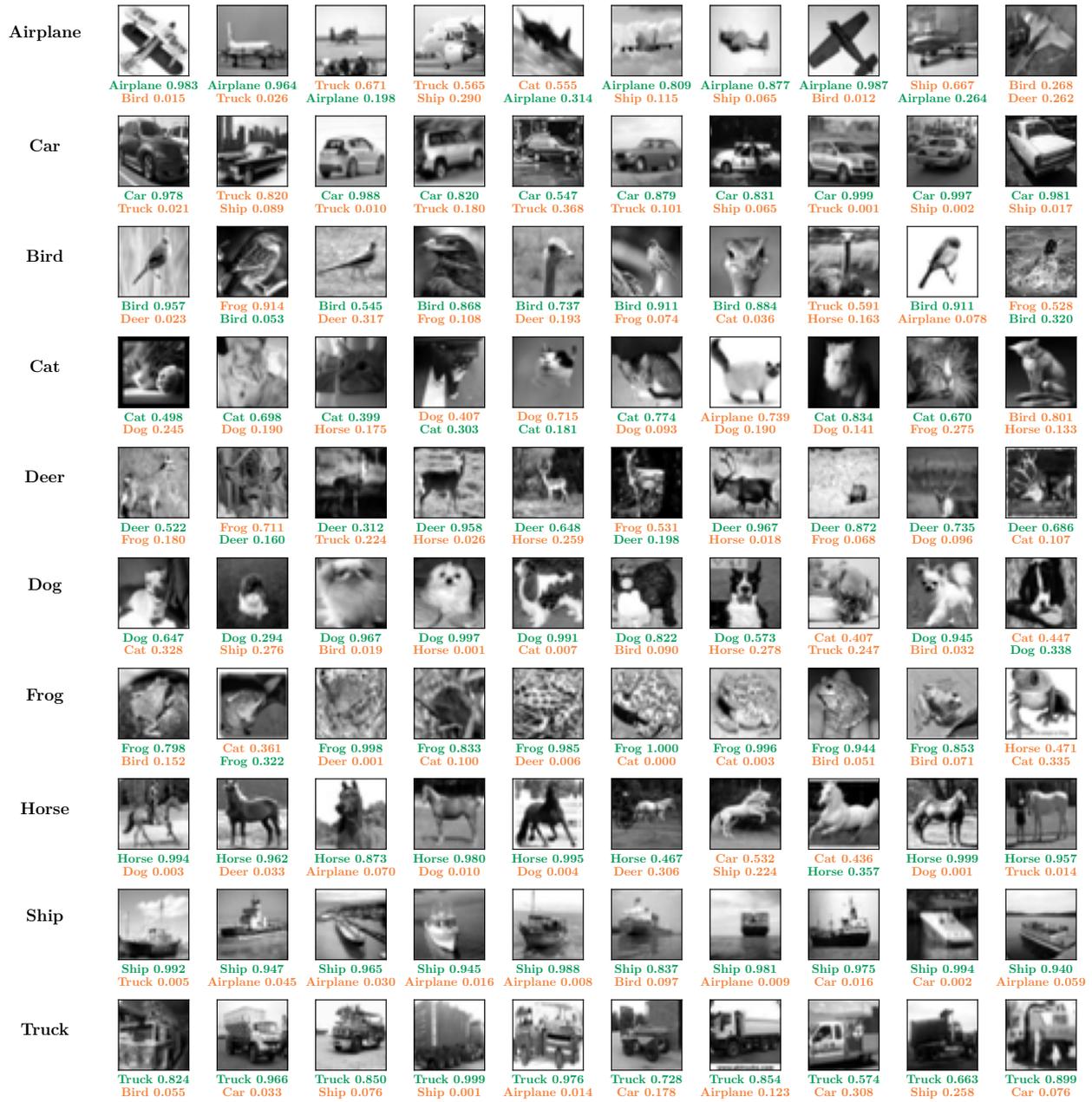

Figure S15: **Additional experimental classification results on random samples from CIFAR-10 test set (Part-2)**. We show SVN$^3$'s predicted categories with top-2 highest probabilities below each test image. Green and Orange colored labels under the images denote the correct and incorrect predictions, respectively.

37

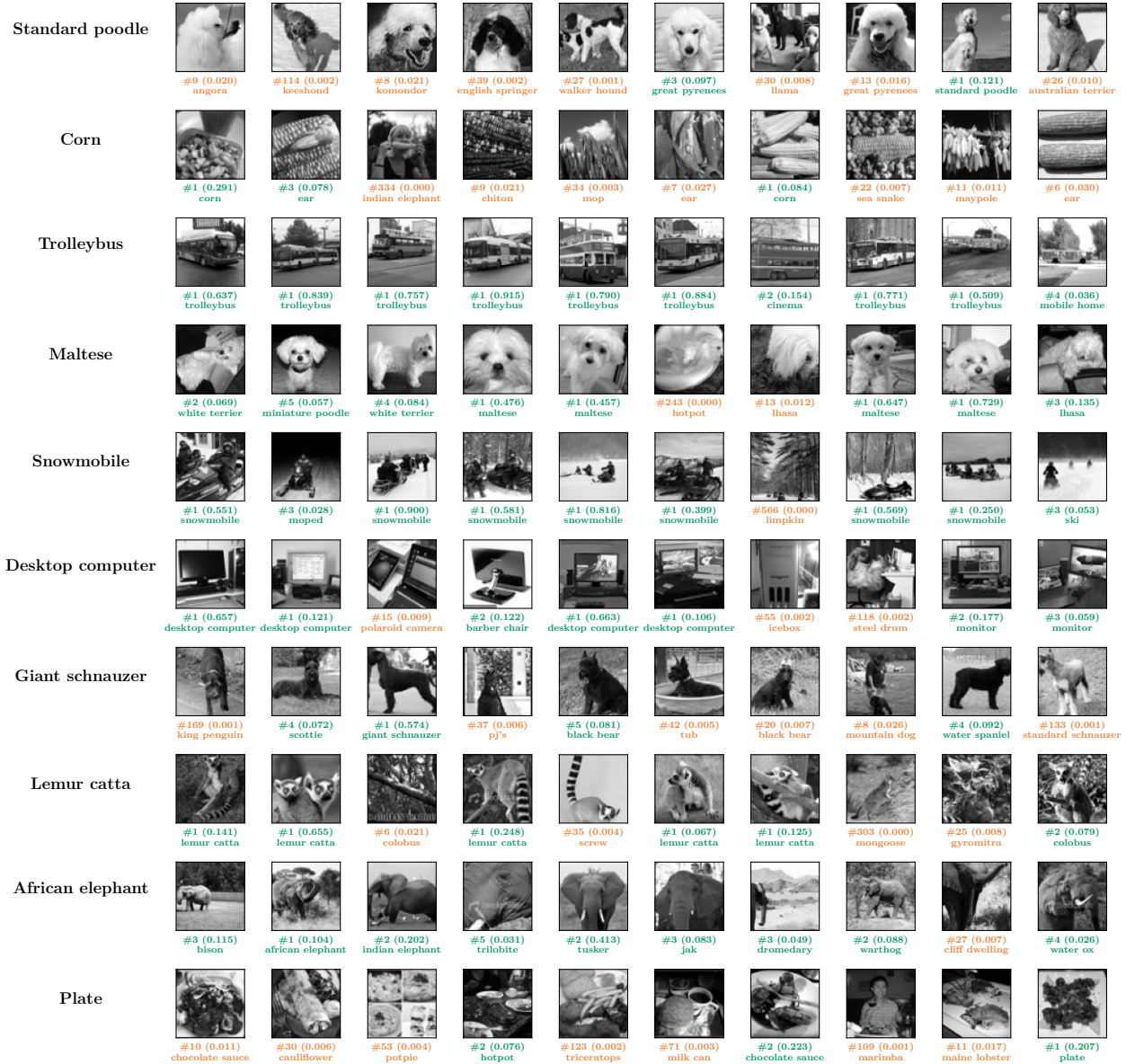

Figure S16: **Additional experimental classification results on random samples from ImageNet-1000 validation set (Part-1)**. 10 classes are randomly selected from 1000 categories in ImageNet (class names are attached on the left side). Below each test image, we show the rank of the ground truth class in SVN$^3$'s predictions, together with the corresponding predicted probabilities in the first row; and the top-1 predicted category label in the second row. Green and Orange colored labels under the images denote the ground truth classes are within or without top-5 predicted categories, respectively.

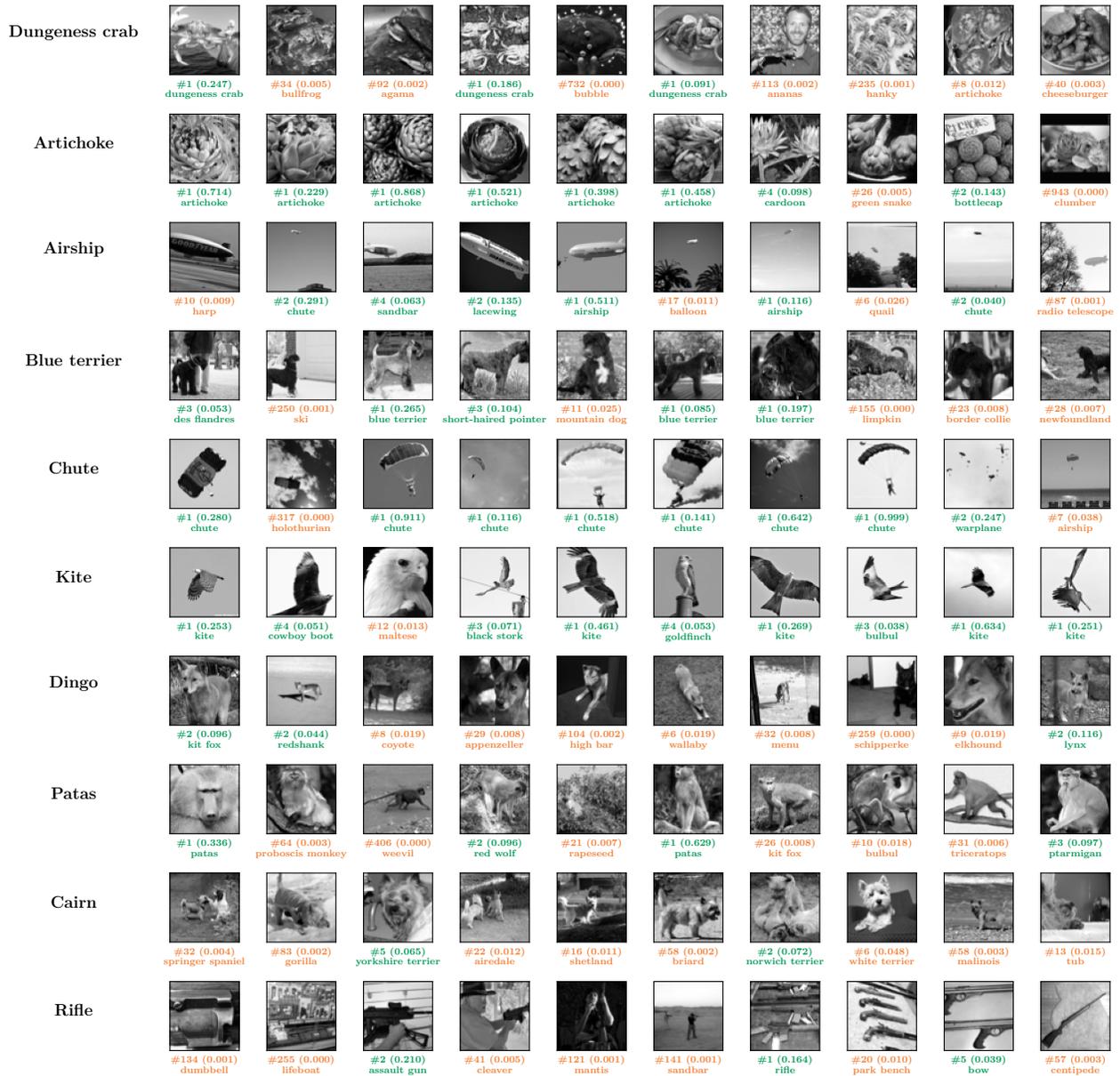

Figure S16: **Additional experimental classification results on random samples from ImageNet-1000 validation set (Part-2)**. 10 classes are randomly selected from 1000 categories in ImageNet (class names are attached on the left side). Below each test image, we show the rank of the ground truth class in SVN$^3$'s predictions, together with the corresponding predicted probabilities in the first row; and the top-1 predicted category label in the second row. Green and Orange colored labels under the images denote the ground truth classes are within or without top-5 predicted categories, respectively.

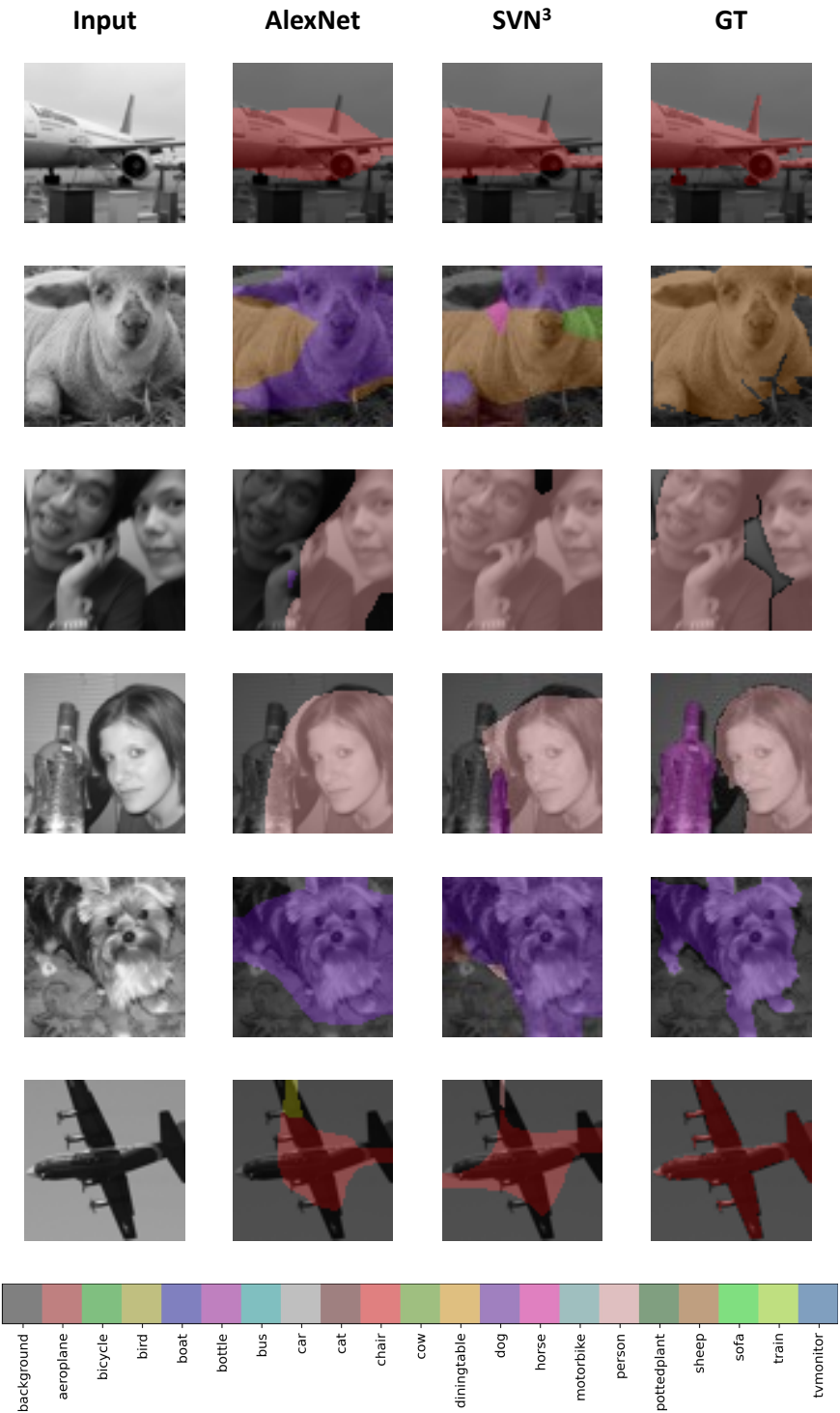

Figure S17: **Additional semantic segmentation results on random samples from PASCAL VOC test set (Part-1)**. We display the input images, the segmentation results produced by AlexNet and SVN$^3$, as well as the ground truth segmentation masks column by column (color encodes the class information as displayed in the bottom).

40

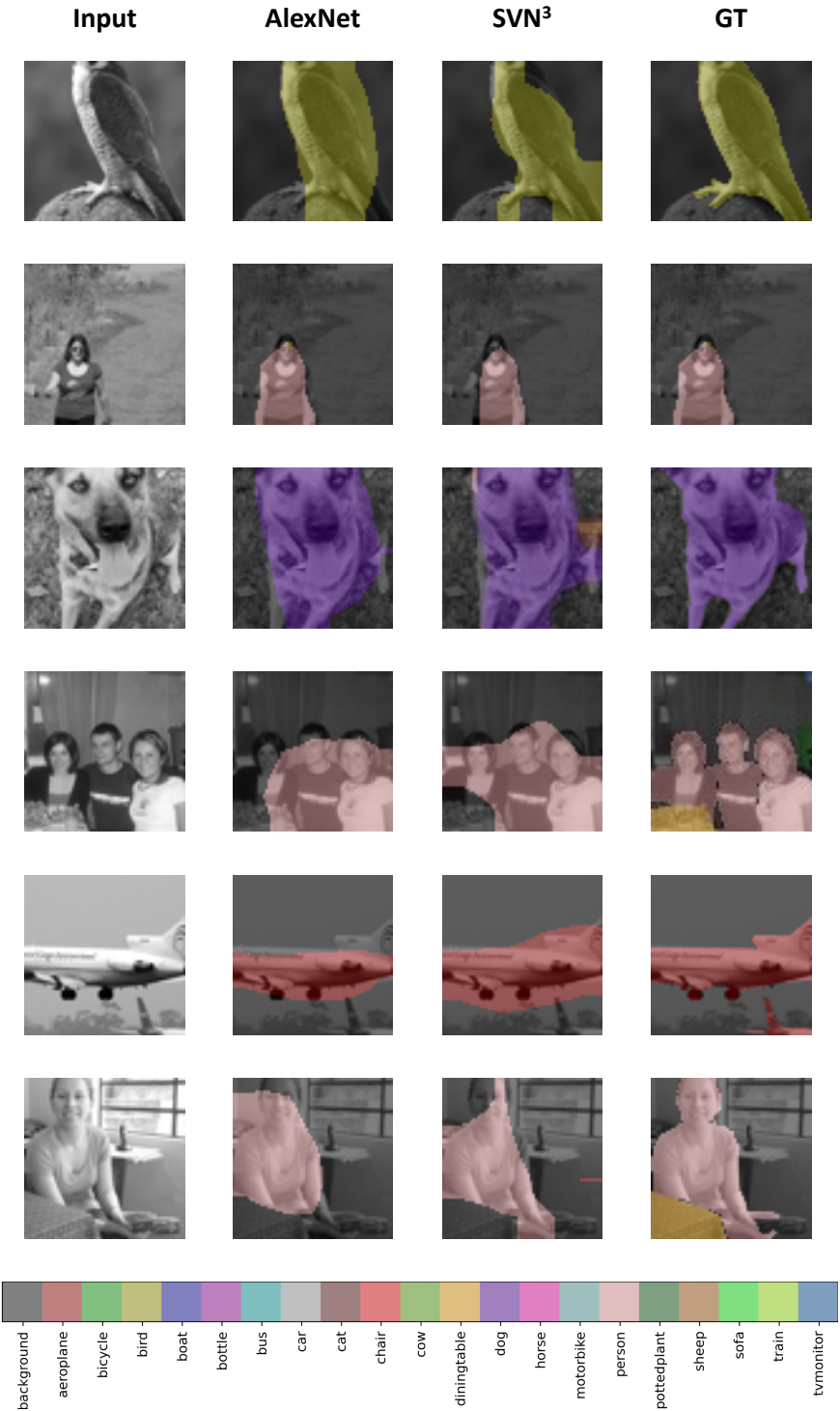

Figure S17: **Additional semantic segmentation results on random samples from PASCAL VOC test set (Part-2).** We display the input images, the segmentation results produced by AlexNet and SVN$^3$, as well as the ground truth segmentation masks column by column (color encodes the class information as displayed in the bottom).

41

Table S6: **Performance comparison of SVN$^3$ and AlexNet on PASCAL VOC image segmentation dataset.** We report three metrics from common semantic segmentation evaluations that are variations on pixel accuracy and region intersection over union (IU). See Supplementary Note 9 for detailed definitions of these metrics.

| SEGMENTATION | ALEXNET | SVN$^3$ (SIMULATION) | SVN$^3$ (EXPERIMENT) |
|---|---|---|---|
| PIXEL ACCURACY ↑ | 66.34 % | 65.73 % | 63.14 % |
| MEAN ACCURACY ↑ | 58.91 % | 57.29 % | 55.76 % |
| MEAN IOU ↑ | 26.92 | 24.62 | 23.64 |


\fi


\begin{thebibliography}{10}
  \expandafter\ifx\csname url\endcsname\relax
    \def\url#1{\texttt{#1}}\fi
  \expandafter\ifx\csname urlprefix\endcsname\relax\def\urlprefix{URL }\fi
  \providecommand{\bibinfo}[2]{#2}
  \providecommand{\eprint}[2][]{\url{#2}}
  
  \bibitem{moore1998cramming}
  \bibinfo{author}{Moore, G.~E.}
  \newblock \bibinfo{title}{Cramming more components onto integrated circuits}.
  \newblock \emph{\bibinfo{journal}{Proceedings of the IEEE}}
    \textbf{\bibinfo{volume}{86}}, \bibinfo{pages}{82--85}
    (\bibinfo{year}{1998}).
  
  \bibitem{waldrop2016chips}
  \bibinfo{author}{Waldrop, M.~M.}
  \newblock \bibinfo{title}{The chips are down for moore’s law}.
  \newblock \emph{\bibinfo{journal}{Nature News}} \textbf{\bibinfo{volume}{530}},
    \bibinfo{pages}{144} (\bibinfo{year}{2016}).
  
  \bibitem{lecun2015deep}
  \bibinfo{author}{LeCun, Y.}, \bibinfo{author}{Bengio, Y.} \&
    \bibinfo{author}{Hinton, G.}
  \newblock \bibinfo{title}{Deep learning}.
  \newblock \emph{\bibinfo{journal}{Nature}} \textbf{\bibinfo{volume}{521}},
    \bibinfo{pages}{436--444} (\bibinfo{year}{2015}).
  
  \bibitem{sevilla2022compute}
  \bibinfo{author}{Sevilla, J.} \emph{et~al.}
  \newblock \bibinfo{title}{Compute trends across three eras of machine
    learning}.
  \newblock \emph{\bibinfo{journal}{arXiv preprint arXiv:2202.05924}}
    (\bibinfo{year}{2022}).
  
  \bibitem{horowitz20141}
  \bibinfo{author}{Horowitz, M.}
  \newblock \bibinfo{title}{1.1 computing's energy problem (and what we can do
    about it)}.
  \newblock In \emph{\bibinfo{booktitle}{2014 IEEE International Solid-State
    Circuits Conference Digest of Technical Papers (ISSCC)}},
    \bibinfo{pages}{10--14} (\bibinfo{organization}{IEEE}, \bibinfo{year}{2014}).
  
  \bibitem{solli2015analog}
  \bibinfo{author}{Solli, D.~R.} \& \bibinfo{author}{Jalali, B.}
  \newblock \bibinfo{title}{Analog optical computing}.
  \newblock \emph{\bibinfo{journal}{Nature Photonics}}
    \textbf{\bibinfo{volume}{9}}, \bibinfo{pages}{704--706}
    (\bibinfo{year}{2015}).
  
  \bibitem{caulfield2010future}
  \bibinfo{author}{Caulfield, H.~J.} \& \bibinfo{author}{Dolev, S.}
  \newblock \bibinfo{title}{Why future supercomputing requires optics}.
  \newblock \emph{\bibinfo{journal}{Nature Photonics}}
    \textbf{\bibinfo{volume}{4}}, \bibinfo{pages}{261--263}
    (\bibinfo{year}{2010}).
  
  \bibitem{miller2017attojoule}
  \bibinfo{author}{Miller, D.~A.}
  \newblock \bibinfo{title}{Attojoule optoelectronics for low-energy information
    processing and communications}.
  \newblock \emph{\bibinfo{journal}{Journal of Lightwave Technology}}
    \textbf{\bibinfo{volume}{35}}, \bibinfo{pages}{346--396}
    (\bibinfo{year}{2017}).
  
  \bibitem{mcmahon2023physics}
  \bibinfo{author}{McMahon, P.~L.}
  \newblock \bibinfo{title}{The physics of optical computing}.
  \newblock \emph{\bibinfo{journal}{Nature Reviews Physics}}
    (\bibinfo{year}{2023}).
  
  \bibitem{miller2010optical}
  \bibinfo{author}{Miller, D.~A.}
  \newblock \bibinfo{title}{Are optical transistors the logical next step?}
  \newblock \emph{\bibinfo{journal}{Nature Photonics}}
    \textbf{\bibinfo{volume}{4}}, \bibinfo{pages}{3--5} (\bibinfo{year}{2010}).
  
  \bibitem{tucker2010role}
  \bibinfo{author}{Tucker, R.~S.}
  \newblock \bibinfo{title}{The role of optics in computing}.
  \newblock \emph{\bibinfo{journal}{Nature Photonics}}
    \textbf{\bibinfo{volume}{4}}, \bibinfo{pages}{405--405}
    (\bibinfo{year}{2010}).
  
  \bibitem{wetzstein2020inference}
  \bibinfo{author}{Wetzstein, G.} \emph{et~al.}
  \newblock \bibinfo{title}{Inference in artificial intelligence with deep optics
    and photonics}.
  \newblock \emph{\bibinfo{journal}{Nature}} \textbf{\bibinfo{volume}{588}},
    \bibinfo{pages}{39--47} (\bibinfo{year}{2020}).
  
  \bibitem{shastri2021photonics}
  \bibinfo{author}{Shastri, B.~J.} \emph{et~al.}
  \newblock \bibinfo{title}{Photonics for artificial intelligence and
    neuromorphic computing}.
  \newblock \emph{\bibinfo{journal}{Nature Photonics}}
    \textbf{\bibinfo{volume}{15}}, \bibinfo{pages}{102--114}
    (\bibinfo{year}{2021}).
  
  \bibitem{wu2021analog}
  \bibinfo{author}{Wu, J.} \emph{et~al.}
  \newblock \bibinfo{title}{Analog optical computing for artificial
    intelligence}.
  \newblock \emph{\bibinfo{journal}{Engineering}}  (\bibinfo{year}{2021}).
  
  \bibitem{liu2016fully}
  \bibinfo{author}{Liu, W.} \emph{et~al.}
  \newblock \bibinfo{title}{A fully reconfigurable photonic integrated signal
    processor}.
  \newblock \emph{\bibinfo{journal}{Nature Photonics}}
    \textbf{\bibinfo{volume}{10}}, \bibinfo{pages}{190--195}
    (\bibinfo{year}{2016}).
  
  \bibitem{kwon2018nonlocal}
  \bibinfo{author}{Kwon, H.}, \bibinfo{author}{Sounas, D.},
    \bibinfo{author}{Cordaro, A.}, \bibinfo{author}{Polman, A.} \&
    \bibinfo{author}{Al{\`u}, A.}
  \newblock \bibinfo{title}{Nonlocal metasurfaces for optical signal processing}.
  \newblock \emph{\bibinfo{journal}{Physical Review Letters}}
    \textbf{\bibinfo{volume}{121}}, \bibinfo{pages}{173004}
    (\bibinfo{year}{2018}).
  
  \bibitem{silva2014performing}
  \bibinfo{author}{Silva, A.} \emph{et~al.}
  \newblock \bibinfo{title}{Performing mathematical operations with
    metamaterials}.
  \newblock \emph{\bibinfo{journal}{Science}} \textbf{\bibinfo{volume}{343}},
    \bibinfo{pages}{160--163} (\bibinfo{year}{2014}).
  
  \bibitem{zhu2017plasmonic}
  \bibinfo{author}{Zhu, T.} \emph{et~al.}
  \newblock \bibinfo{title}{Plasmonic computing of spatial differentiation}.
  \newblock \emph{\bibinfo{journal}{Nature Communications}}
    \textbf{\bibinfo{volume}{8}}, \bibinfo{pages}{1--6} (\bibinfo{year}{2017}).
  
  \bibitem{ferrera2010chip}
  \bibinfo{author}{Ferrera, M.} \emph{et~al.}
  \newblock \bibinfo{title}{On-chip cmos-compatible all-optical integrator}.
  \newblock \emph{\bibinfo{journal}{Nature Communications}}
    \textbf{\bibinfo{volume}{1}}, \bibinfo{pages}{1--5} (\bibinfo{year}{2010}).
  
  \bibitem{xu2020scalable}
  \bibinfo{author}{Xu, X.-Y.} \emph{et~al.}
  \newblock \bibinfo{title}{A scalable photonic computer solving the subset sum
    problem}.
  \newblock \emph{\bibinfo{journal}{Science Advances}}
    \textbf{\bibinfo{volume}{6}}, \bibinfo{pages}{eaay5853}
    (\bibinfo{year}{2020}).
  
  \bibitem{mohammadi2019inverse}
  \bibinfo{author}{Mohammadi~Estakhri, N.}, \bibinfo{author}{Edwards, B.} \&
    \bibinfo{author}{Engheta, N.}
  \newblock \bibinfo{title}{Inverse-designed metastructures that solve
    equations}.
  \newblock \emph{\bibinfo{journal}{Science}} \textbf{\bibinfo{volume}{363}},
    \bibinfo{pages}{1333--1338} (\bibinfo{year}{2019}).
  
  \bibitem{feldmann2019all}
  \bibinfo{author}{Feldmann, J.}, \bibinfo{author}{Youngblood, N.},
    \bibinfo{author}{Wright, C.~D.}, \bibinfo{author}{Bhaskaran, H.} \&
    \bibinfo{author}{Pernice, W.~H.}
  \newblock \bibinfo{title}{All-optical spiking neurosynaptic networks with
    self-learning capabilities}.
  \newblock \emph{\bibinfo{journal}{Nature}} \textbf{\bibinfo{volume}{569}},
    \bibinfo{pages}{208--214} (\bibinfo{year}{2019}).
  
  \bibitem{xu202111}
  \bibinfo{author}{Xu, X.} \emph{et~al.}
  \newblock \bibinfo{title}{11 tops photonic convolutional accelerator for
    optical neural networks}.
  \newblock \emph{\bibinfo{journal}{Nature}} \textbf{\bibinfo{volume}{589}},
    \bibinfo{pages}{44--51} (\bibinfo{year}{2021}).
  
  \bibitem{shen2017deep}
  \bibinfo{author}{Shen, Y.} \emph{et~al.}
  \newblock \bibinfo{title}{Deep learning with coherent nanophotonic circuits}.
  \newblock \emph{\bibinfo{journal}{Nature photonics}}
    \textbf{\bibinfo{volume}{11}}, \bibinfo{pages}{441--446}
    (\bibinfo{year}{2017}).
  
  \bibitem{feldmann2021parallel}
  \bibinfo{author}{Feldmann, J.} \emph{et~al.}
  \newblock \bibinfo{title}{Parallel convolutional processing using an integrated
    photonic tensor core}.
  \newblock \emph{\bibinfo{journal}{Nature}} \textbf{\bibinfo{volume}{589}},
    \bibinfo{pages}{52--58} (\bibinfo{year}{2021}).
  
  \bibitem{ashtiani2022chip}
  \bibinfo{author}{Ashtiani, F.}, \bibinfo{author}{Geers, A.~J.} \&
    \bibinfo{author}{Aflatouni, F.}
  \newblock \bibinfo{title}{An on-chip photonic deep neural network for image
    classification}.
  \newblock \emph{\bibinfo{journal}{Nature}} \bibinfo{pages}{1--6}
    (\bibinfo{year}{2022}).
  
  \bibitem{tait2017neuromorphic}
  \bibinfo{author}{Tait, A.~N.} \emph{et~al.}
  \newblock \bibinfo{title}{Neuromorphic photonic networks using silicon photonic
    weight banks}.
  \newblock \emph{\bibinfo{journal}{Scientific Reports}}
    \textbf{\bibinfo{volume}{7}}, \bibinfo{pages}{1--10} (\bibinfo{year}{2017}).
  
  \bibitem{teugin2021scalable}
  \bibinfo{author}{Te{\u{g}}in, U.}, \bibinfo{author}{Y{\i}ld{\i}r{\i}m, M.},
    \bibinfo{author}{O{\u{g}}uz, {\.I}.}, \bibinfo{author}{Moser, C.} \&
    \bibinfo{author}{Psaltis, D.}
  \newblock \bibinfo{title}{Scalable optical learning operator}.
  \newblock \emph{\bibinfo{journal}{Nature Computational Science}}
    \textbf{\bibinfo{volume}{1}}, \bibinfo{pages}{542--549}
    (\bibinfo{year}{2021}).
  
  \bibitem{lin2018all}
  \bibinfo{author}{Lin, X.} \emph{et~al.}
  \newblock \bibinfo{title}{All-optical machine learning using diffractive deep
    neural networks}.
  \newblock \emph{\bibinfo{journal}{Science}} \textbf{\bibinfo{volume}{361}},
    \bibinfo{pages}{1004--1008} (\bibinfo{year}{2018}).
  
  \bibitem{mengu2019analysis}
  \bibinfo{author}{Mengu, D.}, \bibinfo{author}{Luo, Y.},
    \bibinfo{author}{Rivenson, Y.} \& \bibinfo{author}{Ozcan, A.}
  \newblock \bibinfo{title}{Analysis of diffractive optical neural networks and
    their integration with electronic neural networks}.
  \newblock \emph{\bibinfo{journal}{IEEE Journal of Selected Topics in Quantum
    Electronics}} \textbf{\bibinfo{volume}{26}}, \bibinfo{pages}{1--14}
    (\bibinfo{year}{2019}).
  
  \bibitem{yan2019fourier}
  \bibinfo{author}{Yan, T.} \emph{et~al.}
  \newblock \bibinfo{title}{Fourier-space diffractive deep neural network}.
  \newblock \emph{\bibinfo{journal}{Physical Review Letters}}
    \textbf{\bibinfo{volume}{123}}, \bibinfo{pages}{023901}
    (\bibinfo{year}{2019}).
  
  \bibitem{rahman2021ensemble}
  \bibinfo{author}{Rahman, M. S.~S.}, \bibinfo{author}{Li, J.},
    \bibinfo{author}{Mengu, D.}, \bibinfo{author}{Rivenson, Y.} \&
    \bibinfo{author}{Ozcan, A.}
  \newblock \bibinfo{title}{Ensemble learning of diffractive optical networks}.
  \newblock \emph{\bibinfo{journal}{Light: Science \& Applications}}
    \textbf{\bibinfo{volume}{10}}, \bibinfo{pages}{1--13} (\bibinfo{year}{2021}).
  
  \bibitem{luo2022metasurface}
  \bibinfo{author}{Luo, X.} \emph{et~al.}
  \newblock \bibinfo{title}{Metasurface-enabled on-chip multiplexed diffractive
    neural networks in the visible}.
  \newblock \emph{\bibinfo{journal}{Light: Science \& Applications}}
    \textbf{\bibinfo{volume}{11}}, \bibinfo{pages}{1--11} (\bibinfo{year}{2022}).
  
  \bibitem{hamerly2019large}
  \bibinfo{author}{Hamerly, R.}, \bibinfo{author}{Bernstein, L.},
    \bibinfo{author}{Sludds, A.}, \bibinfo{author}{Solja{\v{c}}i{\'c}, M.} \&
    \bibinfo{author}{Englund, D.}
  \newblock \bibinfo{title}{Large-scale optical neural networks based on
    photoelectric multiplication}.
  \newblock \emph{\bibinfo{journal}{Physical Review X}}
    \textbf{\bibinfo{volume}{9}}, \bibinfo{pages}{021032} (\bibinfo{year}{2019}).
  
  \bibitem{zhou2021large}
  \bibinfo{author}{Zhou, T.} \emph{et~al.}
  \newblock \bibinfo{title}{Large-scale neuromorphic optoelectronic computing
    with a reconfigurable diffractive processing unit}.
  \newblock \emph{\bibinfo{journal}{Nature Photonics}}
    \textbf{\bibinfo{volume}{15}}, \bibinfo{pages}{367--373}
    (\bibinfo{year}{2021}).
  
  \bibitem{shi2022loen}
  \bibinfo{author}{Shi, W.} \emph{et~al.}
  \newblock \bibinfo{title}{Loen: Lensless opto-electronic neural network
    empowered machine vision}.
  \newblock \emph{\bibinfo{journal}{Light: Science \& Applications}}
    \textbf{\bibinfo{volume}{11}}, \bibinfo{pages}{1--12} (\bibinfo{year}{2022}).
  
  \bibitem{zheng2022meta}
  \bibinfo{author}{Zheng, H.} \emph{et~al.}
  \newblock \bibinfo{title}{Meta-optic accelerators for object classifiers}.
  \newblock \emph{\bibinfo{journal}{Science Advances}}
    \textbf{\bibinfo{volume}{8}}, \bibinfo{pages}{eabo6410}
    (\bibinfo{year}{2022}).
  
  \bibitem{chang2018hybrid}
  \bibinfo{author}{Chang, J.}, \bibinfo{author}{Sitzmann, V.},
    \bibinfo{author}{Dun, X.}, \bibinfo{author}{Heidrich, W.} \&
    \bibinfo{author}{Wetzstein, G.}
  \newblock \bibinfo{title}{Hybrid optical-electronic convolutional neural
    networks with optimized diffractive optics for image classification}.
  \newblock \emph{\bibinfo{journal}{Scientific Reports}}
    \textbf{\bibinfo{volume}{8}}, \bibinfo{pages}{1--10} (\bibinfo{year}{2018}).
  
  \bibitem{chen2023accel}
  \bibinfo{author}{Chen, Y.} \& \bibinfo{author}{et~al.}
  \newblock \bibinfo{title}{All-analog photoelectronic chip for high-speed vision
    tasks}.
  \newblock \emph{\bibinfo{journal}{Nature}}  (\bibinfo{year}{2023}).
  
  \bibitem{colburn2019optical}
  \bibinfo{author}{Colburn, S.}, \bibinfo{author}{Chu, Y.},
    \bibinfo{author}{Shilzerman, E.} \& \bibinfo{author}{Majumdar, A.}
  \newblock \bibinfo{title}{Optical frontend for a convolutional neural network}.
  \newblock \emph{\bibinfo{journal}{Applied Optics}}
    \textbf{\bibinfo{volume}{58}}, \bibinfo{pages}{3179--3186}
    (\bibinfo{year}{2019}).
  
  \bibitem{lecun1989handwritten}
  \bibinfo{author}{LeCun, Y.} \emph{et~al.}
  \newblock \bibinfo{title}{Handwritten digit recognition with a back-propagation
    network}.
  \newblock \emph{\bibinfo{journal}{Advances in Neural Information Processing
    Systems}} \textbf{\bibinfo{volume}{2}} (\bibinfo{year}{1989}).
  
  \bibitem{krizhevsky2012imagenet}
  \bibinfo{author}{Krizhevsky, A.}, \bibinfo{author}{Sutskever, I.} \&
    \bibinfo{author}{Hinton, G.~E.}
  \newblock \bibinfo{title}{Imagenet classification with deep convolutional
    neural networks}.
  \newblock \emph{\bibinfo{journal}{Advances in Neural Information Processing
    Systems}} \textbf{\bibinfo{volume}{25}} (\bibinfo{year}{2012}).
  
  \bibitem{khorasaninejad2017metalenses}
  \bibinfo{author}{Khorasaninejad, M.} \& \bibinfo{author}{Capasso, F.}
  \newblock \bibinfo{title}{Metalenses: Versatile multifunctional photonic
    components}.
  \newblock \emph{\bibinfo{journal}{Science}} \textbf{\bibinfo{volume}{358}},
    \bibinfo{pages}{eaam8100} (\bibinfo{year}{2017}).
  
  \bibitem{dorrah2022tunable}
  \bibinfo{author}{Dorrah, A.~H.} \& \bibinfo{author}{Capasso, F.}
  \newblock \bibinfo{title}{Tunable structured light with flat optics}.
  \newblock \emph{\bibinfo{journal}{Science}} \textbf{\bibinfo{volume}{376}},
    \bibinfo{pages}{eabi6860} (\bibinfo{year}{2022}).
  
  \bibitem{deng2009imagenet}
  \bibinfo{author}{Deng, J.} \emph{et~al.}
  \newblock \bibinfo{title}{Imagenet: A large-scale hierarchical image database}.
  \newblock In \emph{\bibinfo{booktitle}{2009 IEEE conference on computer vision
    and pattern recognition}}, \bibinfo{pages}{248--255}
    (\bibinfo{organization}{Ieee}, \bibinfo{year}{2009}).
  
  \bibitem{everingham2010pascal}
  \bibinfo{author}{Everingham, M.}, \bibinfo{author}{Van~Gool, L.},
    \bibinfo{author}{Williams, C.~K.}, \bibinfo{author}{Winn, J.} \&
    \bibinfo{author}{Zisserman, A.}
  \newblock \bibinfo{title}{The pascal visual object classes (voc) challenge}.
  \newblock \emph{\bibinfo{journal}{International journal of computer vision}}
    \textbf{\bibinfo{volume}{88}}, \bibinfo{pages}{303--338}
    (\bibinfo{year}{2010}).
  
  \bibitem{krizhevsky2009learning}
  \bibinfo{author}{Krizhevsky, A.}, \bibinfo{author}{Hinton, G.} \emph{et~al.}
  \newblock \bibinfo{title}{Learning multiple layers of features from tiny
    images}  (\bibinfo{year}{2009}).
  
  \bibitem{fu2022ultracompact}
  \bibinfo{author}{Fu, W.} \emph{et~al.}
  \newblock \bibinfo{title}{Ultracompact meta-imagers for arbitrary all-optical
    convolution}.
  \newblock \emph{\bibinfo{journal}{Light: Science \& Applications}}
    \textbf{\bibinfo{volume}{11}}, \bibinfo{pages}{62} (\bibinfo{year}{2022}).
  
  \bibitem{nilsback2008automated}
  \bibinfo{author}{Nilsback, M.-E.} \& \bibinfo{author}{Zisserman, A.}
  \newblock \bibinfo{title}{Automated flower classification over a large number
    of classes}.
  \newblock In \emph{\bibinfo{booktitle}{2008 Sixth Indian conference on computer
    vision, graphics \& image processing}}, \bibinfo{pages}{722--729}
    (\bibinfo{organization}{IEEE}, \bibinfo{year}{2008}).
  
  \bibitem{bossard2014food}
  \bibinfo{author}{Bossard, L.}, \bibinfo{author}{Guillaumin, M.} \&
    \bibinfo{author}{Van~Gool, L.}
  \newblock \bibinfo{title}{Food-101--mining discriminative components with
    random forests}.
  \newblock In \emph{\bibinfo{booktitle}{Computer Vision--ECCV 2014: 13th
    European Conference, Zurich, Switzerland, September 6-12, 2014, Proceedings,
    Part VI 13}}, \bibinfo{pages}{446--461} (\bibinfo{organization}{Springer},
    \bibinfo{year}{2014}).
  
  \bibitem{parkhi2012cats}
  \bibinfo{author}{Parkhi, O.~M.}, \bibinfo{author}{Vedaldi, A.},
    \bibinfo{author}{Zisserman, A.} \& \bibinfo{author}{Jawahar, C.}
  \newblock \bibinfo{title}{Cats and dogs}.
  \newblock In \emph{\bibinfo{booktitle}{2012 IEEE conference on computer vision
    and pattern recognition}}, \bibinfo{pages}{3498--3505}
    (\bibinfo{organization}{IEEE}, \bibinfo{year}{2012}).
  
  \end{thebibliography}
\end{document}